\documentclass[journal,onecolumn]{IEEEtran}
\usepackage{moreverb,url}
\usepackage{lineno}
\usepackage[square,sort,comma,numbers]{natbib}
\usepackage{makecell} 
\usepackage{ulem}
\usepackage{hyperref}
\usepackage[utf8]{inputenc} 
\usepackage[T1]{fontenc}    
\usepackage{booktabs}       
\usepackage{amsfonts}       
\usepackage{nicefrac}       
\usepackage{microtype}      
\usepackage{xcolor}         
\usepackage{graphicx}
\usepackage{makecell}
\usepackage{multirow}
\usepackage{amssymb} 
\usepackage{amsmath} 
\usepackage{tabularx} 
\usepackage{algorithmic}
\usepackage{textcomp}
\usepackage{graphics}
\usepackage{graphicx}
\usepackage{subcaption}
\usepackage{multicol}
\usepackage{bm}
\usepackage{utfsym}
\usepackage{mathrsfs}
\usepackage{array}
\usepackage{colortbl}
\usepackage{xcolor}
\usepackage[most]{tcolorbox}
\usepackage{wrapfig}
\usepackage{enumitem}
\usepackage{amsmath,amssymb,amsthm,mathtools}

\usepackage{ marvosym }
\usepackage{adjustbox}
\newcolumntype{C}[1]{>{\centering\arraybackslash}m{#1}}

\newcommand{\red}[1]{\textcolor{black}{#1}}
\newcommand{\gray}[1]{}

\definecolor{mycyan}{cmyk}{.3,0,0,0}

\newcommand{\minghuan}[1]{\textcolor{black}{#1}}

\newcommand\BibTeX{{\rmfamily B\kern-.05em \textsc{i\kern-.025em b}\kern-.08em
T\kern-.1667em\lower.7ex\hbox{E}\kern-.125emX}}
\newcommand{\secref}[1]{Section \ref{#1}}

\newcommand{\question}[2]{
    \vspace{-0.1cm}
    \begin{tcolorbox}[
        colback=white!90!gray,     
        colframe=teal!60!black,     
        arc=5pt,                    
        boxsep=5pt,                 
        left=10pt,                  
        right=10pt,                 
        top=2pt,                    
        bottom=2pt,                 
        boxrule=0.8pt,              
        drop shadow=gray!50!white,  
        enhanced jigsaw             
    ]
        \noindent\textbf{\sffamily{Question #1:}} #2
    \label{question:#1}
    \end{tcolorbox}
    \vspace{-0.1cm}
}

\newcommand{\finding}[2]{
    \vspace{-0.1cm}
    \begin{tcolorbox}[
        colback=white!90!gray,     
        colframe=teal!60!black,     
        arc=5pt,                    
        boxsep=5pt,                 
        left=10pt,                  
        right=10pt,                 
        top=2pt,                    
        bottom=2pt,                 
        boxrule=0.8pt,              
        drop shadow=gray!50!white,  
        enhanced jigsaw             
    ]
        \noindent\textbf{\sffamily{Finding #1:}} #2
    \end{tcolorbox}
    \vspace{-0.1cm}
}

\usepackage{pifont}
\newcommand{\cmark}{\ding{51}}%
\newcommand{\xmark}{\ding{55}}%

\usepackage{newfloat} 

\DeclareFloatingEnvironment[
    fileext=loex, 
    listname={List of Extended Figures}, 
    name=Extended Figure, 
    placement=htbp, 
]{extendedfigure}

\DeclareFloatingEnvironment[
    fileext=loex, 
    listname={List of Extended Figures}, 
    name=Supplementary Figure, 
    placement=htbp, 
]{supplementaryfigure}

\DeclareFloatingEnvironment[
    fileext=lotex,
    listname={List of Extended Tables},
    name=Extended Table,
    placement=htbp,
]{extendedtable}

\DeclareFloatingEnvironment[
    fileext=lotex,
    listname={List of Extended Tables},
    name=Supplementary Table,
    placement=htbp,
]{supplementarytable}

\usepackage{etoolbox}

\makeatletter

\let\orig@extendedtable\extendedtable
\let\orig@endextendedtable\endextendedtable

\renewenvironment{extendedtable}[1][htbp]
  {%
    \begin{table}[#1]
    \def\@captype{extendedtable}
    \small 
    \centering 
    \setlength{\tabcolsep}{6pt} 
  }
  {%
    \end{table}%
  }

\renewenvironment{extendedtable*}[1][tbp]
  {%
    \begin{table*}[#1]
    \def\@captype{extendedtable}
    \small
    \centering
    \setlength{\tabcolsep}{6pt}
    
  }
  {%
    \end{table*}%
  }
\makeatother

\begin{document}

\newtheorem{lemma}{Lemma}
\newtheorem{corollary}{Corollary}
\newtheorem{assumption}{Assumption}
\newtheorem{observation}{Observation}

\newcommand{\fig}[1]{Fig.~\ref{#1}}
\newcommand{\eq}[1]{Eq.~(\ref{#1})}
\newcommand{\ineq}[1]{Ineq.~(\ref{#1})}
\newcommand{\tb}[1]{Tab.~\ref{#1}}
\newcommand{\se}[1]{Section~\ref{#1}}
\newcommand{\ap}[1]{Appendix~\ref{#1}}
\newcommand{\pa}[1]{Part~\ref{#1}}
\newcommand{\lm}[1]{Lemma~\ref{#1}}
\newcommand{\prop}[1]{Proposition~\ref{#1}}
\newcommand{\alg}[1]{Algo.~\ref{#1}}
\newcommand{\theo}[1]{Theorem~\ref{#1}}
\newcommand{\defi}[1]{Definition~\ref{#1}}
\newcommand{\assum}[1]{Assumption~\ref{#1}}
\newcommand{\observe}[1]{Observation~\ref{#1}}

\newcommand*{\dif}{\mathop{}\!\mathrm{d}}
\newcommand*{\kl}{\mathrm{KL}}
\newcommand{\bbI}{\ensuremath{\mathbb{I}}} 
\newcommand{\bbE}{\ensuremath{\mathbb{E}}} 
\newcommand{\bbR}{\ensuremath{\mathbb{R}}} 
\newcommand{\caA}{\ensuremath{\mathcal{A}}} 
\newcommand{\caS}{\ensuremath{\mathcal{S}}} 
\newcommand{\caAt}{\ensuremath{\mathcal{\tilde{A}}}} 
\newcommand{\caSt}{\ensuremath{\mathcal{\tilde{S}}}} 
\newcommand{\caN}{\ensuremath{\mathcal{N}}} 
\newcommand{\caM}{\ensuremath{\mathcal{M}}} 
\newcommand{\caMt}{\ensuremath{\mathcal{\tilde{M}}}}
\newcommand{\caD}{\ensuremath{\mathcal{D}}} 
\newcommand{\caG}{\ensuremath{\mathcal{G}}} 
\newcommand{\caL}{\ensuremath{\mathcal{L}}} 
\newcommand{\caT}{\ensuremath{\mathcal{T}}} 
\newcommand{\caO}{\ensuremath{\mathcal{O}}} 
\newcommand{\caTt}{\ensuremath{\mathcal{\tilde{T}}}}
\newcommand{\caB}{\ensuremath{\mathcal{B}}} 
\newcommand{\kld}{\text{D}_{\text{KL}}} 
\newcommand{\jsd}{\text{D}_{\text{JS}}}
\newcommand{\fd}{\text{D}_{\text{f}}} 
\newcommand{\iter}[2]{{#1}^{(#2)}}
\newcommand{\piE}{{\pi_E}}
\newcommand{\hr}{\hat{r}}
\newcommand{\hpi}{\hat{\pi}}

\newcommand{\hytt}[1]{\texttt{\hyphenchar\font=\defaulthyphenchar #1}}

\newcommand{\tc}[1]{\textcolor{red}{#1}} 

\newcommand{\figleft}{{\em (Left)}}
\newcommand{\figcenter}{{\em (Center)}}
\newcommand{\figright}{{\em (Right)}}
\newcommand{\figtop}{{\em (Top)}}
\newcommand{\figbottom}{{\em (Bottom)}}
\newcommand{\captiona}{{\em (a)}}
\newcommand{\captionb}{{\em (b)}}
\newcommand{\captionc}{{\em (c)}}
\newcommand{\captiond}{{\em (d)}}

\newcommand{\newterm}[1]{{\bf #1}}

\def\figref#1{figure~\ref{#1}}
\def\Figref#1{Figure~\ref{#1}}
\def\twofigref#1#2{figures \ref{#1} and \ref{#2}}
\def\quadfigref#1#2#3#4{figures \ref{#1}, \ref{#2}, \ref{#3} and \ref{#4}}
\def\secref#1{section~\ref{#1}}
\def\Secref#1{Section~\ref{#1}}
\def\twosecrefs#1#2{sections \ref{#1} and \ref{#2}}
\def\secrefs#1#2#3{sections \ref{#1}, \ref{#2} and \ref{#3}}
\def\eqref#1{equation~\ref{#1}}
\def\Eqref#1{Equation~\ref{#1}}
\def\plaineqref#1{\ref{#1}}
\def\chapref#1{chapter~\ref{#1}}
\def\Chapref#1{Chapter~\ref{#1}}
\def\rangechapref#1#2{chapters\ref{#1}--\ref{#2}}
\def\algref#1{algorithm~\ref{#1}}
\def\Algref#1{Algorithm~\ref{#1}}
\def\twoalgref#1#2{algorithms \ref{#1} and \ref{#2}}
\def\Twoalgref#1#2{Algorithms \ref{#1} and \ref{#2}}
\def\partref#1{part~\ref{#1}}
\def\Partref#1{Part~\ref{#1}}
\def\twopartref#1#2{parts \ref{#1} and \ref{#2}}

\def\ceil#1{\lceil #1 \rceil}
\def\floor#1{\lfloor #1 \rfloor}
\def\1{\bm{1}}
\newcommand{\train}{\mathcal{D}}
\newcommand{\valid}{\mathcal{D_{\mathrm{valid}}}}
\newcommand{\test}{\mathcal{D_{\mathrm{test}}}}

\def\eps{{\epsilon}}

\def\reta{{\textnormal{$\eta$}}}
\def\ra{{\textnormal{a}}}
\def\rb{{\textnormal{b}}}
\def\rc{{\textnormal{c}}}
\def\rd{{\textnormal{d}}}
\def\re{{\textnormal{e}}}
\def\rf{{\textnormal{f}}}
\def\rg{{\textnormal{g}}}
\def\rh{{\textnormal{h}}}
\def\ri{{\textnormal{i}}}
\def\rj{{\textnormal{j}}}
\def\rk{{\textnormal{k}}}
\def\rn{{\textnormal{n}}}
\def\ro{{\textnormal{o}}}
\def\rp{{\textnormal{p}}}
\def\rq{{\textnormal{q}}}
\def\rr{{\textnormal{r}}}
\def\rs{{\textnormal{s}}}
\def\rt{{\textnormal{t}}}
\def\ru{{\textnormal{u}}}
\def\rv{{\textnormal{v}}}
\def\rw{{\textnormal{w}}}
\def\rx{{\textnormal{x}}}
\def\ry{{\textnormal{y}}}
\def\rz{{\textnormal{z}}}

\def\rvepsilon{{\mathbf{\epsilon}}}
\def\rvtheta{{\mathbf{\theta}}}
\def\rva{{\mathbf{a}}}
\def\rvb{{\mathbf{b}}}
\def\rvc{{\mathbf{c}}}
\def\rvd{{\mathbf{d}}}
\def\rve{{\mathbf{e}}}
\def\rvf{{\mathbf{f}}}
\def\rvg{{\mathbf{g}}}
\def\rvh{{\mathbf{h}}}
\def\rvu{{\mathbf{i}}}
\def\rvj{{\mathbf{j}}}
\def\rvk{{\mathbf{k}}}
\def\rvl{{\mathbf{l}}}
\def\rvm{{\mathbf{m}}}
\def\rvn{{\mathbf{n}}}
\def\rvo{{\mathbf{o}}}
\def\rvp{{\mathbf{p}}}
\def\rvq{{\mathbf{q}}}
\def\rvr{{\mathbf{r}}}
\def\rvs{{\mathbf{s}}}
\def\rvt{{\mathbf{t}}}
\def\rvu{{\mathbf{u}}}
\def\rvv{{\mathbf{v}}}
\def\rvw{{\mathbf{w}}}
\def\rvx{{\mathbf{x}}}
\def\rvy{{\mathbf{y}}}
\def\rvz{{\mathbf{z}}}

\def\erva{{\textnormal{a}}}
\def\ervb{{\textnormal{b}}}
\def\ervc{{\textnormal{c}}}
\def\ervd{{\textnormal{d}}}
\def\erve{{\textnormal{e}}}
\def\ervf{{\textnormal{f}}}
\def\ervg{{\textnormal{g}}}
\def\ervh{{\textnormal{h}}}
\def\ervi{{\textnormal{i}}}
\def\ervj{{\textnormal{j}}}
\def\ervk{{\textnormal{k}}}
\def\ervl{{\textnormal{l}}}
\def\ervm{{\textnormal{m}}}
\def\ervn{{\textnormal{n}}}
\def\ervo{{\textnormal{o}}}
\def\ervp{{\textnormal{p}}}
\def\ervq{{\textnormal{q}}}
\def\ervr{{\textnormal{r}}}
\def\ervs{{\textnormal{s}}}
\def\ervt{{\textnormal{t}}}
\def\ervu{{\textnormal{u}}}
\def\ervv{{\textnormal{v}}}
\def\ervw{{\textnormal{w}}}
\def\ervx{{\textnormal{x}}}
\def\ervy{{\textnormal{y}}}
\def\ervz{{\textnormal{z}}}

\def\rmA{{\mathbf{A}}}
\def\rmB{{\mathbf{B}}}
\def\rmC{{\mathbf{C}}}
\def\rmD{{\mathbf{D}}}
\def\rmE{{\mathbf{E}}}
\def\rmF{{\mathbf{F}}}
\def\rmG{{\mathbf{G}}}
\def\rmH{{\mathbf{H}}}
\def\rmI{{\mathbf{I}}}
\def\rmJ{{\mathbf{J}}}
\def\rmK{{\mathbf{K}}}
\def\rmL{{\mathbf{L}}}
\def\rmM{{\mathbf{M}}}
\def\rmN{{\mathbf{N}}}
\def\rmO{{\mathbf{O}}}
\def\rmP{{\mathbf{P}}}
\def\rmQ{{\mathbf{Q}}}
\def\rmR{{\mathbf{R}}}
\def\rmS{{\mathbf{S}}}
\def\rmT{{\mathbf{T}}}
\def\rmU{{\mathbf{U}}}
\def\rmV{{\mathbf{V}}}
\def\rmW{{\mathbf{W}}}
\def\rmX{{\mathbf{X}}}
\def\rmY{{\mathbf{Y}}}
\def\rmZ{{\mathbf{Z}}}

\def\ermA{{\textnormal{A}}}
\def\ermB{{\textnormal{B}}}
\def\ermC{{\textnormal{C}}}
\def\ermD{{\textnormal{D}}}
\def\ermE{{\textnormal{E}}}
\def\ermF{{\textnormal{F}}}
\def\ermG{{\textnormal{G}}}
\def\ermH{{\textnormal{H}}}
\def\ermI{{\textnormal{I}}}
\def\ermJ{{\textnormal{J}}}
\def\ermK{{\textnormal{K}}}
\def\ermL{{\textnormal{L}}}
\def\ermM{{\textnormal{M}}}
\def\ermN{{\textnormal{N}}}
\def\ermO{{\textnormal{O}}}
\def\ermP{{\textnormal{P}}}
\def\ermQ{{\textnormal{Q}}}
\def\ermR{{\textnormal{R}}}
\def\ermS{{\textnormal{S}}}
\def\ermT{{\textnormal{T}}}
\def\ermU{{\textnormal{U}}}
\def\ermV{{\textnormal{V}}}
\def\ermW{{\textnormal{W}}}
\def\ermX{{\textnormal{X}}}
\def\ermY{{\textnormal{Y}}}
\def\ermZ{{\textnormal{Z}}}

\def\vzero{{\bm{0}}}
\def\vone{{\bm{1}}}
\def\vmu{{\bm{\mu}}}
\def\vtheta{{\bm{\theta}}}
\def\va{{\bm{a}}}
\def\vb{{\bm{b}}}
\def\vc{{\bm{c}}}
\def\vd{{\bm{d}}}
\def\ve{{\bm{e}}}
\def\vf{{\bm{f}}}
\def\vg{{\bm{g}}}
\def\vh{{\bm{h}}}
\def\vi{{\bm{i}}}
\def\vj{{\bm{j}}}
\def\vk{{\bm{k}}}
\def\vl{{\bm{l}}}
\def\vm{{\bm{m}}}
\def\vn{{\bm{n}}}
\def\vo{{\bm{o}}}
\def\vp{{\bm{p}}}
\def\vq{{\bm{q}}}
\def\vr{{\bm{r}}}
\def\vs{{\bm{s}}}
\def\vt{{\bm{t}}}
\def\vu{{\bm{u}}}
\def\vv{{\bm{v}}}
\def\vw{{\bm{w}}}
\def\vx{{\bm{x}}}
\def\vy{{\bm{y}}}
\def\vz{{\bm{z}}}

\def\evalpha{{\alpha}}
\def\evbeta{{\beta}}
\def\evepsilon{{\epsilon}}
\def\evlambda{{\lambda}}
\def\evomega{{\omega}}
\def\evmu{{\mu}}
\def\evpsi{{\psi}}
\def\evsigma{{\sigma}}
\def\evtheta{{\theta}}
\def\eva{{a}}
\def\evb{{b}}
\def\evc{{c}}
\def\evd{{d}}
\def\eve{{e}}
\def\evf{{f}}
\def\evg{{g}}
\def\evh{{h}}
\def\evi{{i}}
\def\evj{{j}}
\def\evk{{k}}
\def\evl{{l}}
\def\evm{{m}}
\def\evn{{n}}
\def\evo{{o}}
\def\evp{{p}}
\def\evq{{q}}
\def\evr{{r}}
\def\evs{{s}}
\def\evt{{t}}
\def\evu{{u}}
\def\evv{{v}}
\def\evw{{w}}
\def\evx{{x}}
\def\evy{{y}}
\def\evz{{z}}

\def\mA{{\bm{A}}}
\def\mB{{\bm{B}}}
\def\mC{{\bm{C}}}
\def\mD{{\bm{D}}}
\def\mE{{\bm{E}}}
\def\mF{{\bm{F}}}
\def\mG{{\bm{G}}}
\def\mH{{\bm{H}}}
\def\mI{{\bm{I}}}
\def\mJ{{\bm{J}}}
\def\mK{{\bm{K}}}
\def\mL{{\bm{L}}}
\def\mM{{\bm{M}}}
\def\mN{{\bm{N}}}
\def\mO{{\bm{O}}}
\def\mP{{\bm{P}}}
\def\mQ{{\bm{Q}}}
\def\mR{{\bm{R}}}
\def\mS{{\bm{S}}}
\def\mT{{\bm{T}}}
\def\mU{{\bm{U}}}
\def\mV{{\bm{V}}}
\def\mW{{\bm{W}}}
\def\mX{{\bm{X}}}
\def\mY{{\bm{Y}}}
\def\mZ{{\bm{Z}}}
\def\mBeta{{\bm{\beta}}}
\def\mPhi{{\bm{\Phi}}}
\def\mLambda{{\bm{\Lambda}}}
\def\mSigma{{\bm{\Sigma}}}

\newcommand{\tens}[1]{\bm{\mathsfit{#1}}}
\def\tA{{\tens{A}}}
\def\tB{{\tens{B}}}
\def\tC{{\tens{C}}}
\def\tD{{\tens{D}}}
\def\tE{{\tens{E}}}
\def\tF{{\tens{F}}}
\def\tG{{\tens{G}}}
\def\tH{{\tens{H}}}
\def\tI{{\tens{I}}}
\def\tJ{{\tens{J}}}
\def\tK{{\tens{K}}}
\def\tL{{\tens{L}}}
\def\tM{{\tens{M}}}
\def\tN{{\tens{N}}}
\def\tO{{\tens{O}}}
\def\tP{{\tens{P}}}
\def\tQ{{\tens{Q}}}
\def\tR{{\tens{R}}}
\def\tS{{\tens{S}}}
\def\tT{{\tens{T}}}
\def\tU{{\tens{U}}}
\def\tV{{\tens{V}}}
\def\tW{{\tens{W}}}
\def\tX{{\tens{X}}}
\def\tY{{\tens{Y}}}
\def\tZ{{\tens{Z}}}

\def\gA{{\mathcal{A}}}
\def\gB{{\mathcal{B}}}
\def\gC{{\mathcal{C}}}
\def\gD{{\mathcal{D}}}
\def\gE{{\mathcal{E}}}
\def\gF{{\mathcal{F}}}
\def\gG{{\mathcal{G}}}
\def\gH{{\mathcal{H}}}
\def\gI{{\mathcal{I}}}
\def\gJ{{\mathcal{J}}}
\def\gK{{\mathcal{K}}}
\def\gL{{\mathcal{L}}}
\def\gM{{\mathcal{M}}}
\def\gN{{\mathcal{N}}}
\def\gO{{\mathcal{O}}}
\def\gP{{\mathcal{P}}}
\def\gQ{{\mathcal{Q}}}
\def\gR{{\mathcal{R}}}
\def\gS{{\mathcal{S}}}
\def\gT{{\mathcal{T}}}
\def\gU{{\mathcal{U}}}
\def\gV{{\mathcal{V}}}
\def\gW{{\mathcal{W}}}
\def\gX{{\mathcal{X}}}
\def\gY{{\mathcal{Y}}}
\def\gZ{{\mathcal{Z}}}

\def\sA{{\mathbb{A}}}
\def\sB{{\mathbb{B}}}
\def\sC{{\mathbb{C}}}
\def\sD{{\mathbb{D}}}
\def\sF{{\mathbb{F}}}
\def\sG{{\mathbb{G}}}
\def\sH{{\mathbb{H}}}
\def\sI{{\mathbb{I}}}
\def\sJ{{\mathbb{J}}}
\def\sK{{\mathbb{K}}}
\def\sL{{\mathbb{L}}}
\def\sM{{\mathbb{M}}}
\def\sN{{\mathbb{N}}}
\def\sO{{\mathbb{O}}}
\def\sP{{\mathbb{P}}}
\def\sQ{{\mathbb{Q}}}
\def\sR{{\mathbb{R}}}
\def\sS{{\mathbb{S}}}
\def\sT{{\mathbb{T}}}
\def\sU{{\mathbb{U}}}
\def\sV{{\mathbb{V}}}
\def\sW{{\mathbb{W}}}
\def\sX{{\mathbb{X}}}
\def\sY{{\mathbb{Y}}}
\def\sZ{{\mathbb{Z}}}

\def\emLambda{{\Lambda}}
\def\emA{{A}}
\def\emB{{B}}
\def\emC{{C}}
\def\emD{{D}}
\def\emE{{E}}
\def\emF{{F}}
\def\emG{{G}}
\def\emH{{H}}
\def\emI{{I}}
\def\emJ{{J}}
\def\emK{{K}}
\def\emL{{L}}
\def\emM{{M}}
\def\emN{{N}}
\def\emO{{O}}
\def\emP{{P}}
\def\emQ{{Q}}
\def\emR{{R}}
\def\emS{{S}}
\def\emT{{T}}
\def\emU{{U}}
\def\emV{{V}}
\def\emW{{W}}
\def\emX{{X}}
\def\emY{{Y}}
\def\emZ{{Z}}
\def\emSigma{{\Sigma}}

\newcommand{\etens}[1]{\mathsfit{#1}}
\def\etLambda{{\etens{\Lambda}}}
\def\etA{{\etens{A}}}
\def\etB{{\etens{B}}}
\def\etC{{\etens{C}}}
\def\etD{{\etens{D}}}
\def\etE{{\etens{E}}}
\def\etF{{\etens{F}}}
\def\etG{{\etens{G}}}
\def\etH{{\etens{H}}}
\def\etI{{\etens{I}}}
\def\etJ{{\etens{J}}}
\def\etK{{\etens{K}}}
\def\etL{{\etens{L}}}
\def\etM{{\etens{M}}}
\def\etN{{\etens{N}}}
\def\etO{{\etens{O}}}
\def\etP{{\etens{P}}}
\def\etQ{{\etens{Q}}}
\def\etR{{\etens{R}}}
\def\etS{{\etens{S}}}
\def\etT{{\etens{T}}}
\def\etU{{\etens{U}}}
\def\etV{{\etens{V}}}
\def\etW{{\etens{W}}}
\def\etX{{\etens{X}}}
\def\etY{{\etens{Y}}}
\def\etZ{{\etens{Z}}}

\newcommand{\pdata}{p_{\rm{data}}}
\newcommand{\ptrain}{\hat{p}_{\rm{data}}}
\newcommand{\Ptrain}{\hat{P}_{\rm{data}}}
\newcommand{\hatpi}{\hat{\pi}}
\newcommand{\hatrhob}{\rho_{\hat{\pi}_b}}
\newcommand{\pmodel}{p_{\rm{model}}}
\newcommand{\Pmodel}{P_{\rm{model}}}
\newcommand{\ptildemodel}{\tilde{p}_{\rm{model}}}
\newcommand{\pencode}{p_{\rm{encoder}}}
\newcommand{\pdecode}{p_{\rm{decoder}}}
\newcommand{\precons}{p_{\rm{reconstruct}}}

\newcommand{\laplace}{\mathrm{Laplace}} 

\newcommand{\E}{\mathbb{E}}
\newcommand{\Ls}{\mathcal{L}}
\newcommand{\R}{\mathbb{R}}
\newcommand{\emp}{\tilde{p}}
\newcommand{\lr}{\alpha}
\newcommand{\reg}{\lambda}
\newcommand{\rect}{\mathrm{rectifier}}
\newcommand{\softmax}{\mathrm{softmax}}
\newcommand{\sigmoid}{\sigma}
\newcommand{\softplus}{\zeta}
\newcommand{\KL}{D_{\mathrm{KL}}}
\newcommand{\TV}{D_{\mathrm{TV}}}
\newcommand{\Var}{\mathrm{Var}}
\newcommand{\standarderror}{\mathrm{SE}}
\newcommand{\Cov}{\mathrm{Cov}}
\newcommand{\normlzero}{L^0}
\newcommand{\normlone}{L^1}
\newcommand{\normltwo}{L^2}
\newcommand{\normlp}{L^p}
\newcommand{\normmax}{L^\infty}

\newcommand{\parents}{Pa} 

\let\ab\allowbreak

\title{What Matters in Building Vision-Language-Action Models for Generalist Robots}

\author{Xinghang Li$^{1,2,6}$, Peiyan Li$^{2,3}$, Long Qian$^{6}$, Minghuan Liu$^{2,4}$, Dong Wang$^{1,2}$, Jirong Liu$^{2,4}$,\\ Bingyi Kang$^{2}$, Xiao Ma$^{2}$, Xinlong Wang$^{6}$, Di Guo$^7$, Tao Kong$^{2,\textrm{\Letter}}$, Hanbo Zhang$^{5,\textrm{\Letter}}$, Huaping Liu$^{1,\textrm{\Letter}}$  \\
$^1$Department of Computer Science and Technology, Tsinghua University,\\ $^2$ByteDance Research, $^3$CASIA MAIS-NLPR,\\
$^4$Shanghai Jiao Tong University, $^5$National University of Singapore\\
$^6$Beijing Academy of Artificial Intelligence, $^7$Beijing University of Posts and Telecommunications\\
\texttt{taokongcn@gmail.com}, \texttt{zhanghb@comp.nus.edu.sg},\\
\texttt{hpliu@tsinghua.edu.cn}\\
}

\maketitle

\begin{abstract}
To utilize Foundation Vision Language Models (VLMs) for robotic tasks and motion planning, the community has proposed different methods for injecting action components into VLMs and building the Vision-Language-Action models (VLAs). 
In this work, we disclose the key factors that significantly influence the performance of VLA \minghuan{on robot manipulation problems} and focus on answering three essential design choices: which backbone to select, how to formulate the VLA architectures, and when to add cross-embodiment data. The obtained results convince us firmly to explain why we prefer VLA and develop a new family of VLAs, RoboVLMs, which require very few manual designs and achieve a new state-of-the-art performance in three simulation tasks and real-world experiments. Through our extensive experiments, which include over 8 VLM backbones, 4 policy architectures, and over 600 distinct designed experiments, we provide a detailed guidebook for the future design of VLAs. In addition to the study, the highly flexible RoboVLMs framework, which supports easy integrations of new VLMs and free combinations of various design choices, is made public to facilitate future research. We open-source all details, including codes, models, datasets, and toolkits, along with detailed training and evaluation recipes at: \url{robovlms.github.io}.
\end{abstract}

\begin{IEEEkeywords}
Robot Foundation Models, Vision-Language-Action Models, Generalist Robot Policies
\end{IEEEkeywords}

\begin{figure}[htbp]
    \centering
    \includegraphics[width=0.9\linewidth]{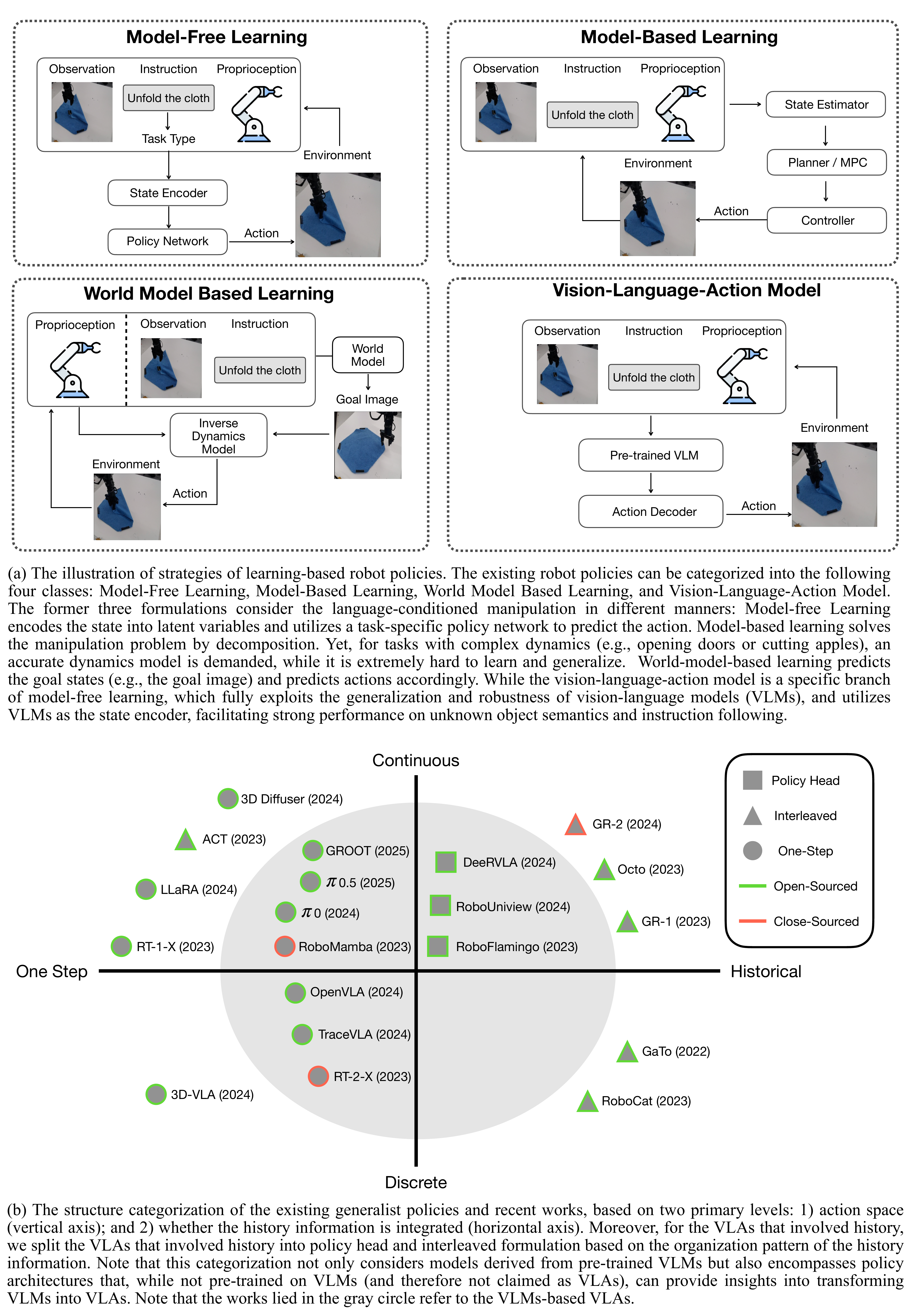}

    \caption{The strategies of learning-based robot policies and the categorization and the recent generalist policies.}
    \label{fig:exist_works}
\end{figure}

\begin{figure}[htbp]
    \centering
    \includegraphics[width=0.9\textwidth]{imgs/illustrate.pdf}
    \caption{
    \minghuan{This work mainly considers three key ingredients for building VLAs based on VLMs: {\bf How} to formulate the problem modeled by VLAs (\textit{i.e.}, observation horizon, action space, and how to aggregate history knowledge); {\bf Which} backbone to use (\textit{i.e.}, various VLM structures, and the data scaled used for the VLM training); and {\bf When} to use extra data source to improve the VLA performance in the tested scenario (\textit{e.g.}, in domain robot data on different tasks, and cross-embodiment data). With our proposed \textit{RoboVLMs}, we can easily transfer various VLMs into generalist robot policies, while supporting multiple embodiments, scenarios, and tasks.}
    }
    \label{fig:illustration}
\end{figure}

\section{Introduction}
\label{sec:intro}

Building generalizable robot policies capable of perceiving, reasoning, and interacting with the physical environment given human instructions has been a long-standing challenge in robotics~\citep{bousmalis2023robocat,brohan2023rt,black2024pi_0,o2023open, liu2025embodied}.
\red{Recently, there has been an active exploration of \textit{Vision-Language-Action Models} (VLAs), namely, a family of learning-based robot foundation models directly built upon generalist \textit{Vision-Language Models} (VLMs),
and they have demonstrated promising results of semantic generality} in both simulated and real-world tasks~\citep{brohan2023rt,kim2024openvla,li2023vision} (Although the rigorous definition of VLAs is not consistent in different works, we regard the ones directly built upon large-scale VLMs along with robot-domain data as the key factor to identify VLAs in this work.).
\textcolor{black}{As shown in Fig. 1a, existing robot policies can be categorized into four main groups:
(1) \textbf{Model-Free Learning}, which encode the state into a latent representation and use a policy network to predict actions, enabling generalization across different embodiments;
(2) \textbf{Model-Based Learning}, which rely on the robot's affordances and explicit models of the robot and environment, limiting their applicability to specific configurations;
(3) \textbf{World Model Based Learning}, which predicts future goal images and derives actions via inverse dynamics models; and
(4) \textbf{Vision-Language-Action (VLA) Model}, representing a specialized branch of model-free learning that leverages the generalization and robustness of Vision-Language Models (VLMs) by employing them as state encoders.}
Therefore, a natural question arises: \textbf{Why} do we prefer \textbf{VLA}s built upon large-scale pre-trained VLMs?
As shown in Fig. 1a, VLA is a newly raised model-free learning framework that receives natural language instruction as input and learns a generalist policy, rather than task-specific policies in traditional model-free learning. The widely-believed reason is that VLAs may inherit the \red{power} of VLMs, which have been shown powerful in learning generalized and robust representations of multi-modal data, such as text, images/videos, through extensive training on web-scale data. 
Such capabilities can facilitate the adaptation of robot foundation models to \gray{bridge the gap between} highly diverse open-world scenes \red{with} limited robotic data. 
\gray{The VLAs built upon VLMs could exhibit strong robustness and generalization over novel scenarios.}
\minghuan{However, it remains an open problem how large-scale vision-language pre-training facilitates generalist robot policies.}

While VLAs have shown early promise, effectively transferring pre-trained VLMs into high-performing robot policies remains non-trivial. Challenges still remain in how to sufficiently and efficiently select and utilize: 1) vision-language backbones; 2) formulation that best leverages multi-modal representations; and 3) the increasing robot data. As shown in Fig.~\ref{fig:illustration}, to address these challenges, we focus on three critical questions: 

\red{\textbf{1. Which kind of VLM backbones are suitable for VLAs?}}
\red{Selecting an appropriate VLM backbone is a fundamental yet underexplored design choice in adapting vision-language models for robotic control.}
\red{Modern VLMs vary widely in architecture. They differ in model capacity, visual encoder structure, fusion mechanisms, and data scale.}
\red{While these models are originally trained for general vision-language tasks, their effectiveness in policy learning depends on how well they can encode physical object properties, spatial relationships, and task-relevant semantics.}
\red{Existing works are based on diversified VLM structures, and most of them show promising results on adaptation to diverse and complicated manipulation skills.}
\red{Yet, there is no comprehensive study on how backbone choices impact downstream robot manipulation performance, and hence, it remains an open problem: Which kind of VLM backbones facilitates the VLA learning most.}

\red{\textbf{2. How should VLAs be formulated to best leverage VLMs?}} Beyond the diversity of different backbones, for generalist robot policies, including VLAs, the structures are more complex and vary in form. Based on the most prevalent existing work~\citep{li2023vision, team2024octo, wu2023unleashing, brohan2023rt, kim2024openvla, o2023open, nair2022r3m, jiang2022vima, zhen20243d, black2024pi_0}, we propose a categorization based on 1) how the history and action information are incorporated in VLAs and 2) whether the action space is continuous or discrete.
As shown in Fig. 1b, four types of structure formulations are considered. For history information modeling, two forms are identified: 1) \emph{one-step modeling}, which utilizes only the current state or observation to produce actions; and 2) \emph{history modeling}, which processes a sliding window of historical states or observations. Regarding the aggregation of history information, we classify it into two approaches: a) \emph{interleaved modeling}, which integrates historical observation and action sequences in an interleaved format; and b) \emph{policy head}, which separately processes each historical step and fuses the information at a distinct policy head for action prediction.
\textcolor{black}{These above-mentioned formulations leverage the pre-trained VLMs in their different ways}. Hence, they may have different features in terms of robustness, generalization ability, and data efficiency when faced with \textcolor{black}{various} types of environments and tasks. Therefore, it is practically important but underexplored to understand: How should we formulate VLAs to sufficiently leverage the power of VLMs in practice?

\red{\textbf{3. When should we leverage the public diverse robot data for optimal training?}} In addition to the VLA itself, the quality and diversity of the training data used to develop VLAs are equally critical. With recent progress achieved by well-known VLAs~\citep{team2024octo, brohan2023rt, kim2024openvla, o2023open, black2024pi_0}, large-scale data from different embodiments (also known as cross-embodiment data) is important to further improve performance in terms of robustness and generalization against out-of-distribution tasks and environments. Yet, \minghuan{they differ largely in detailed training recipes: some utilize cross-embodiment data to 1) pre-train a foundation VLA from VLMs by refining representations closer to robotic manipulation tasks~\citep{black2024pi_0, zhou2025vision} and 2) post-train the VLA with in-domain data; while others co-train VLAs with cross-embodiment data alongside in-domain data~\citep{team2024octo,brohan2023rt,kim2024openvla,o2023open}. Moreover, by being sufficiently pre-trained on cross-embodiment datasets, VLAs are expected to learn new skills with minimal demonstrations~\citep{finn2017model}.}
Consequently, in the case of developing efficient VLAs, when to leverage the large-scale cross-embodiment data becomes an intriguing issue.

To thoroughly study the aforementioned key questions,  
we propose a general framework, {\bf RoboVLMs}, to easily transfer the VLMs into VLAs and implement a fair comparison. We evaluate these models on two popular robot manipulation benchmarks in simulation: \textbf{CALVIN}~\citep{mees2022calvin} and \textbf{SimplerEnv}~\citep{radosavovic2023real}.
Moreover, we also trained and evaluated the built VLAs on a real-world robot manipulation dataset, consisting of 100 manipulation tasks and a total of 74K trajectories.
Specifically, we initially selected three commonly used VLMs--LLaVA, Flamingo, and KosMos, as backbones, combining each with the four VLA structures illustrated in Fig. 1b to examine the effects of action space, observation horizon, and history aggregating methods.
With the finding that the policy head modeling with continuous action space performs best, we compare 8 various VLMs as the backbone with policy head formulation to answer which backbone is more suitable. Meanwhile, we compare the generalization and data efficiency of different VLA structures. For the question of when to leverage cross-embodiment data, we compare co-training (the VLAs trained with Open X-Embodiment), finetuning (the VLAs trained with the target dataset), and post-training (the VLAs co-trained with Open X-Embodiment and further finetuned with the target dataset). Finally, to confirm the real-world applicability of the VLAs with the optimal configuration, we trained and evaluated them in real-world robot manipulation scenarios, demonstrating generalization across 1) unseen distractors, 2) unseen backgrounds, 3) unseen target objects, and 4) novel skill descriptions.

Through our extensive and comprehensive studies, we derive conclusions on how to build high-performance VLAs around the following questions:

\noindent{\bf Why do we prefer VLAs?} 
VLAs built upon pre-trained VLMs have proven to be both effective and efficient for generalist robot policies. Across all experiments, including simulations and real-world manipulation tasks, our VLA consistently outperforms open-source state-of-the-art VLAs by a significant margin. Furthermore, pre-trained VLMs exhibit notable advantages in generalization and data efficiency, making them highly desirable for real-world robotic applications.

\noindent{\bf Which VLM backbone is more suitable for VLAs?}
Our extensive study on 8 different VLM backbones shows two distinguished VLM backbones, namely KosMos~\citep{peng2023kosmos} and Paligemma~\citep{beyer2024paligemma}, which significantly outperform the others. These results highlight that comprehensive vision-language pretraining is essential for achieving superior VLA performance.

\noindent{\bf How should we formulate VLAs?}
Through extensive study and experiments, continuous actions consistently outperform auto-regressive discrete actions, while incorporating historical context is crucial for enhancing performance and addressing partial observability.
For the model architecture, Vision-Language Models (VLMs) integrated directly with policy heads demonstrate superior performance compared to other formulations due to the consistent usage, i.e., vision-language tokens should be processed in their original pretraining format, with a policy head added to integrate past visual and proprioceptive observations for effective decision-making.
Finally, larger VLMs further enhance efficiency, requiring fewer data to achieve higher performance.

\noindent{\bf When should we leverage cross-embodiment datasets?} 
While it is widely believed that pre-training or post-training with cross-embodiment data improves performance, this belief has not been rigorously validated. Our findings reveal that pre-training with cross-embodiment data does not consistently yield significant improvements in final performance. However, post-training a cross-embodiment pre-trained model on the target dataset can lead to notable performance gains. Additionally, leveraging manipulation data from the same robots or tasks provides a clear boost in performance.

\gray{Throughout the study, we propose a new framework, {\bf RoboVLMs}, which transfers VLMs into VLAs, and provides a unified, flexible, easy-to-use, open-source framework that enables seamless integration of any VLM into VLAs with minimal effort, allowing robotics practitioners to investigate, compare, and deploy future VLAs. Further, the VLAs built by RoboVLMs demonstrate strong performance in generalization, dexterity, and flexibility across a wide range of benchmarks and real-world tasks.}
\gray{We open-source the code, along with model weights, and comprehensive guidelines to facilitate the reproducibility of all the results. Our goal is to shed light on the robotics community and help build generalized robots.}

\section{Main Results and Findings}
\label{sec:resutls}

\gray{Vision-language-action models (VLAs) are commonly defined as models fine-tuned from pre-trained large-scale vision-language models (VLMs) using imitation learning~\citep{li2023vision,brohan2023rt}. By leveraging the robust vision-language representation capabilities of VLMs, VLAs offer a promising approach for developing generalist robotic policies capable of handling complex tasks. However, this approach is not universally accepted as the sole or optimal solution. For instance, modular approaches utilize pre-trained vision and language modules to encode latent representations of multi-modal inputs~\citep{mees2022matters,brohan2022rt}, while alternative methods rely on direct training with diverse robotic datasets~\citep{team2024octo}. Even within VLA research, there is no consensus on architectures or training recipes~\citep{li2023vision,cheang2024gr,kim2024openvla, brohan2023rt}.}

The primary objective of this work is to establish VLAs as robust generalist robotic policies by thoroughly analyzing contemporary VLA architectures and identifying the key factors driving their performance.

As shown in Extended Table.~\ref{tab:chapter_split}, we aim to answer the four questions (why/which/how/when) described in the former section. The 4 questions are further divided into 6 executable research problems, and successively experiments are designed to answer each.

\begin{figure}[!t]
\centering
\includegraphics[width=0.8\linewidth]{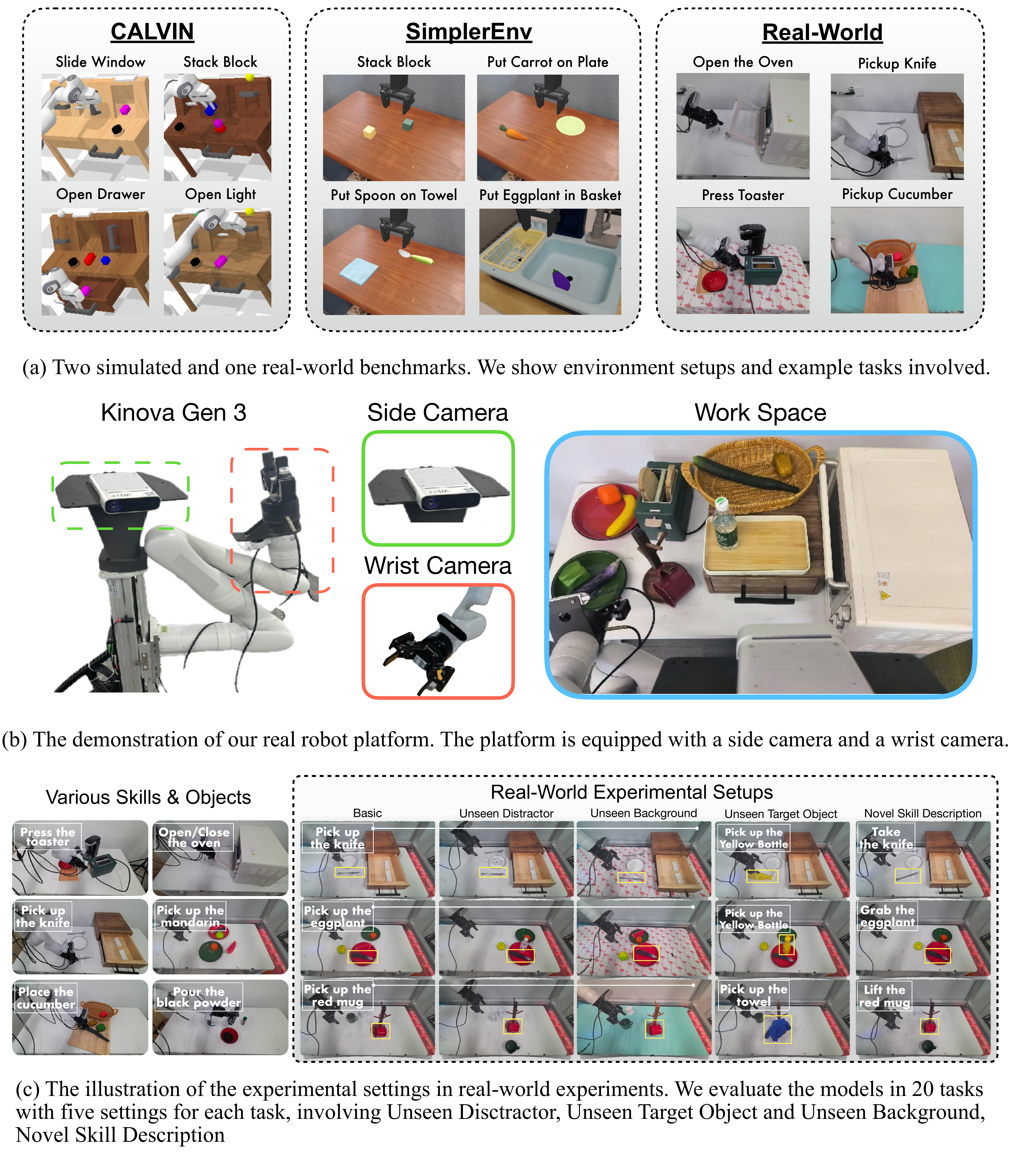}
\caption{Illustration of the involved simulations and real-world benchmarks.}
\label{fig:env_ill}
\end{figure}

\subsection{Benchmarks and Evaluation Metrics}

To comprehensively evaluate the performance of VLAs, in this work, we benchmark all models on a diverse set of benchmarks and robotic manipulation tasks in both simulation and the real world. Specifically, as shown in Fig. 3a, we choose two well-known and widely-used simulation benchmarks ({\bf CALVIN}~\citep{mees2022calvin} and {\bf SimplerEnv}~\citep{torne2024reconciling}) as our testbed. Moreover, we deploy and test our system on a real-world robot manipulation benchmark, including more than 100 tasks, to evaluate RoboVLMs.

\textbf{CALVIN}~\citep{mees2022calvin} is a simulation benchmark for multitask table-top manipulation.  The dataset contains four splits A, B, C, and D according to different scene settings and provides 34 basic tasks with 24K human teleoperated demonstrations annotated with language instructions in total. Evaluation metrics include the success rates of finishing $1\sim 5$ consecutive tasks, as well as the average number of achieved tasks successfully executed (shortened as \textit{Avg. Len.}).
    
\textbf{SimplerEnv}~\citep{li2024evaluating} is designed as a suite of real-to-sim environments and enables the evaluation of robot policies in simulation. It creates a comparable arena for benchmarking the success rate of robot policies in private real-world settings such as Google Robot~\citep{brohan2022rt,brohan2023rt} and Bridge V2~\citep{walke2023bridgedata}.
Evaluation metrics for SimplerEnv are the success rate of each task, simply counted by the number of successfully finished tasks over all valid trials.

\textbf{Real Robot Benchmark}~\citep{cheang2024gr} consists of over 70K teleoperated human trajectories used to fine-tune robot policies, covering 105 manipulation tasks. To evaluate the performance of models on this benchmark, we adopt the approach outlined in~\citep{li2024gr}, testing each model on one \textit{Simple} setting and four challenging \textit{Unseen} settings. Examples of these settings are shown in Fig. 3c. In total, we evaluate each VLA across 20 tasks, with 5 settings per task and 3 rollouts per setting, reporting the average success rate for each setting. A detailed description of the benchmarks is provided in Appendix.

\gray{All tasks included in these benchmarks are driven by single-arm robots, leading to a 7-DoF action - the 6D pose~\footnote{Represented in Euler angles.} of the gripper and one-dimensional open/close status. Robot observation is accessible from proprioceptive sensory information, visual observation, and language input.}

\subsection{Why do we prefer VLAs?}

\begin{figure}[htbp]
    \centering
    \includegraphics[width=0.85\linewidth]{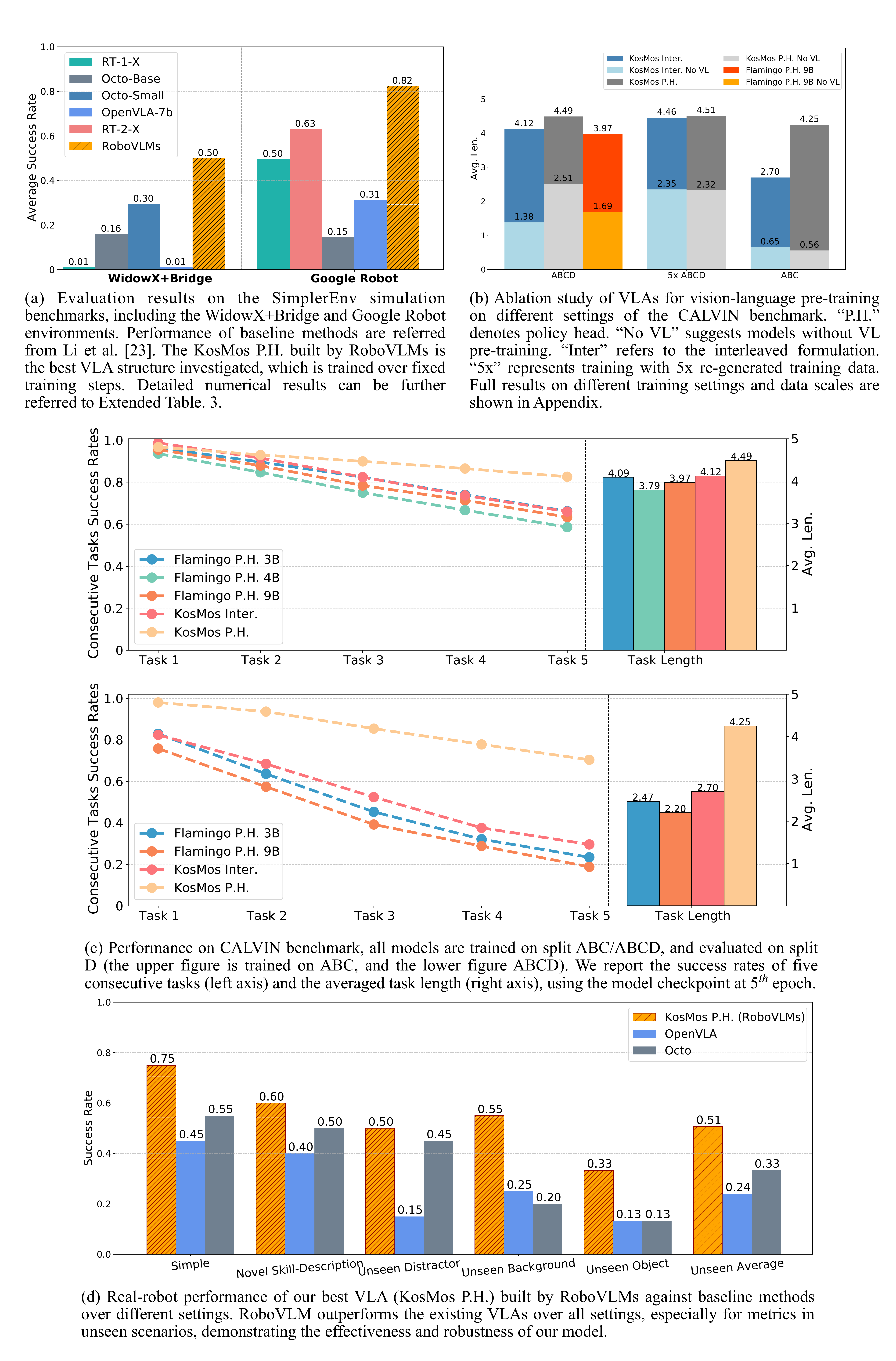}
    
    \caption{The experimental results for RoboVLMs in Simulations and real world.}
    \label{fig:total}  
\end{figure}

This section investigates one of the key questions: \textit{Why do we want VLAs?} To answer this question, we need to answer the following prerequisite sub-question first:
\question{1.1}{Are VLAs a proper choice for building generalist robot policies?}
Concretely, we demonstrate the best-performing VLA resulting from our study, which sets a new state-of-the-art result on both \textbf{CALVIN} and \textbf{SimplerEnv} benchmarks, outperforming all other robot policies with a clear margin.
All results are shown in Extended Table.~\ref{tab:performance_calvin} and Fig. 4a. From these tables, we can see that our strongest RoboVLM exceeds the existing state-of-the-art generalist policies by a large margin and establishes a strong baseline for robot manipulation tasks both in simulation and real-world experiments. Concretely, we can easily observe the following facts:
\begin{itemize}
    \item 
    On \textbf{CALVIN} benchmark, our best model achieves the highest performance in all metrics and demonstrates superior generalization ability when transferring from ABC to D (a novel scene unseen in the training splits) with an absolute improvement of 12.6\% for the execution of a single task and a total improvement of 30.3\% for 5 consecutive tasks. 
    On average, under zero-shot settings, our model can finish 4.25 tasks out of 5 tasks for each single rollout, outperforming the previous SOTA model (GR-1) by 1.19 tasks.
    
    \item 
    On \textbf{SimplerEnv}, our model achieves the highest average performance on both \textit{WidowX + Bridge} and \textit{Google Robot} environments,
    demonstrating the general effectiveness and robustness against different settings and diverse manipulation tasks.

\end{itemize}

We also investigated the impact of vision-language pre-training on the generalization and data efficiency (Fig. 4b and Extended Table.~\ref{tab:data_eff_calvin}).
For the generalization in CALVIN, we adopt the official setting: training models on the split of ABC and validating performance on D. 
To evaluate data efficiency, we conduct experiments on model scales ranging from 3B to 9B and various data scales: 10\% training data (\textbf{0.1x ABCD}), the standard setting (\textbf{ABCD}), and 500\% training data (\textbf{5x ABCD}). The additional data originates from the officially released unlabeled datasets, following the setups as introduced in \citet{wu2023unleashing}. The detailed results on different data scales are shown in Appendix.

We can see that vision-language pre-training is essential for both generalization and data efficiency. This observation is intuitive, as an aligned vision-language representation provides a robust foundation for visual understanding, enabling the policy to focus on learning manipulation skills.
Therefore, we can conclude that
\finding{1.1}{VLA is a promising path to generalist robot policies.}

However, although VLAs perform well in simulation, it is still an open problem whether VLAs are suitable for real-robot applications due to the sim-to-real gap \citep{zhao2020sim}. We propose the second open question:

\question{1.2}{How does the best VLA built with RoboVLMs perform in real-world scenarios?}

As discussed above, we deploy the best-performing RoboVLM model, that is, the one based on the decoder-only KosMos in real-world scenarios to validate its effectiveness.
As shown in Fig. 3c, our experiment involves 20 tasks with multiple skills, including \texttt{Open}, \texttt{Close}, \texttt{Press Button}, \texttt{Pick \& Place}, etc. For each task, we evaluate five settings, with the basic setting, novel skill description, unseen distractors, unseen target object, and unseen background.

Our robot system for real experiments is built on a 7-DoF Kinova Gen3 robot arm paired with a Robotiq 2F-85 gripper. Please refer to Sec.~\ref{sec:method} for more details of the real robot. 
For input, we take the RGB images for the two cameras equipped on the robot head and wrist separately. The head camera provides an overview of the workspace while the gripper camera offers a close observation of the interaction area between the end effector and the environment.

We fine-tune Octo-Base, OpenVLA, and KosMos P.H. built by RoboVLMs on the real robot benchmark and compare their performance. The result is shown in Fig. 4d. We observe that the best VLA (KosMos P.H.) built by RoboVLMs achieves the best performance in all evaluation setups, especially on \textit{Simple} and \textit{Unseen Background}, demonstrating their effectiveness and generalization ability, which is consistent with the results in SimplerEnv and CALVIN simulation.

The qualitative results are shown in Appendix, including success rollouts in various settings and some representative failure cases.
KosMos P.H. not only outperforms baseline models in basic setting tasks like \texttt{Open Drawer}, \texttt{Pickup Eggplant} and \texttt{Press the Toaster Switch}, but also achieves better performance over unseen objects, distractors, and backgrounds. Furthermore, as shown in Extended Fig. 1, KosMos P.H. emerges with self-correction ability, it can realize the incorrect positions of the end effector and correct its future trajectory to complete the task successfully. Note that this ability does not appear in the other tested baselines, and this kind of data is not contained in the training dataset.

\finding{1.2}{The best setup VLA built by RoboVLMs appears to have strong effectiveness and robustness in real scenarios.}

In the following, we will explain our empirical studies and corresponding findings in detail, including the settings, results, and takeaways in building VLAs from pre-trained VLMs.

\subsection{Which kind of VLM backbones are suitable for VLAs?}

When building a VLM-based VLA, the first question is which VLM backbone is the most appropriate one for our VLA:
\question{2}{Which type of VLMs is most suitable for constructing VLAs?}

To thoroughly investigate this question, it would be ideal to conduct experiments in highly controlled settings. However, training VLMs on large-scale vision-language datasets is extremely resource-intensive. Therefore, we base our VLAs on a diverse set of off-the-shelf, pre-trained, large-scale vision-language backbones with varying architectures, training data scales, model sizes, and latent embeddings. These include Flamingo model family~\citep{alayrac2022flamingo} (Encoder-Decoder), and a series of decoder-only VLMs, including LLaVA~\citep{liu2024visual}, Qwen-VL~\citep{bai2023qwen}, MoonDream~\citep{moondream}, UForm~\citep{uform-gen}, Paligemma~\citep{beyer2024paligemma}, and KosMos~\citep{peng2023kosmos}. 
Noticeably, in this section, for fair comparisons, all the models are trained with static images instead of both static and hand cameras.

\gray{Although this approach may not offer a fully controlled comparison, our extensive experiments aim to provide insights into the impact of different VLM backbones on VLAs.}
The results, presented in Extended Table.~\ref{tab:calvin_backbone}, reveals the following observation:
KosMos and Paligemma demonstrate the distinctively better performance. From Extended Table.~\ref{tab:calvin_backbone}, we can see that these two backbones are much better than others with a significantly clear margin. Their superior performance benefits from sufficient vision-language pre-trained on large vision-language datasets. This outcome is intuitive, as extensive pretraining facilitates stronger alignment between visual and linguistic features, an alignment critical for language-conditioned manipulation tasks.

    
\gray{We discuss more influencing factors and interesting findings in Sec.~\ref{sec:discuss}.}
\finding{2}{
VLAs benefit from sufficient vision-language pre-training on larger vision-language datasets of VLMs backbone.
}

\subsection{How should VLAs be formulated to best leverage VLMs?}
\label{sec:how}

In this section, our study addresses questions regarding VLA formulations, including different design choices of VLA structures and various backbones. To answer these questions, we conduct a series of controlled experimental studies to evaluate various VLA formulations on the CALVIN benchmark for rapid evaluation.

\question{3.1}{How to select the best-performing VLA structure?}

\noindent More specifically, how should we model observations, states, and actions in robot manipulation tasks within the context of a VLA? To explore this question, we implement several variants, leveraging various open-source VLM backbones such as OpenFlamingo~\citep{o2023open}, LLaVA~\citep{liu2024visual}, and KosMos~\citep{peng2023kosmos}. These variants incorporate different historical information modeling strategies, and action spaces, as discussed and categorized in Sec.\ref{sec:intro}.

\begin{table*}[!t]
\caption{The ablation study on \textbf{CALVIN} benchmark over the effect of action space, history integration, and history organizing format. All variants are trained on split ABCD and tested on split D. ``Disc." is short for discrete and ``Cont." represents continuous action space. 
Note that for VLAs with LLaVA backbone, we utilize a perceiver resampler to downsample its vision tokens to 64 for fair comparison.
Results are reported with models trained maximally within 5 epochs on the ABCD training splits, all structures predict an action chunk with size 10 and only execute the first step action for inference.
}
\centering
\begin{tabularx}{\textwidth}{l|>{\centering\arraybackslash}X>{\centering\arraybackslash}X|>{\centering\arraybackslash}X>{\centering\arraybackslash}X>{\centering\arraybackslash}X>{\centering\arraybackslash}X>{\centering\arraybackslash}X|>{\centering\arraybackslash}X}
\toprule
\multirow{2}{*}{Backbone} & \multirow{2}{*}{Structure} & \multirow{2}{*}{\makecell{Action\\Space}} & \multicolumn{5}{c|}{Consecutive tasks success rates} &  \multirow{2}{*}{\makecell{\textit{Avg.}\\\textit{Len.}}} \\ \cmidrule{4-8}
 &  &  & 1 & 2 & 3 & 4 & 5 &  \\ \midrule
\multirow{4}{*}{LLaVA} & One-Step & Disc. & 0.809 & 0.484 & 0.278 & 0.175 & 0.103 & 1.85 \\
 & One-Step & Cont. & 0.793 & 0.592 & 0.420 & 0.329 & 0.235 & 2.37 \\
 & Interleaved & Cont. & {\bf 0.892} & 0.645 & 0.436 & 0.282 & 0.181 & 2.44\\
 & Policy-Head & Cont. & 0.873 & {\bf 0.678} & {\bf 0.506} & {\bf 0.376} & {\bf 0.275} & {\bf 2.71} \\ \midrule
\multirow{3}{*}{Flamingo} & One-Step & Disc. & 0.681 & 0.318 & 0.133 & 0.062 & 0.029 & 1.22 \\
 & One-Step & Cont. & 0.681 & 0.354 & 0.158 & 0.076 & 0.035 & 1.30 \\
 & Policy-Head & Cont. & {\bf 0.964} & {\bf 0.896} & {\bf 0.824} & {\bf 0.740} & {\bf 0.662} & {\bf 4.09} \\
\midrule
\multirow{4}{*}{KosMos} & One-Step & Disc. & 0.424 & 0.097 & 0.023 & 0.005 & 0.002 & 0.55 \\
& One-Step & Cont. & 0.935 & 0.868 & 0.814 & 0.768 & 0.702 &  4.09 \\
& Interleaved & Cont. & {\bf 0.987} & 0.915 & 0.824 & 0.737 & 0.660 & 4.12 \\
& Policy-Head & Cont. & 0.967 & {\bf 0.930} & {\bf 0.899} & {\bf 0.865} & {\bf 0.826} & {\bf 4.49} \\ 
\midrule
\multirow{4}{*}{PaliGemma} & One-Step & Disc. & 0.316 & 0.096 & 0.021 & 0.005 & 0.001 & 0.44  \\
& One-Step & Cont. & 0.933 & 0.863 & 0.808 & 0.751 & 0.688 & 4.04   \\
& Interleaved & Cont. & 0.949 & 0.896 & 0.851 & 0.803 & 0.754 & 4.25 \\
& Policy-Head & Cont. & {\bf 0.984} & {\bf 0.933} & {\bf 0.888} & {\bf 0.835} & {\bf 0.779} & {\bf 4.42} \\ 
\bottomrule
\end{tabularx}
\label{tab:alblation_calvin}
\end{table*}

The performance of various VLA structures in CALVIN is summarized in Table.~\ref{tab:alblation_calvin}. From these results, we draw the following key observations:

\begin{itemize}
    \item \textbf{Continuous action matters:} By comparing two types of action spaces, {\bf continuous} and {\bf discrete}, as shown in Table.~\ref{tab:alblation_calvin}, we observe that under the single-frame formulation, continuous action spaces consistently outperform discrete ones, particularly as task horizons increase. This finding is intuitive: continuous actions can represent high-precision floating-point values, whereas discrete actions are limited to indexing action intervals. For long-horizon tasks, the accumulation of compounding errors significantly degrades the performance of discrete actions.
    
    \item \textbf{History observation matters:} As shown in Table.~\ref{tab:alblation_calvin}, under the same VLM structure (either Encoder-Decoder or Decoder-only), models incorporating history observations as input consistently outperform one-step models, achieving substantially higher success rates across all tasks. This improvement holds regardless of the history fusion strategy used. Furthermore, increasing the length of an observable history can enhance performance, albeit at the cost of higher computational overhead.
    
    \item \textbf{Policy head improves history fusion:} Among the formulations utilizing history, the interleaved history formulation performs worse than merging history via an additional policy head. We hypothesize that the policy head preserves the VLM's original vision-language fusion capabilities while effectively integrating historical information. Moreover, the interleaved formulation incurs significantly higher memory and FLOP costs during both training and inference. This suggests that incorporating history with an additional policy head is a more effective and efficient approach for VLAs.

\end{itemize}

\finding{3.1}{The VLA achieves its best performance when using multi-step historical observations as inputs and continuous actions as outputs. For integrating history with continuous action space, the policy head structure performs better.}

However, beyond the performance itself, one of the most important challenges for modern VLAs is achieving generalization to novel objects and environmental settings, which is critical for practical deployment across various robots and scenarios. Conversely, when generalization is insufficient, fine-tuning the policy with a few new demonstrations becomes ideal. Thus, VLAs should inherit the generalization capabilities of VLMs in open-world settings while maintaining high data efficiency when additional in-domain training samples are available. 

Therefore, we further investigate the generalization and data efficiency for different VLAs formulations.

\question{3.2}{How do different formulations affect the generalization and data efficiency for VLAs?}

To address the question, we empirically study and evaluate the generalization and data efficiency of various VLA formulations, aiming to provide practical insights for training high-performing VLAs.
Specifically, we assess the generalization and data efficiency of different VLAs built with RoboVLMs by training models with different architectures and formulations on varying data scales using the CALVIN datasets.
As discussed earlier, we focus on comparing the interleaved and policy head formulations using the OpenFlamingo and KosMos backbones, which have shown strong potential among all configurations. Note that the interleaved formulation can only be paired with a decoder-only structure. The results presented in Fig. 4c and  Extended Table. 4, lead to the following observations:

\begin{itemize}
    \item \textbf{For generalization performance} (Fig. 4c), our best model, based on the KosMos backbone and leveraging a policy head for history fusion, exhibits only a slight performance drop in zero-shot settings. In contrast, other formulations experience significant performance declines. This finding highlights that the model architecture significantly impacts generalization. This conclusion is further supported by results in Fig. 4a, where tasks in the evaluation set are paired with novel instructions, and Fig. 4d, where our best model outperforms others by a large margin across all unseen tasks.

    \item \textbf{For data efficiency,} we observe trends similar to those for generalization. Our best model consistently achieves the highest performance when training data is scaled down, with a notably slower performance decline compared to other formulations. Additionally, comparisons of encoder-decoder VLAs at different scales reveal that larger models tend to be more data efficient.

\end{itemize}

\finding{3.2}{Leveraging policy head for history fusion is the best in terms of generalization and data efficiency.}

\begin{table*}[!t]
    \centering
    \caption{The performance of VLAs with different training objectives and action execution paradigms, and whether to introduce the MoE structure. We utilize PaliGemma, which aligns with the settings of $\pi$ series, as the VLM backbone and implement Fully Connecting (FC) with MSE and Flow Matching (FM) loss. Note that the FC with Flow Matching loss and MoE structure is identity with $\pi_0$, and for fair comparison, we do not consider state as input. For executing paradigms, we verify three different paradigms: {\bf Chunk} represents that the VLAs execute the entire chunk. {\bf First} refers to VLAs execute the first action in the chunk. {\bf Ensemble} refers to ensemble the first action with the corresponding actions in history chunks.}
    \label{tab:combined1}
    
    \begin{subtable}{\textwidth}
        \centering
        \caption{Performance with different training objectives and execution paradigms}
        \begin{tabularx}{\textwidth}{l@{\extracolsep{\fill}}|>{\centering\arraybackslash}X|>{\centering\arraybackslash}X|>{\centering\arraybackslash}X>{\centering\arraybackslash}X>{\centering\arraybackslash}X>{\centering\arraybackslash}X>{\centering\arraybackslash}X|>{\centering\arraybackslash}X} 
\toprule
\multirow{2}{*}{\makecell{Training\\Objective}} & \multirow{2}{*}{\makecell{Training\\Split}} & \multirow{2}{*}{\makecell{Executing\\Paradigm}} & \multicolumn{5}{c|}{Consecutive tasks success rates} & \multirow{2}{*}{\makecell{\textit{Avg.}\\\textit{Len.}}}\\ \cmidrule{4-8} 
 &  &  & 1 & 2 & 3 & 4 & 5 &  \\ \midrule
Flow Matching\ \  & \multirow{6}{*}{ABC}  & Chunk & 0.910 & 0.812 & 0.730 & 0.652 & 0.573 & 3.68 \\
Flow Matching\ \  & & First  & 0.793  & 0.586  & 0.442  & 0.351  & 0.273 & 2.45 \\
Flow Matching\ \  & & Ensemble  & 0.838  & 0.710  & 0.614  & 0.545  & 0.437 & 3.14 \\ 
MSE+BCE  & & Chunk  & 0.901 & 0.802 & 0.703 & 0.620 & 0.544 & 3.57 \\
MSE+BCE & & First  & 0.822  & 0.564  & 0.371  & 0.265  & 0.169 & 2.19 \\
MSE+BCE & & Ensemble  & 0.880  & 0.730  & 0.606  & 0.508  & 0.414 & 3.14 \\ 
\midrule
Flow Matching\ \  & \multirow{6}{*}{ABCD}  & Chunk  & 0.937 & 0.872 &  0.811 & 0.764 & 0.705 & 4.09 \\
Flow Matching\ \ &  & First  & 0.898  & 0.799 & 0.701 & 0.618 & 0.544 & 3.56 \\
Flow Matching\ \ & & Ensemble &  0.945 & 0.882  & 0.818  & 0.770  & 0.708 & 4.12\\
MSE+BCE  & & Chunk  & 0.933 & 0.863 & 0.808 & 0.751 & 0.688 & 4.04 \\
MSE+BCE & & First  & 0.925 & 0.727 & 0.579 &  0.473 & 0.378 & 3.06 \\
MSE+BCE & & Ensemble  &  0.925 & 0.849 & 0.784  & 0.721  & 0.664 & 3.94 \\ 

\bottomrule
\end{tabularx}
\label{tab:train_loss}
    \end{subtable}
    
    \vspace{0.5cm}
    
    \begin{subtable}{\textwidth}
        \centering
        \caption{Effectiveness of Mix-of-Expert structure}
        \begin{tabularx}{\textwidth}{l@{\extracolsep{\fill}}|>{\centering\arraybackslash}X|>{\centering\arraybackslash}X>{\centering\arraybackslash}X>{\centering\arraybackslash}X>{\centering\arraybackslash}X>{\centering\arraybackslash}X|>{\centering\arraybackslash}X} 
\toprule
\multirow{2}{*}{\makecell{Training\\Objective}} & \multirow{2}{*}{\makecell{Using\\MoE}} & \multicolumn{5}{c|}{Consecutive tasks success rates} & \multirow{2}{*}{\makecell{\textit{Avg.}\\\textit{Len.}}}\\ \cmidrule{3-7} 
 &  & 1 & 2 & 3 & 4 & 5 &  \\ \midrule
MSE+BCE & \checkmark (ABC) & 0.932 & 0.824 & 0.732 & 0.644 & 0.556 & 3.69   \\
MSE+BCE & \ding{55} (ABC) & 0.901 & 0.802 & 0.703 & 0.620 & 0.544 & 3.57 \\
Flow Matching \ \ & \checkmark (ABC) & 0.940 & 0.855 & 0.768 & 0.679  & 0.597  & 3.84   \\
Flow Matching \ \  & \ding{55} (ABC) & 0.910 & 0.812 & 0.730 & 0.652 & 0.573 & 3.68 \\ 
\midrule
MSE+BCE & \checkmark (ABCD) & 0.945 & 0.823 & 0.678 & 0.561 & 0.457 & 3.46  \\
MSE+BCE & \ding{55} (ABCD) & 0.933 & 0.863 & 0.808 & 0.751 & 0.688 & 4.04 \\
Flow Matching \ \ & \checkmark (ABCD) & 0.911 & 0.828 & 0.759 & 0.707 & 0.639 & 3.84 \\
Flow Matching \ \  & \ding{55} (ABCD) & 0.937 & 0.872 &  0.811 & 0.764 & 0.705 & 4.10 \\

\bottomrule
\end{tabularx}
\label{tab:moe}
    \end{subtable}
\end{table*}

\red{Yet, the historical formulation is demonstrated to have the best performance, most recent learning-based techniques favor the one-step-continuous formulation for its computational efficiency, with an acceptable performance drop on short-horizon tasks. This offers a practical trade-off, particularly for real-robot deployment~\citep{black2024pi_0, kim2024openvla}. To further investigate their impacts on the performance of resulting policies, we based them on the state-of-the-art one-step-continuous formulation with the best-performing PaliGemma backbone, to align with their original backbone setting and ensure fair comparison.}

\question{3.3}{How do training objectives and inference-time aggregation strategies impact the performance?}

\red{Learning-based policies vary in training objectives and inference, leading to different execution behaviors. For example, in behavior cloning, the most prevailing way is to use MSE or BCE losses for action generation, while diffusion policies learn denoising trajectories through generative modeling. We investigate different training objectives along with the widely-exploited inference-time aggregation strategies for a comprehensive understanding.}

In detail, we compare the following two objectives, which are the most prevailing ones in current days:
\begin{itemize}
    \item \red{{\bf Flow Matching~\cite{lipman2022flow, black2024pi_0}} is a diffusion-based objective. It regards the denoising process as a continuous flow and samples timestep $t\in [0, 1]$ to enforce the model to predict the velocity at $t$. For inference, the model randomly samples an action-chunk shaped gaussion noise, and predicts the denoising velocity based on the current action chunk and time step, the denoised velocity is added to the action chunk vector until the maximum denoising step is reached. }
    \item \red{{\bf MSE+BCE~\cite{wu2023unleashing, li2023vision}}. It formulates the action as an $ n+1$-dimensional vector, with $n$ continuous actions representing arms or end-effector poses, plus a separate gripper action. It utilizes Mean Square Error (MSE) to learn the continuous actions and Binary Cross Entropy (BCE) to learn the state of the gripper.}
\end{itemize}

\red{Besides, we also compare different inference-time action aggregation strategies including \textit{Chunk}, \textit{First}, and \textit{Ensemble}:}
\begin{itemize}
    \item \red{\textbf{Chunk}: A chunk will be fully executed before the next inference.}
    \item \red{\textbf{First}: Only the first action in the chunk will be executed, and then run the next inference.}
    \item \red{\textbf{Ensemble}: An ensemble strategy \cite{zhao2023learning} is used to aggregate the predictions at the same time step from different action chunks.}
\end{itemize}

From Table.~\ref{tab:train_loss} we could observe that:
\begin{itemize}

\item \red{\textbf{Diffusion policies do not significantly improve performance compared to MSE+BCE.} While Flow Matching slightly outperforms MSE+BCE in all experiments, the performance gap is not significant. This suggests that the added inference-time complexity of diffusion models offers limited benefit for short-horizon tasks where deterministic objectives already perform well.}
    
\item \red{\textbf{One-Step-Continuous policies benefit most from executing full action chunks.} Across both Flow Matching and MSE+BCE, the Chunk paradigm consistently yields the best performance in novel scenarios (Training Split ABC). This is likely because the models learn to generate trajectories from a multimodal distribution, and executing only the first action or aggregating across chunks (as in Ensemble) can introduce mode collapse or distort the intended trajectory. Chunk execution preserves temporal coherence within the sampled sequence. While in seen scenarios (Training Split ABCD), executing action chunks does not achieve significant performance improvement compared with ensembling. This indicates that in seen scenarios, the diversity of trajectories might be decreased so that executing action chunks is comparable with ensembling. However, chunking execution could decrease the inference times of VLAs and achieve a real-time deployment (over 30HZ), so that for the rest of the experiments, we utilize chunking execution for one-step-continuous policies.}

\item \red{\textbf{Executing only the first action underperforms across both objectives.} The First paradigm results in significantly lower performance regardless of training objective, especially as the sequential task progresses. This suggests that relying solely on the first action from a sampled or predicted sequence lacks the temporal context and consistency, which are particularly necessary for long-horizon task success.}

\end{itemize}

\finding{3.3}{\red{For \texttt{One-Step-Continuous} formulation, Diffusion loss and MSE loss could achieve a similar performance. For inference-time aggregation strategy, it is important to keep execution consistency, particularly for long-horizon tasks and multimodal actions.}}

\red{Inspired by the Mixture-of-Experts (MoE) in deep learning \citep{moe}, which dynamically routes inputs through specialized feed-forward networks, $\pi_{0}$ \cite{black2024pi_0} proposes augmenting each VLM layer with a dedicated action expert. This design, used in the $\pi_0$ \cite{black2024pi_0} and $\pi_{0.5}$ \cite{intelligence2025pi_}series, preserves vision-language reasoning by routing vision and language tokens through the original VLM modules, while action tokens are handled by the expert and fused via self-attention. Given its demonstrated effectiveness, we also investigate the impact of this structure under our policy formulation.}
\question{3.4}{Does Mix-of-Expert structure benefit VLA learning?}


\red{To assess the effectiveness of the MoE design, we adopt the same setup as the $\pi$-series as shown in Extended Fig. 2b, where the vision-language (VL) tokens are processed by the standard VLM feedforward network (FFN), and the action tokens are routed through a separate action expert FFN. The two sets of tokens interact via self-attention, maintaining joint reasoning while allowing specialization. We evaluate this structure under two representative training objectives: MSE+BCE and Flow Matching, which follow both our default settings in this paper and the original setting in $\pi$-series, respectively.}

From Table.~\ref{tab:moe}, we observe that
\begin{itemize}
    \item \red{MoE structure consistently improves generalization in the visually zero-shot setting (Training Split ABC). For both MSE+BCE and Flow Matching training objectives, using MoE yields higher success rates across all five consecutive tasks. This suggests that MoE enhances the model’s ability to generalize to unseen scenarios. This demonstrates that the mixture-of-experts (MoE) structure is beneficial in preserving the pretrained vision-language (VL) capabilities, as the action prediction does not impose strong supervision on the visual tokens. }

    \item \red{For seen scenarios (Training Split ABCD), introducing the MoE structure does not provide additional benefits. Both training objectives achieve higher success rates without MoE, suggesting that for scenes similar or identical to the training data, the original VLM backbone is sufficient. This may originate from that the MoE structure tends to preserve the general pretrained vision-language (VL) representations, whereas directly fine-tuning the VLA backbone without MoE allows the model to specialize instead of generalize, and fit more effectively to the seen tasks, leading to better performance under familiar settings.}

\end{itemize}

\finding{3.4}{Introducing the Mix-of-Expert structure can improve the generalization of VLAs, while it can not boost the performance in seen scenarios.}

\subsection{When should we leverage the public diverse robot data for optimal training?}

In recent work, it has been a dominant trend to leverage large-scale cross-embodiment robot manipulation datasets to improve the performance of VLAs~\citep{kim2024openvla, black2024pi_0, brohan2023rt, o2023open}. 
However, it is still not fully clear if it helps and an important question remains:
\question{4}{How do large-scale cross-embodiment datasets contribute to VLAs?}

To address this, we break the question into two sub-questions:
\begin{enumerate}
    \item What types of data from large-scale cross-embodiment datasets are the most beneficial for building VLAs?
    \item When and how should these data be utilized effectively?
\end{enumerate}

\textcolor{black}{In this section, we conduct a series of experiments to investigate different strategies for using external large-scale cross-embodiment datasets. Specifically, we explore three settings as shown in Extended Fig.~\ref{fig:cross_setting}}:
\begin{itemize}
    \item \minghuan{\textit{\textbf{Co-train}}: Training the model with in-domain manipulation data and cross-embodiment datasets. Previous work like RT-2~\citep{brohan2023rt}, OpenVLA~\citep{kim2024openvla}, and OCTO~\citep{team2024octo} only consider this one-stage training.}
    
    \item \minghuan{\textit{\textbf{Post-train}}: First, Co-training the VLMs on in-domain and cross-embodiment datasets as above; then, fine-tuning with in-domain manipulation tasks. This has been adopted by $\pi_0$~\citep{black2024pi_0}.}

    \item \minghuan{\textit{\textbf{Finetune}}: Without involving cross-embodiment datasets, fine-tuning the model with only in-domain manipulation tasks. For RT1 dataset that includes the exact target tasks, we further split the RT1 dataset into target tasks and extra in-domain data.}
\end{itemize}

Our experiments in this section use the best-performing KosMos backbone with a policy head for history fusion as the base model. We use Open X-Embodiment (OXE)~\citep{o2023open} as the cross-embodiment dataset, which comprises a diverse range of robot manipulation data collected worldwide and is the most widely used one in recent works~\citep{brohan2023rt, black2024pi_0, kim2024openvla, team2024octo}.

\noindent\textit{D.1 In Domain Evaluation}

We first test on the general in-domain tasks (i.e., tasks inside OXE).
For comparison, we also evaluate a baseline setting, \textit{\textbf{Finetune}}, where the VLA is trained exclusively on in-domain data. 
Additionally, for \textbf{Google Robot}, we include both \textit{RT-Partial \textbf{Finetune}} and \textit{RT \textbf{Finetune}}, where \textit{RT-Partial \textbf{Finetune}} involves only the trajectories with the same task type as the evaluating tasks, and \textit{RT \textbf{Finetune}} involves co-finetuning the policy with additional data from the same robot across different tasks. For \textit{Bridge}, we only evaluate \textit{Bridge \textbf{Finetune}}, which finetunes the policy with the entire \textit{Bridge-V2} dataset, since the training dataset does not contain trajectories with the same instructions as the evaluated tasks.

The comparison of the two stages of using cross-embodiment data is shown in Extended Fig.~\ref{fig:cross_emb}, showing the evaluation results for \textbf{SimplerEnv}-\textit{Google Robot} and \textbf{SimplerEnv}-\textit{Bridge}, from which the following observations can be drawn:
\begin{itemize}
    \item \textbf{\textit{Co-training} on cross-embodiment data without post-training does not help significantly.} Comparing \textit{OXE \textbf{Co-train}} and \textit{RT-Partial \textbf{Finetune}} reveals that for both \textit{Google Robot} and \textit{Bridge}, co-training with cross-embodiment data does not lead to substantial performance improvements. In particular, for \textit{Google Robot}, training with additional in-domain data (\textit{RT \textbf{Finetune}}) from different tasks achieves higher success rates (compared with \textit{RT-Partial \textbf{Finetune}}). This indicates that in-domain data, even if task-agnostic, is more effective for improving model performance than cross-embodiment data.


    \item \red{\textbf{\textit{Post-training} shows potential benefits over high-frequency tasks in the first co-training stage.} The average performance of the post-trained model exceeds that of the model fine-tuned exclusively on in-domain data on \textit{Bridge} (50\% for post-trained and 44\% for fine-tuned). While for \textit{Google Robot}, the post-trained model only outperforms the fine-tuned model over \texttt{pick coke can} tasks, and drops its performance on all other tasks. We consider the main reason is that the proportion of \texttt{pick-and-place}tasks in the OpenX-Embodiment dataset is very large, so that the post-trained model gains higher performance over \texttt{pick-and-place} tasks, while it loses performance over tasks involving other skills like \texttt{move near} and \texttt{open/close drawer} as a trade-off. 
    In other words, cross-embodiment co-training enables VLAs to learn skills for high-frequency tasks from other types of embodiments and transfer the acquired knowledge to the target embodiment during the post-training phase. This helps VLAs further improve the overall success rate of these tasks, though at the cost of sacrificing low-frequency tasks and skills in the co-training stage.}

    \item \red{\textbf{In-domain data plays a critical role in achieving strong task performance.} Even task-agnostic in-domain data from the same embodiment (e.g., RT Finetune) proves more effective than large-scale cross-embodiment data for fine-tuning VLAs, as shown by the experiments on \textit{Google Robot} dataset, especially for fitting specific tasks. While cross-embodiment pre-training can provide useful initialization in some cases, leveraging in-domain data remains essential for robust and consistent policy performance.}
    
\end{itemize}

\noindent\textit{D.2 Out-of-Distribution Evaluation}

To evaluate the impact of cross-embodiment datasets more comprehensively, we further perform experiments on \textbf{CALVIN}, whose data is not part of OXE. We mainly focus on whether cross-embodiment datasets benefit \textit{\textbf{few-shot learning}} (we randomly sample 10 trajectories for each task from data split ABCD for training and test on D) on out-of-distribution tasks (w.r.t. to OXE). To keep settings consistent, we only take the image from the static on-head camera as input. The results are illustrated in Extended Fig.~\ref{fig:few_shot}, and we conclude that: 

\begin{itemize}
    \item \textbf{Cross-embodiment \textit{pre-training} improves \textit{few-shot learning} performance.} In the few-shot setting for CALVIN, with a single-view on-head camera, pre-training significantly improves the performance by 17.2\% in terms of single-task execution and 0.25 more executed tasks in each rollout.
    It can be concluded that pre-training on large-scale cross-embodiment datasets benefits the learning of more effective representations for robot manipulation, which can quickly adapt to novel manipulation tasks with unseen objects and environmental settings.
\end{itemize}

\finding{4}{Extra in-domain data, even from different tasks, shows beneficial, and an extra large-scale cross-embodiment co-training before the post-training stage further improves high-frequency tasks as well as few-shot performance.}

\section{Method and Material}
\label{sec:method}

We consider the problem of controlling a robot to accomplish a set of tasks, based on language instructions $l$, and historical observations $o_{t-H+1:t}$ at every time step $t$ with the maximum historical length $H$.
In this paper, we mainly consider a table-top robot arm, therefore the observations $o_t$ are sensor inputs and images, e.g., $o_t = (s_t, I_t)$, from either a third-view camera, a gripper camera, or both.
We build a controlling policy $\pi(a|o_{t-H+1:t},l)$, where the action $a$ is modeled as a 7-dim vector, including the 6-DoF pose of the gripper, along with its open/close status. 
In this section, we will first introduce the formulation of vision-language models (Section \ref{sec:vlm}), based on which we build our VLA formulations.

\subsection{Vision Language Model}
\label{sec:vlm}
Vision-language models (VLMs), also referred to as multi-modal large language models, integrate vision into their input modality, enabling them to process and reason about both visual and textual information. Typically, VLMs are prompted with images and/or text to generate text~\citep{liu2024visual, bai2023qwen, alayrac2022flamingo, moondream, uform-gen, beyer2024paligemma}, facilitating applications such as image captioning, visual question answering, and goal-directed planning. 
This process can be formally described as:
\begin{equation}
    \hat{l} = \text{VLM}(I, l_{\text{prompt}})~.
\end{equation}
Here, $I$ and $l_{\text{prompt}}$ denote the image and text prompts, respectively, while $\hat{l}$ represents the text output generated by the VLM. For instance, in a visual question-answering task, $l_{\text{prompt}}$ corresponds to the question, and $l_{\text{target}}$ corresponds to the generated answer.
Training a VLM typically involves minimizing a cross-entropy loss to predict discrete language tokens, which can be expressed as:
\begin{equation}
    \ell_{\text{VLM}} = \text{CrossEntropy}(\hat{l}, l_{\text{target}})~,
\end{equation}
where $l_{\text{target}}$ is the ground truth text.
By pre-training on millions or even billions of paired vision-language data, VLMs acquire robust representations of both visual and textual modalities.
To effectively handle these two distinct modalities, VLMs typically employ a vision processor and a language decoder, connected through various vision-language feature fusion mechanisms. Among the available options, vision transformer (ViTs)~\citep{dosovitskiy2020image} and perceiver resampler~\citep{jaegle2021perceiver} are widely adopted choices for the vision processor~\citep{li2023vision, team2024octo, liu2024robomamba, wu2023unleashing}. The ViT module reshapes each input image $I$ into patches and encodes them as visual tokens \texttt{[OBS]}:
\begin{equation}
    \texttt{[OBS]} = (x^v_1, \cdots, x^v_N) = \text{ViT}(I)~,
\end{equation}
where $N$ represents the token number, $x^v_i$ represents the $i$th token, which, in ViT, is an embedding vector encoding a patch of the input image.

Existing VLMs can be broadly categorized into two structural paradigms: encoder-decoder and decoder-only architectures. These two structures differ mainly in their multi-modal fusion strategies.

{\noindent\bf Encoder-decoder}. Encoder-decoder architectures are composed of two main components: 
an encoder that is typically responsible for extracting features from inputs using input embedding modules as discussed above, and a decoder that generates the output (e.g., text or multi-modal predictions) auto-regressive. 
Feature fusion between the encoder and decoder is usually achieved by cross-attention layers in the decoder. 
This structure excels in tasks requiring a detailed understanding of the input modalities, such as image captioning and visual reasoning, due to its ability to explicitly encode multi-modal information prior to generation~\citep{nagrani2021attention, xu2021vlm}. Representative models include Flamingo~\citep{alayrac2022flamingo} and OFA~\citep{wang2022ofa}.

{\noindent\bf Decoder-only}. Decoder-only architectures, in contrast, rely on a unified transformer framework where both the input modalities (vision and text) and the output sequences are processed in the same autoregressive decoder. In these models, visual features are first embedded into token-like representations (via vision processors), which are then concatenated with textual tokens and passed through the decoder. 
Multi-modal feature fusion occurs naturally through the self-attention mechanism, allowing the decoder to model dependencies between visual and textual inputs during token generation. Decoder-only architectures are more flexible and scalable, making them well-suited for tasks like instruction following, multi-modal question answering, and open-ended generation. Examples of decoder-only models include GPT-4V~\citep{yang2023dawn} and LLaVA~\citep{liu2024visual}.

\subsection{Vision-Language-Action Models}
\label{sec:vla}


Vision-language-action models (VLAs) are predominantly applied to robotic tasks, where they serve as a generalist robot policy $\pi$ capable of handling complicated tasks. In formal, VLAs predict action sequences based on previous observations at the current time step $t$:
\begin{equation}\label{eq:vla-general}
    a_{t:t+L-1} = \text{VLA}(o_{t-H+1:t}, l_{\text{prompt}})~,
\end{equation}
where $a_{t:t+L-1}$ are a sequence of predicted 7-dim actions, $L$ is the action sequence length and $H$ is the history observation length.
Different from VLMs, the observations of VLAs $o_{t-H+1:t}$ usually contain proprioceptive states $s_{t-H+1:t}$ like the joint angles and end-effector positions besides the visual inputs $I_{t-H+1:t}$.
As discussed in Section \ref{sec:vla}, we abstract and categorize VLAs into four representative structures based on 1) historical information modeling and 2) action space. Before we describe these distinct models in form, we first introduce the general pre-process and prediction principles when dealing with robot actions in different action spaces.

\noindent\textit{1) Action Pre-process}

\noindent\textbf{Action Normalization.}
For both continuous and discrete action spaces, we normalize each dimension of the 7-DoF action. Following \citet{kim2024openvla}, we count the 1\textsuperscript{st} and 99\textsuperscript{th} quantile of the actions in the training data and use the quantiles to clamp each dimension of the action~\citep{brohan2023rt}:
\begin{equation}
    a^{i'} = \min(a^i_{ 99\textsuperscript{th}}, \max(a^i_{1\textsuperscript{st}}, a^i))~,
\end{equation}
where $a^{i'}$ is the clamped value of the $i$-th dimension of action $a$. For the next step, we normalize each dimension of the clamped action $a^{'}$ with the 1\textsuperscript{st} and 99\textsuperscript{th} quantile of the actions:
\begin{equation}
    \Tilde{a}^i = 2 \times (a^{i'} - a^i_{1\textsuperscript{st}}) / (a^i_{ 99\textsuperscript{th}} - a^i_{1\textsuperscript{st}}) - 1~.
\end{equation}
$\Tilde{a} = [\Tilde{a}^1, \Tilde{a}^2,\cdots, \Tilde{a}^7]$ is the normalized action, each dimension is in the range of $[-1, 1]$, the last dimension representing the open/close status of the gripper $\in \{-1, 1\}$.
During inference time, we will reversely map the predicted actions into unnormalized actions.

\noindent\textbf{Action Discretization.}
For discrete action representation, we need to further discretize the normalized action $\Tilde{a}$. Following \citet{brohan2023rt, kim2024openvla}, we map continuous robot actions to discrete tokens used by the VLM's tokenizer. Specifically, we discretize each robot action dimension into one of 256 bins separately. For each dimension, we set the width of the bin to uniformly divide the interval between the quantile of 1\textsuperscript{st} and 99\textsuperscript{th} of actions in the training data.

Using this discretization, we transform $\Tilde{a}$ to $\overline{a}$ with 7~discrete integers $\in [0 \dots 255]$. To avoid damaging the original special token positions in the language tokenizer, we add an offset (default set to 10) and replace the last offset $\sim$ 256+offset tokens with a discretized index.

\noindent\textit{2) Action Prediction}
\noindent{\bf Continuous Actions.}
We optimize the mean square error (MSE) and binary cross entropy (BCE) for the predicted action sequence $a_{t:t+L-1}$ with the ground truth action sequence $\Tilde{a}_{t:t+L-1}$:
\begin{equation}
    l_{\text{VLA}} = \sum_{i=t}^{t+L-1} \text{MSE}(\hat{a}_{i,pose}, \Tilde{a}_{i, pose}) + \lambda * \text{BCE}(a_{i, gripper}, \Tilde{a}_{i, gripper})~,
\end{equation}
where the MSE loss is computed for the first six dimensions, and the BCE loss is for the last gripper dimension between the predicted action $\hat{a}_{t:t+L-1}$ and the ground truth $\Tilde{a}_{t:t+L-1}$, $\lambda$ is the balancing weight. 

\noindent{\bf Discrete Actions.}
Discrete-action models predict action tokens $\texttt{ACT}^i$ for each action dimension $i$.
These tokens are the index of discretized bins from continuous action by dimension, which can be easily de-tokenized to recover the action vector.
The optimization object has a similar cross-entropy (CE) format as text generation widely used in VLM training:
\begin{equation}
    l_{\text{VLA}} = \sum_{i=t}^{t+L-1} \sum_{j=1}^7 \text{CE}(\texttt{[ACT]}^{j}_{i}, \tilde{a}^j_i)~,
\end{equation}
where $\texttt{[ACT]}^{j}_{i}$ represents the index of bins of the $j$$\textsuperscript{th}$ dimension of the predicted action token $\texttt{[ACT]}$ at time $i$, and $\overline{a}^j_i$ is the corresponding ground truth.
During inference time, after getting the predicted action token $\texttt{ACT}^i$, we re-project the discrete tokens to the center of the corresponding bins into a continuous form for achieving tasks.

\subsection{VLA Structures}
As shown in Extended Fig. 2a, there are mainly four kinds of VLA structures categorized by action space and history aggregating method, namely one-step-continuous-action models, one-step-discrete-action models, interleaved-continuous-action models, and policy-head-continuous-action models. Note that our proposed framework RoboVLMs could transfer VLMs to arbitrary VLA structures with no effort. We will illustrate the four VLA structures in detail in the following section.
\noindent\textit{1) One-step Models}
One-step models predict future action sequences using only the observation at the current time step $t$, i.e., a history length of 1.
\begin{equation}\label{eq:vla-onestep}
    \hat{a}_{t:t+L-1} = \text{VLA}(o_{t}, l_{\text{prompt}})~.
\end{equation}
For one-step models, we formulate two variants: {continuous-action model} and {discrete-action model}:

\noindent \textbf{Continuous-action model}: In the continuous-action formulation, the VLM model first predicts a learnable token \texttt{[LRN]} using the VLM backbone. This is achieved by fusing visual and language tokens (in an encoder-decoder architecture) or concatenating multi-modal tokens (in a decoder-only architecture). An MLP is then used to predict the action vector:
\begin{equation}\label{eq:vla-onestep-cont}
\begin{aligned}
    \texttt{[LRN]} &= \text{VLM}(o_{t}, l_{\text{prompt}})~,\\
    \hat{a}_{t:t+L-1} &= \text{MLP}(\texttt{[LRN]})~.
\end{aligned}
\end{equation}
The one-step continuous-action models include ACT~\citep{zhao2023learning}, BC-Z~\citep{jang2022bc}, MVP~\citep{radosavovic2023real}, R3M~\citep{nair2022r3m}, VIMA~\citep{jiang2022vima}, 3D Diffuser~\citep{ke20243d}, RoboMamba~\citep{liu2024robomamba}, and $\pi_0$~\citep{black2024pi_0}.

\noindent \textbf{Discrete-action model}: For discrete action prediction, we directly follow the straightforward next-word prediction similar to VLMs, where actions are discretized into tokens like texts:
\begin{equation}\label{eq:vla-onestep-disc}
\begin{aligned}
    \texttt{[ACT]}^{1:7}_{t:t+L-1} &= \text{VLM}(o_{t}, l_{\text{prompt}})~.
\end{aligned}
\end{equation}
The one-step discrete-action models include RT-1~\citep{brohan2022rt}, RT-2~\citep{brohan2023rt}, 3D-VLA~\citep{zhen20243d}, LAPA~\citep{ye2024latent}, OpenVLA~\citep{kim2024openvla}, and Embodied-COT~\citep{zawalski2024robotic}.

\noindent\textit{2) Interleaved-Continuous-Action Models}
Interleaved models receive observation-action sequences:
\begin{equation}
    O_t = (\texttt{[OBS]}_{t-H+1}, \texttt{[LRN]}), ..., (\texttt{[OBS]}_{t}, \texttt{[LRN]})~,
\end{equation}
where $O_t$ represents the input token sequence at time instant $t$, $\texttt{[OBS]}$ denotes observation tokens and $\texttt{[LRN]}$ denotes the learnable action token and is  duplicated for $H$ times and insert into $O_t$ with an interleaved format. The VLM backbone fuses this sequence (in a decoder-only structure) and predicts the action sequence through an MLP based on each action token:
\begin{equation}\label{eq:vla-interleaved}
\begin{aligned}
    \texttt{[LRN]}_{t-H+1:t} &= \text{VLM}(O_t)~,\\
    \hat{a}_{t:t+L-1} &= \text{MLP}(\texttt{[LRN]}_{t})~.
\end{aligned}
\end{equation}
The $\texttt{[LRN]}_{t}$ which is utilized to predict the action chunk  $\hat{a}_{t:t+L-1}$, represents the $\texttt{[LRN]}$ inserted after $\texttt{[OBS]}_{t}$ and fused with the observations before $t$.
The loss and action unnormalization procedure is identical with one-step continuous action models. At time instant $t$ of inference, the input sequence contains only the current observation $\texttt{[OBS]}_{t}$ and the language instruction $l_{prompt}$, we add the learnable token $\texttt{[ACT]}$ at the end of the input sequence and pass the sequence to the VLM to predict the action. After the robot executes the predicted action, we add the new observation $\texttt{[OBS]}_{t+1}$ and language instruction $l_{prompt}$ to the input sequence to predict the action in the current step.
The interleaved-continuous-action models include GR-1~\citep{wu2023unleashing}, OCTO~\citep{team2024octo}, GR-2~\citep{cheang2024gr}. Note that the interleaved-discrete-action models like GATO~\citep{reed2022generalist} and RoboCat~\citep{bousmalis2023robocat} are out of consideration. 

\subsubsection{Policy-Head-Continuous-Action Models}
Unlike interleaved models, which fuse historical information within the VLM backbone, policy-head VLAs only require the VLM to provide single-step multi-modal representations at each time step $t$:
\begin{equation}\label{eq:policy_head_obs}
\begin{aligned}
    o_t &= (\texttt{[OBS]}_t, \texttt{[LRN]})~, \\
    \texttt{[LRN]}_t &= \text{VLM}(o_{t}, l_{\text{prompt}})~.
\end{aligned}
\end{equation}
Historical information is then modeled and actions are predicted through an additional policy head $h$, such as an RNN~\citep{medsker2001recurrent, chung2014empirical, graves2012long}, transformer~\citep{vaswani2017attention, floridi2020gpt}, or diffusion model~\citep{chi2023diffusion}:
\begin{equation}\label{eq:vla-policy-head}
\begin{aligned}
     a_{t:t+L-1} &= \text{h}(\texttt{[LRN]}_{t-H+1}, ..., \texttt{[LRN]}_{t})~.
\end{aligned}
\end{equation}
The action chunk $a_{t:t+L-1}$ with sequence length $L$ is predicted based on learnable tokens $\texttt{[LRN]}_{t-H+1}, ..., \texttt{[LRN]}_{t}$.
Each of $\texttt{[LRN]}_{t}$ is identical.
Note that the interleaved-continuous-action model is only available for decoder-only backbones.
The policy-head-continuous-action model can be built based on VLM backbones with both encoder-decoder and decoder-only structures.
The main difference comes from the language decoder. The input sequence of the encoder-decoder VLM fuses only contains the text and learnable action tokens, it fuses the multi-modal input with cross-attention where the text tokens combined with the learnable tokens are the keys and values, and vision tokens are the queries. The decoder-only backbone directly concatenates the vision, language, and learnable tokens as input and utilizes self-attention to fuse the multi-modal features. The policy-head-continuous-action models include RoboFlamingo~\citep{li2023vision}, RoboUniview~\citep{liu2024robouniview}, and DeeRVLA~\citep{yue2024deer}.

At every inference step $t$, the current observation $\texttt{[OBS]}_{t}$ and language instruction $l_{\text{prompt}}$ along with a learnable token $\texttt{[LRN]}$ is concatenated as a complete input sequence, which is further passed into the VLM backbone. After the policy head takes $\texttt{[LRN]}$ and predicts the current action sequences, the robot steps with the predicted actions and obtains the new observation for the next round of prediction. 

\subsection{Real Robot Platform}
As shown in Fig. 3b, our real robot platform consists of a Kinova Gen-3 robot arm, equipped with a Robotiq 2F-85 parallel-jaw gripper and two cameras: one static camera for capturing the workspace and another camera mounted on the end-effector. The static camera is a Kinect Azure, while the wrist-mounted camera is a RealSense D435i. The workspace is a 55 cm x 24 cm table, and there are more than 40 objects distributed across the evaluated scenes.

As described in Section II, we define one simple setting and four unseen settings for evaluation on this platform. For the convenience of readers, we provide a more detailed introduction to these settings here. In the \textit{Simple} setting, the scene is designed to closely resemble those in the training data. This setting is used to assess the model's ability to fit the training distribution. In the \textit{Unseen Distractors} setting, previously unseen distractor objects are introduced into the scene, but the manipulated objects are still part of the training data. The \textit{Unseen Backgrounds} setting changes the background by adding two new tablecloths that were not present in the training data. One tablecloth differs in color from the white background, while the other features entirely different patterns. In the \textit{Unseen Objects} setting, the robot is tasked with manipulating objects that were not included in the training dataset. The unseen objects used in this setting are the same as those in the Unseen Distractors setting. Finally, in the \textit{Novel Skill Description} setting, we use GPT-4 to generate three unseen synonyms for the verbs in the instructions and randomly select one to replace the original verb, creating a novel skill description. For instance, "press" may be replaced with "hit," "pick up" with "take," "close" with "shut," "pour" with "sprinkle," and so on.

\subsection{Discussions about Structures}
Although we have discussed four representative VLA structures, by a two-level categorization of historical information and action space, readers may notice that these two dimensions are not entirely orthogonal, and not all combinations are discussed and implemented in this work. This is due to architectural limitations and implementation challenges.
For instance, interleaved and policy-head models with discrete action spaces have not been implemented so far, because interleaved models are typically combined with action chunk prediction, where the default lower triangular attention mask cannot effectively mask subsequent actions for later steps.

\section{Data availability}
Datasets used in this study could be found at \url{https://github.com/mees/calvin} (CALVIN), \url{https://github.com/google-deepmind/open_x_embodiment} (Open X-Embodiment), and \url{https://huggingface.co/datasets/robovlms/bytedance_robot_benchmark_20} (BDRBench20).

\section{Code availability}
The source code of this study could be found at \url{https://github.com/Robot-VLAs/RoboVLMs}. It is also available at \url{https://zenodo.org/records/17757179}~\citep{li2025robovlms}.

\section{Acknowledgments}
This work was supported in part by the National Natural Science Fund under Grant 62025304 and in part by the National Natural Science Fund for Key International Collaboration under Grant 62120106005. We thank all the members of the robotics research team at ByteDance Research for their assistance in real-world data collection, setup design, robot maintenance, and experiments. The author Minghuan Liu is supported by the ByteDance Scholarship. We also want to thank @YouJiacheng for his active and instructive discussion on X.

\section{Contributions}

\noindent{\textbf{Project Leads}}: Huaping Liu, Xinghang Li, Hanbo Zhang, Minghuan Liu

\noindent{\textbf{Methodology and Codebase}}: Xinghang Li

\noindent{\textbf{Model Training and Evaluation (experimental design, implementation)}}: Xinghang Li, Long Qian, Hanbo Zhang, Dong Wang, Minghuan Liu, Xiao Ma, Jirong Liu

\noindent{\textbf{Real-Robot Deployment \& Experiments}}: Xinghang Li, Peiyan Li

\noindent{\textbf{Paper (logic, figures, visualizations, writing)}}: Xinghang Li, Minghuan Liu, Hanbo Zhang, Long Qian, Bingyi Kang, Xiao Ma, Peiyan Li, Jirong Liu, Di Guo, Huaping Liu, Tao Kong

\noindent{\textbf{Advising}}: Huaping Liu, Tao Kong, Hanbo Zhang, Xiao Ma, Bingyi Kang, Di Guo, Xinlong Wang

\newpage

\begin{extendedfigure*}[t]
  \centering
  \includegraphics[width=1.0\textwidth]{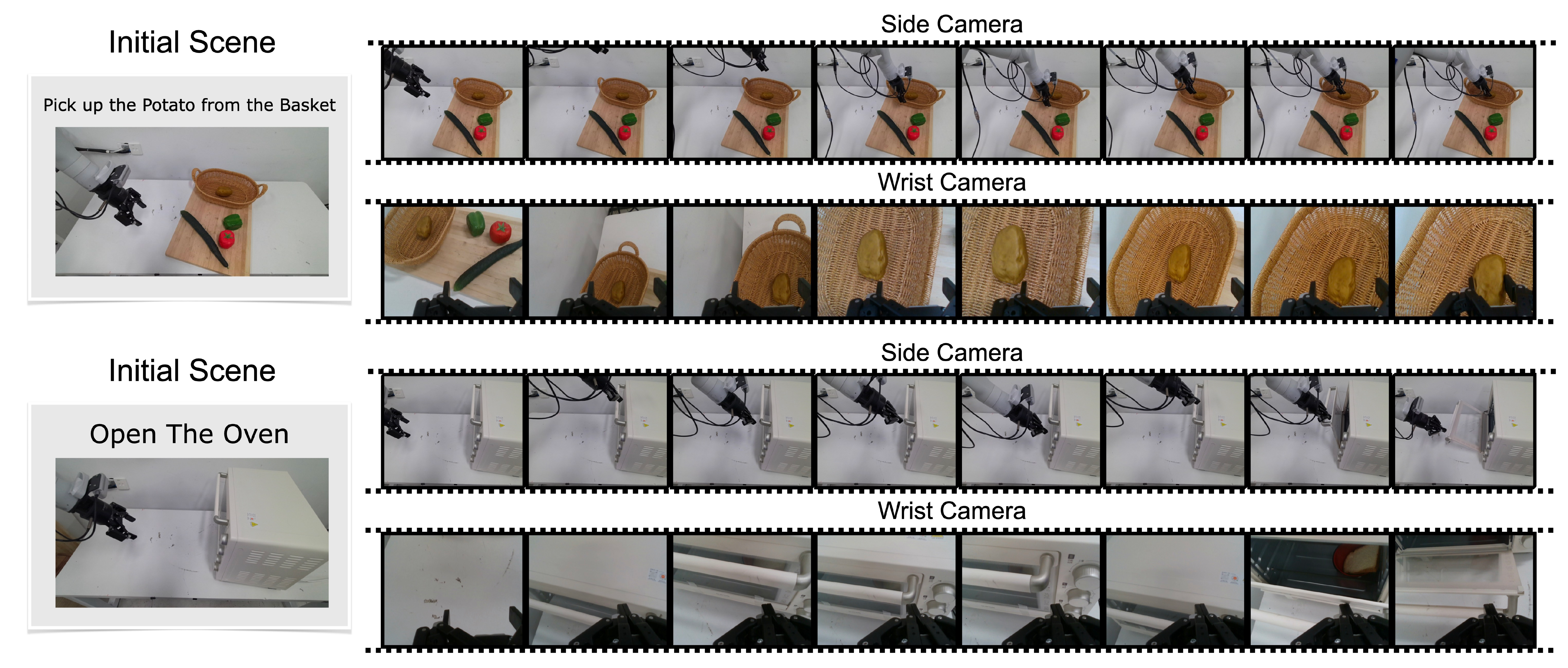}
  \caption{Visualization for rollouts that the best setting VLA built by RoboVLMs emerges the ability of self-correction. For instance, in the \texttt{Open The Oven} task, the robot\'s first attempt does not reach the oven handle, and it adjusts the end-effector position to re-locate the handle at the second attempt. Note that the training dataset does not contain this kind of data.}
  \label{fig:real_correction}
\end{extendedfigure*}

\begin{extendedfigure}  
  \centering
  \includegraphics[width=0.95\linewidth]{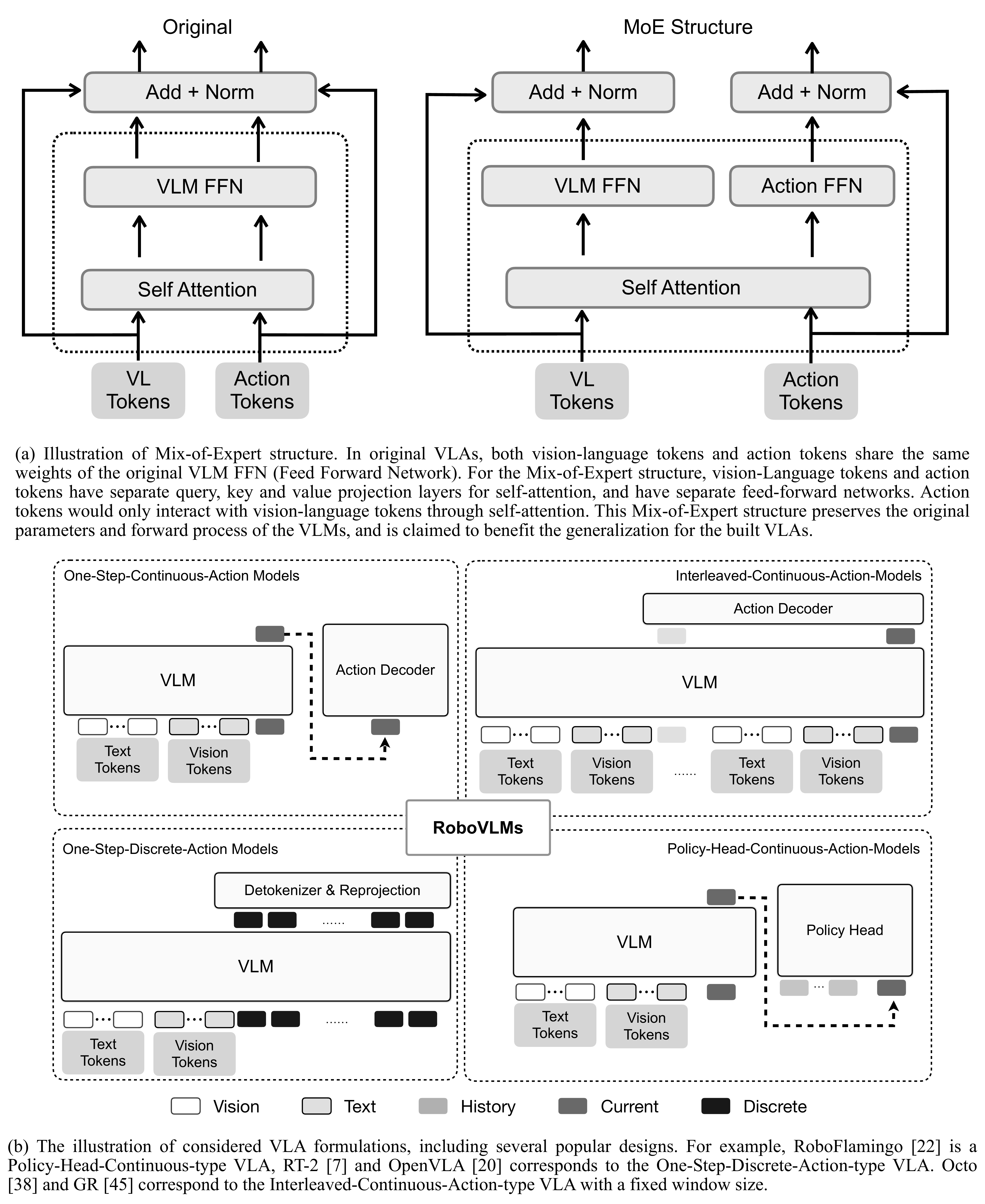}
  \caption{The architectures of MoE and considered VLAs.}
\end{extendedfigure}

\begin{extendedfigure}[htbp]
    \centering
\includegraphics[width=0.8\textwidth]{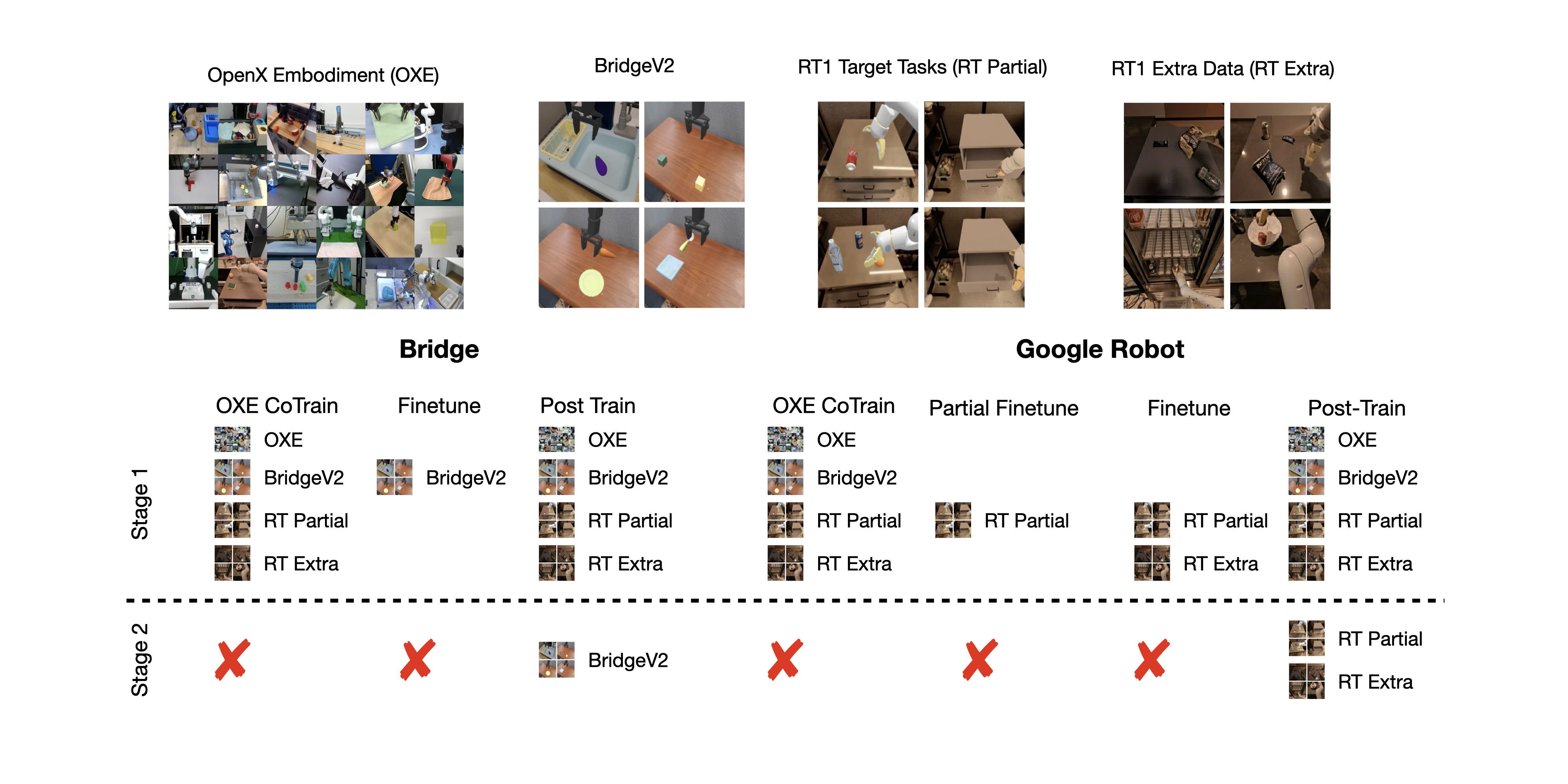}
    \caption{Illustration of training configurations for Bridge and Google Robot. The red crosses (\ding{55}) denote the excluded training stage, and the small icons represent the datasets used, corresponding to those shown above in the figure. Take Bridge Post Train as an example, two stages are employed: Stage 1: Co-training with cross-embodiment data, Bridge V2 dataset, RT1 target dataset, and RT1 extra data. Stage 2: Post-training refinement using only the Bridge V2 dataset.}
    \label{fig:cross_setting}
\end{extendedfigure}

\begin{extendedfigure}  
  \centering
  \includegraphics[width=0.85\linewidth]{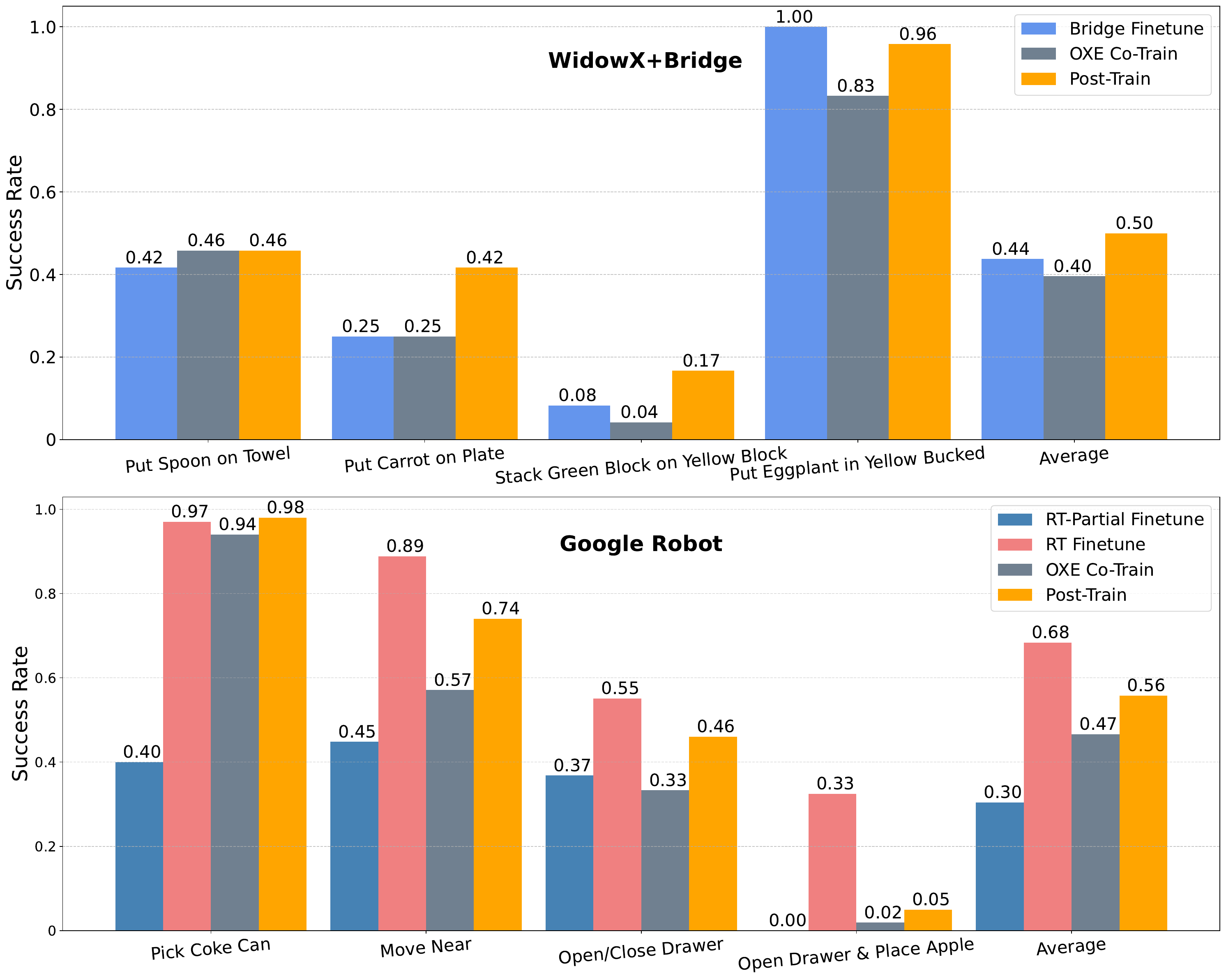}
  \caption{Ablation studies for cross-embodiment training on \textbf{SimpleEnv}. We evaluate four different training recipes. On the \textit{WidowX+Bridge} environments, we test (1) Bridge \textbf{\textit{Finetune}} finetunes the VLA directly on the full Bridge datasets (tested tasks not included); (2) OXE \textbf{\textit{Co-Train}} Co-trains the VLA on OXE dataset~\citep{o2023open}; (3) \textit{\textbf{Post-Train}} trains the OXE Co-trained VLA on Bridge datasets. 
  On the \textit{Google Robot} environments, we test (1)
  \textit{RT-Partial \textbf{Finetune}} finetunes the VLA on tested RT tasks only; (2) \textit{RT \textbf{Finetune}} finetunes the VLA on the full RT dataset (tested tasks included), along with (3) \textit{OXE \textbf{Co-Train}} and (4) \textit{\textbf{Post-Train}} on the tested RT tasks stage.}  
  \label{fig:cross_emb}  
  \vspace{-15pt}
\end{extendedfigure}

\begin{extendedfigure}[htbp]
  \centering
\includegraphics[width=0.4\textwidth]{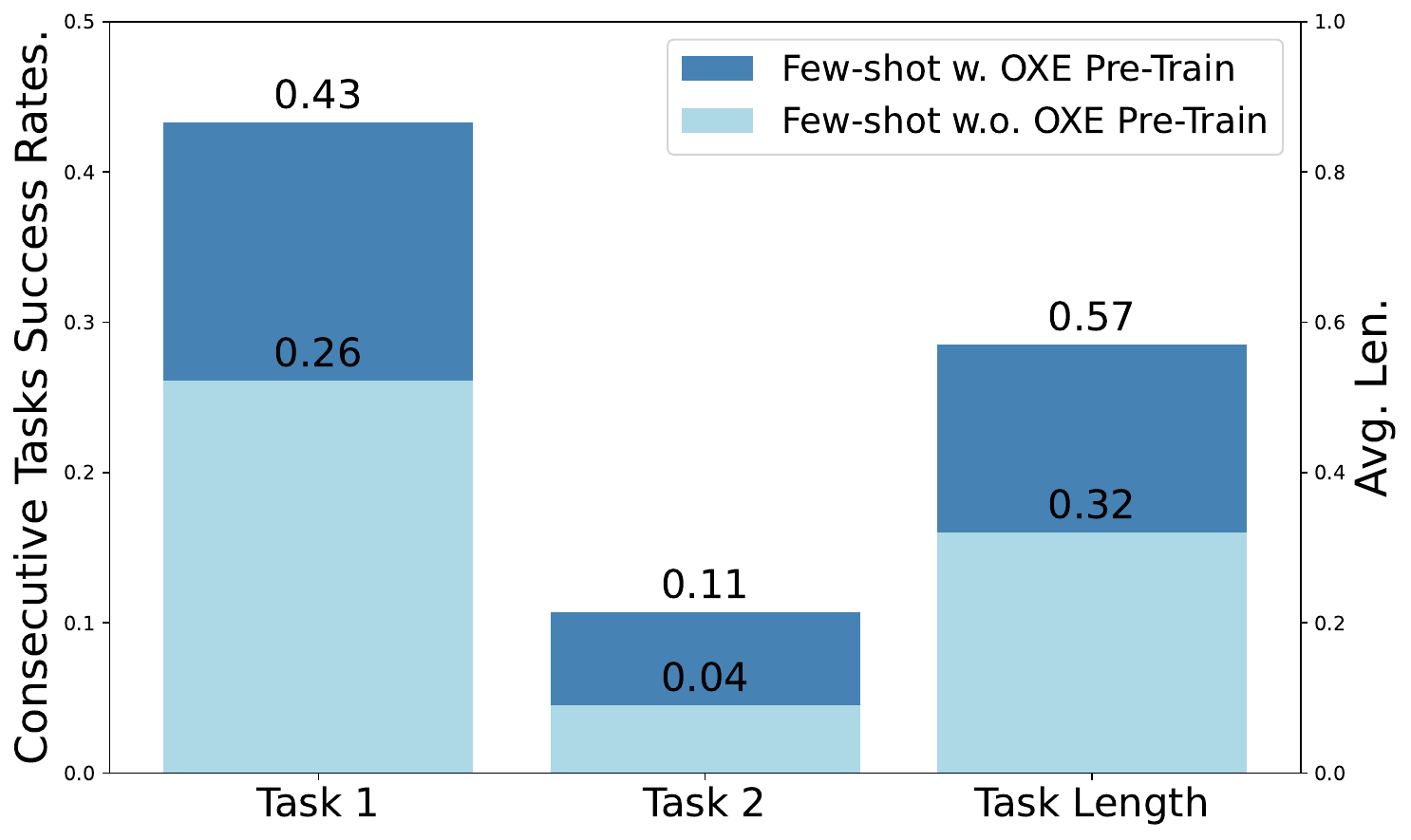}
  \caption{The effect of cross-embodiment pre-training on OXE datasets for \textit{\textbf{few-shot learning}}.}
  \label{fig:few_shot}
\end{extendedfigure}

\begin{extendedtable*}
\caption{The essential questions, research questions, and findings of the study.
}

\centering
\begin{tabular}
{m{0.25\textwidth}m{0.33\textwidth}m{0.35\textwidth}}
\textbf{Essential Questions} & \textbf{Research Questions} & \textbf{Research Findings} \\
\toprule
\multirow{4}{*}{\textit{Q1}: Why \textbf{VLAs}} & \textit{Q1.1}: Are \textcolor{red}{VLAs} a \textcolor{red}{proper choice} for building generalist robot policies? & \textit{A1.1}: VLA is a \textcolor{red}{promising path} towards generalist robot policies. \\
& \textit{Q1.2}: How does the best VLA built with RoboVLMs perform in real-world scenarios? & \textit{A1.2}: The best setup VLA built with \textcolor{red}{RoboVLMs} appears strong \textcolor{red}{effectiveness} and \textcolor{red}{robustness} in \textcolor{red}{real-world} scenarios. \\
\midrule
\multirow{1}{*}{\textit{Q2}: Which \textbf{Backbone}} & \textit{Q2}: Which type of \textcolor{red}{VLMs} are more suitable for constructing VLAs? & \textit{A2}: \textcolor{red}{Sufficient} vision-language pre-trained on \textcolor{red}{large} vision-language datasets benefit VLAs \\ \midrule
\multirow{8}{*}{\textit{Q3}: How to \textbf{Formulate}}
& \multirow{1}{*}{\textit{Q3.1}}: How to select the best-performing VLA \textcolor{red}{structure}? & \textit{A3.1}: \textcolor{red}{Continuous} action space with \textcolor{red}{policy head} to integrate history is the best structure.\\ 
& \textit{Q3.2}: How do different formulations affect the \red{generalization} and \red{data efficiency} for VLAs? & \textit{A3.2}: Leveraging \red{policy head} for history fusion is the best in terms of generalization and data efficiency. \\
& \textit{Q3.3}: How to design appropriate \textcolor{red}{training objective} and \textcolor{red}{execution paradigm}? & \textit{A3.3}: \textcolor{red}{Diffusion} and \textcolor{red}{MSE} training loss are comparable, executing \textcolor{red}{action chunk} is efficient and effective. \\
& \textit{Q3.4}: Does \red{Mix-of-Expert} structure benefit VLA learning? & \textit{A3.4}: Introducing the Mix-of-Expert structure can improve the \red{performance} and \red{generalization} of VLAs. \\
\midrule
\multirow{1}{*}{\textit{Q4}: When to Leverage \textbf{Extra Data}} & \textit{Q4}: Which training stage should introduce  \textcolor{red}{cross-embodiment} datasets? & \textit{A4}: 1) \textcolor{red}{Extra in-domain data} shows beneficial; 2) \textcolor{red}{Post-training} further improves high-frequency tasks as well as few-shot performance. \\
\bottomrule
\end{tabular}
\label{tab:chapter_split}
\end{extendedtable*}

\begin{extendedtable*}[!t]
\caption{Simulation performances on \textbf{CALVIN} benchmark, all models are trained on split ABCD/ABC, and evaluated on split D. KosMos P.H. represents the VLA utilizing KosMos-2 as backbone and policy head as architecture, built with the RoboVLMs framework, and is maximally trained for 5 epochs. We will continue to use the expression of backbone and structure to represent the VLAs built with RoboVLMs in the following paper. Note that KosMos refers to the VLM backbone we utilized, and ``P.H." refers to the policy head formulation.
}
\centering
\begin{tabularx}{\textwidth}{l@{\extracolsep{\fill}}|>{\centering\arraybackslash}X|>{\centering\arraybackslash}X|>{\centering\arraybackslash}X>{\centering\arraybackslash}X>{\centering\arraybackslash}X>{\centering\arraybackslash}X>{\centering\arraybackslash}X|>{\centering\arraybackslash}X} 
\toprule
\multirow{2}{*}{Method} & \multirow{2}{*}{VLA?} & \multirow{2}{*}{Train} & \multicolumn{5}{c|}{Consecutive tasks success rates} & \multirow{2}{*}{\makecell{\textit{Avg.}\\\textit{Len.}}}  \\ \cmidrule{4-8} 
 &  &  & 1 & 2 & 3 & 4 & 5 &  \\ \hline
MCIL & \xmark & \multirow{8}{*}{ABCD} & 0.373 & 0.027 & 0.002 & 0.000 & 0.000 & 0.40 \\
R3M (Frozen) & \xmark &  & 0.085 & 0.005 & 0.001 & 0.000 & 0.000 & 0.10 \\
Voltron (Frozen) & \xmark &  & 0.101 & 0.003 & 0.001 & 0.000 & 0.000 & 0.11 \\
Voltron (Fine-tuned) & \xmark &  & 0.837 & 0.566 & 0.352 & 0.208 & 0.115 & 2.08 \\
RT-1 & \xmark &  & 0.844 & 0.617 & 0.438 & 0.323 & 0.227 & 2.45 \\
HULC & \xmark &  & 0.889 & 0.733 & 0.587 & 0.475 & 0.383 & 3.06 \\
GR-1 & \cmark &  & 0.949 & 0.896 & 0.844 & 0.789 & 0.731 & 4.21 \\
{KosMos P.H. (RoboVLMs)} & \cmark &  & {\bf 0.967} & {\bf 0.930} & {\bf 0.899} & {\bf 0.865} & {\bf 0.826} & {\bf 4.49} \\ \midrule 
MCIL & \xmark & \multirow{7}{*}{ABC} & 0.304 & 0.013 & 0.002 & 0.000 & 0.000 & 0.31 \\
Voltron (Frozen) & \xmark &  & 0.026 & 0.001 & 0.000 & 0.000 & 0.000 & 0.03 \\
Voltron (Fine-tuned)& \xmark &  & 0.569 & 0.272 & 0.105 & 0.038 & 0.014 & 1.00 \\
RT-1 & \xmark &  & 0.533 & 0.222 & 0.094 & 0.038 & 0.013 & 0.90 \\
HULC & \xmark &  & 0.418 & 0.165 & 0.057 & 0.019 & 0.011 & 0.67 \\
GR-1 & \cmark &  & 0.854 & 0.712 & 0.596 & 0.497 & 0.401 & 3.06 \\
{KosMos P.H. (RoboVLMs)} & \cmark &  & {\bf 0.980} & {\bf 0.936} & {\bf 0.854} & {\bf 0.778} & {\bf 0.704} & {\bf 4.25} \\ \bottomrule
\end{tabularx}
\label{tab:performance_calvin}
\end{extendedtable*}

\begin{extendedtable*}[!t]
    \centering
    \caption{The detailed performance on SimplerEnv.}
    \label{tab:simpler_merge}
    
    \begin{subtable}{\textwidth}
        \centering
        \caption{Simulation performance on \textbf{SimplerEnv}-\textit{WidowX+Bridge} environments. We report both the final success rates
(final) along with the sub-task success rates (e.g., \texttt{Grasp Spoon}).}
        \resizebox{\textwidth}{!}{
\begin{tabular}{l|cc|cc|cc|cc}
\toprule
\multirow{2}{*}{Method} & \multicolumn{2}{c|}{\texttt{Put Spoon on Towel}} & \multicolumn{2}{c|}{\texttt{Put Carrot on Plate}} & \multicolumn{2}{c|}{\texttt{Stack Green Block on Yellow Block}} & \multicolumn{2}{c}{\texttt{Put Eggplant in Yellow Basket}} \\ \cmidrule{2-9} 
 & \texttt{Grasp Spoon} & final & \texttt{Grasp Carrot} & final & \texttt{Grasp Green Block} & final & \texttt{Grasp Eggplant} & final \\ \midrule
RT-1-X & 0.167 & 0.000 & 0.208 & 0.042 & 0.083 & 0.000 & 0.000 & 0.000 \\
Octo-Base & 0.347 & 0.125 & 0.528 & 0.083 & 0.319 & 0.000 & 0.667 & 0.431 \\
Octo-Small & {\bf 0.778} & {\bf 0.472} & 0.278 & 0.097 & 0.403 & 0.042 & 0.875 & 0.569 \\
OpenVLA-7b & 0.041 & 0.000 & 0.333 & 0.000 & 0.125 & 0.000 & 0.083 & 0.041 \\
RoboVLMs (Ours) & 0.583 & 0.458 & {\bf 0.458} & {\bf 0.417} & {\bf 0.375} & {\bf 0.167} & {\bf 0.958} & {\bf 0.958} \\ \bottomrule
\end{tabular}
\vspace{-50pt}}
\label{tab:simpler_bridge}
    \end{subtable}
    
    \vspace{0.5cm}
    
    \begin{subtable}{\textwidth}
        \centering
        \caption{Simulation performance on \textbf{SimplerEnv}-\textit{Google Robot} environments.}
        \begin{tabularx}{\textwidth}{l|>{\centering\arraybackslash}X>{\centering\arraybackslash}X>{\centering\arraybackslash}X>{\centering\arraybackslash}X|c|ccc|c}
\toprule
\multirow{4}{*}{Method} & \multicolumn{4}{c|}{\multirow{2}{*}{\texttt{Pick Coke Can}}} & \multirow{2}{*}{\texttt{Move Near}} & \multicolumn{3}{c|}{\multirow{2}{*}{\texttt{Open/Close Drawer}}} & \multirow{2}{*}{\begin{tabular}[c]{@{}c@{}}\texttt{Open Top Drawer} \\ \texttt{and Place Apple} \end{tabular}} \\
 & \multicolumn{4}{c}{} &  & \multicolumn{3}{c}{} &  \\ \cmidrule{2-10} 
 & \begin{tabular}[c]{@{}c@{}}Horizontal\\ Laying\end{tabular} & \begin{tabular}[c]{@{}c@{}}Vertical\\ Laying\end{tabular} & Standing & Average & Average & Open & Close & Average & Average  \\ \midrule
RT-1 (Converged) & 0.960 & 0.900 & 0.710 & {\bf 0.857} & 0.442 & {\bf 0.601} & 0.861 & {\bf 0.730} & 0.000 \\
RT-1 ($15\%$) & 0.860 & 0.790 & 0.480 & 0.710 & 0.354 & 0.463 & 0.667 & 0.565 & 0.000 \\
RT-1-X & 0.820 & 0.330 & 0.550 & 0.567 & 0.317 & 0.296 & {\bf 0.891} & 0.597 & 0.213 \\
RT-2-X & 0.740 & 0.740 & 0.880 & 0.787 & 0.779 & 0.157 & 0.343 & 0.250 & 0.037 \\
Octo-Base & 0.210 & 0.210 & 0.090 & 0.170 & 0.042 & 0.009 & 0.444 & 0.227 & 0.000 \\
RT-1 (Begin) & 0.050 & 0.000 & 0.030 & 0.027 & 0.050 & 0.000 & 0.278 & 0.139 & 0.000 \\
OpenVLA-7b & 0.270 & 0.030 & 0.190 & 0.163 & 0.462 & 0.194 & 0.518 & 0.356 & 0.000 \\
RoboVLMs (Ours) & {\bf 1.000} & {\bf 0.910} & {\bf 1.000} & {\bf 0.970} & {\bf 0.888} & 0.565 & 0.537 & 0.551 & {\bf 0.325} \\ \bottomrule

\end{tabularx}
\label{tab:simpler_rt}
    \end{subtable}

    \vspace{0.5cm}
    
    \begin{subtable}{\textwidth}
        \centering
        \caption{Results on WidowX + Bridge setup for cross-embodiment dataset.}
        \resizebox{\textwidth}{!}{
\begin{tabular}{l|cc|cc|cc|cc}
\toprule
\multirow{2}{*}{Method} & \multicolumn{2}{c|}{\texttt{Put Spoon on Towel}} & \multicolumn{2}{c|}{\texttt{Put Carrot on Plate}} & \multicolumn{2}{c|}{\texttt{Stack Green Block on Yellow Block}} & \multicolumn{2}{c}{\texttt{Put Eggplant in Yellow Basket}} \\ \cmidrule{2-9} 
 & \texttt{Grasp Spoon} & final & \texttt{Grasp Carrot} & final & \texttt{Grasp Green Block} & final & \texttt{Grasp Eggplant} & final \\ \midrule
Cross-Emb Pre-Train & 0.458 & {\bf 0.458} & 0.333 & 0.250 & {\bf 0.458} & 0.042 & 0.916 & 0.833 \\
In-domain Full Finetune & {\bf 0.625} & 0.417 & 0.500 & 0.250 & 0.417 & 0.083 & {\bf 1.000} & {\bf 1.000} \\
Post Train & 0.583 & {\bf 0.458} & {\bf 0.458} &  {\bf 0.417} &  0.375 & {\bf 0.167} & 0.958 & 0.958 \\ 
\hline
\end{tabular}
\label{tab:cross_bridge}
}
    \end{subtable}

    \vspace{0.5cm}
    
    \begin{subtable}{\textwidth}
        \centering
        \caption{Results on WidowX + Bridge setup for cross-embodiment dataset.}
        \resizebox{\textwidth}{!}{
\begin{tabularx}{\textwidth}{l>{\centering\arraybackslash}X>{\centering\arraybackslash}X>{\centering\arraybackslash}X>{\centering\arraybackslash}X|>{\centering\arraybackslash}X|ccc|cc}
\toprule
\multirow{4}{*}{Method} & \multicolumn{4}{c|}{\multirow{2}{*}{\texttt{Pick Coke Can}}} & \multirow{2}{*}{\texttt{Move Near}} & \multicolumn{3}{c|}{\multirow{2}{*}{\texttt{Open/Close Drawer}}} & \multirow{2}{*}{\begin{tabular}[c]{@{}c@{}}\texttt{Open Top Drawer} \\ \texttt{and Place Apple} \end{tabular}} \\
 & \multicolumn{4}{c|}{} &  & \multicolumn{3}{c|}{} &  \\ \cmidrule{2-10} 
 & \begin{tabular}[c]{@{}c@{}}Horizontal\\ Laying\end{tabular} & \begin{tabular}[c]{@{}c@{}}Vertical\\ Laying\end{tabular} & Standing & Average & Average & Open & Close & Average & Average  \\ \midrule
Cross-Emb Pre-Train & 0.940 & 0.840 & 0.890 & 0.890 & 0.571 & 0.222 & 0.444 & 0.333 & 0.019 \\
Target Task Finetune & 0.420 & 0.320 & 0.440 & 0.400 & 0.448 & 0.167 & 0.567 & 0.368 & 0.000 \\
In-domain Full Finetune & {\bf 1.000} & 0.910 & {\bf 1.000} & 0.970 & {\bf 0.888} & {\bf 0.565} & 0.537 & {\bf 0.551} & {\bf 0.325} \\
Post Train & 0.990 & {\bf 0.960} & {\bf 1.000} & {\bf 0.980} & 0.740 & 0.250 & {\bf 0.670} & 0.460 & 0.050 \\ \hline
\end{tabularx}
\label{tab:cross_rt}
}
    \end{subtable}

\end{extendedtable*}

\begin{extendedtable*}[!t]
\caption{
The performance of VLAs implemented with different formulations and training data scales. The results for 0.1x and 1x data are the best-behaved model checkpoints within 5 epochs, and the results for 5x data are the model performance at 1\textsuperscript{st} epoch. We name different implemented VLAs by their VLM backbones and the way of history modeling. Results are reported with models trained maximally  within 5 epochs on the ABCD training splits.
}
\centering
\begin{tabularx}{\textwidth}{l@{\extracolsep{\fill}}|>{\centering\arraybackslash}X|>{\centering\arraybackslash}X>{\centering\arraybackslash}X>{\centering\arraybackslash}X>{\centering\arraybackslash}X>{\centering\arraybackslash}X|>{\centering\arraybackslash}X} 
\toprule
\multirow{2}{*}{\makecell{VLA\\Architecture}} & \multirow{2}{*}{\makecell{Data\\Scale}} & \multicolumn{5}{c|}{Consecutive tasks success rates} & \multirow{2}{*}{\makecell{\textit{Avg.}\\\textit{Len.}}}\\ \cmidrule{3-7} 
 &  & 1 & 2 & 3 & 4 & 5 &  \\ \midrule
Flamingo P.H. 3B \quad & \multirow{5}{*}{0.1x} & 0.120 & 0.007  & 0.000  & 0.000  & 0.000  &  0.13\\
Flamingo P.H. 4B &  & 0.448 & 0.084  & 0.014  & 0.003  & 0.001  &  0.55\\
Flamingo P.H. 9B &  & 0.547   & 0.190  & 0.067  & 0.020  & 0.003  &  0.83 \\
KosMos Inter. &  & 0.938 & 0.701 & 0.445 & 0.270 & 0.140 & 2.49 \\
KosMos P.H. &  & 0.958 & 0.684 & 0.431 & 0.270 & 0.176 &  2.52 \\ \midrule
Flamingo P.H. 3B & \multirow{5}{*}{1x} & 0.964 & 0.896 & 0.824 & 0.740 & 0.662 & 4.09\\
Flamingo P.H. 4B &  & 0.936 & 0.847 & 0.750 & 0.667 & 0.586 & 3.79\\
Flamingo P.H. 9B &   & 0.955 & 0.879 & 0.784 & 0.714 & 0.634 & 3.97  \\
KosMos Inter. &  & 0.987 & 0.915 & 0.824 & 0.737 & 0.660 & 4.12 \\
KosMos P.H. &  & 0.967 & 0.930 & 0.899 & 0.865 & 0.826 & 4.49 \\ \midrule
Flamingo P.H. 3B & \multirow{3}{*}{5x} & 0.971 & 0.916 & 0.856 & 0.794 & 0.716 & 4.21 \\
KosMos Inter. &  & 0.989 & 0.940 & 0.892 & 0.842 & 0.795 & 4.46 \\
KosMos P.H. &  & 0.968 & 0.937 & 0.903 & 0.872 & 0.830 & 4.51 \\ \bottomrule
\end{tabularx}
\label{tab:data_eff_calvin}
\end{extendedtable*}

\begin{extendedtable*}[!t]
\caption{The performance of the built VLAs based on VLMs with different image token numbers and VL pre-train data scales. The first three rows are flamingo backbones with encoder-decoder structures, the rest backbones are decoder-only structures. Note that for VLMs with multi-stage training, the data scale refers to the data amount utilized for the final stage of fine-tuning. ``UNK'' denotes unknown. 
Results are reported with the model checkpoints trained with 5 epochs on the ABCD training splits, all models are trained with a single side view image for fair comparison. 
We surprisingly found that both LLaVA and Qwen behave badly without an additional resampler to downsample the number of tokens. 
}
\centering
\begin{tabularx}{\textwidth}{l@{\extracolsep{\fill}}|>{\centering\arraybackslash}X>{\centering\arraybackslash}X>{\centering\arraybackslash}X|>{\centering\arraybackslash}X>{\centering\arraybackslash}X>{\centering\arraybackslash}X>{\centering\arraybackslash}X>{\centering\arraybackslash}X|>{\centering\arraybackslash}X} 
\toprule
\multirow{2}{*}{Backbone} & \multirow{2}{*}{\#Token} & \multirow{2}{*}{\makecell{Data\\Scale}} & \multirow{2}{*}{\makecell{Model\\Size}} & \multicolumn{5}{c|}{Consecutive tasks success rates} & \multirow{2}{*}{\makecell{Avg.\\Len.}} \\ \cmidrule{5-9} 
 &  &  &  & 1 & 2 & 3 & 4 & 5 &  \\ \midrule
Flamingo & 64 & 1B+ & 3B & 0.692 & 0.418 & 0.241 & 0.14 & 0.074 & 1.57 \\
Flamingo & 64 & 1B+ & 4B & 0.689 & 0.456 & 0.281 & 0.181 & 0.107 & 1.71\\
Flamingo & 64 & 1B+ & 9B & 0.744 & 0.485 & 0.298 & 0.187 & 0.112 & 1.83\\ \midrule
Qwen-VL & 256 & 350K & 9B & 0.221 & 0.062 & 0.014 & 0.002 & 0.000 & 0.30 \\
MoonDream & 576 & UNK & 3B & 0.717 & 0.473 & 0.296 & 0.198 & 0.127 & 1.81 \\
Uform & 256 & 10M & 1.3B & 0.778 & 0.577 & 0.407 & 0.300 & 0.216 & 2.28 \\
KosMos & 64 & 90M & 2B & 0.922 & 0.807 & 0.701 & 0.615 & 0.549 & 3.59 \\ 
Paligemma & 256 & 10B & 3B & 0.931 & 0.836 & 0.752 & 0.683 & 0.616 & 3.82 \\
\bottomrule
\end{tabularx}
\label{tab:calvin_backbone}
\end{extendedtable*}

\clearpage

\newpage

\bibliographystyle{plainnat}
\bibliography{references}

@article{zhou2025vision,
  title={Vision-Language-Action Model with Open-World Embodied Reasoning from Pretrained Knowledge},
  author={Zhou, Zhongyi and Zhu, Yichen and Wen, Junjie and Shen, Chaomin and Xu, Yi},
  journal={arXiv preprint arXiv:2505.21906},
  year={2025}
}

@article{intelligence2025pi_,
  title={A Vision-Language-Action Model with Open-World Generalization},
  author={Intelligence, Physical and Black, Kevin and Brown, Noah and Darpinian, James and Dhabalia, Karan and Driess, Danny and Esmail, Adnan and Equi, Michael and Finn, Chelsea and Fusai, Niccolo and others},
  journal={arXiv preprint arXiv:2504.16054},
  year={2025}
}

@inproceedings{finn2017model,
  title={Model-agnostic meta-learning for fast adaptation of deep networks},
  author={Finn, Chelsea and Abbeel, Pieter and Levine, Sergey},
  booktitle={International conference on machine learning},
  pages={1126--1135},
  year={2017},
  organization={PMLR}
}

@inproceedings{jaegle2021perceiver,
  title={Perceiver: General perception with iterative attention},
  author={Jaegle, Andrew and Gimeno, Felix and Brock, Andy and Vinyals, Oriol and Zisserman, Andrew and Carreira, Joao},
  booktitle={International conference on machine learning},
  pages={4651--4664},
  year={2021},
  organization={PMLR}
}

@article{o2023open,
  title={Open x-embodiment: Robotic learning datasets and rt-x models},
  author={O'Neill, Abby and Rehman, Abdul and Gupta, Abhinav and Maddukuri, Abhiram and Gupta, Abhishek and Padalkar, Abhishek and Lee, Abraham and Pooley, Acorn and Gupta, Agrim and Mandlekar, Ajay and others},
  journal={arXiv preprint arXiv:2310.08864},
  year={2023}
}

@article{team2024octo,
  title={Octo: An open-source generalist robot policy},
  author={Team, Octo Model and Ghosh, Dibya and Walke, Homer and Pertsch, Karl and Black, Kevin and Mees, Oier and Dasari, Sudeep and Hejna, Joey and Kreiman, Tobias and Xu, Charles and others},
  journal={arXiv preprint arXiv:2405.12213},
  year={2024}
}

@article{brohan2022rt,
  title={Rt-1: Robotics transformer for real-world control at scale},
  author={Brohan, Anthony and Brown, Noah and Carbajal, Justice and Chebotar, Yevgen and Dabis, Joseph and Finn, Chelsea and Gopalakrishnan, Keerthana and Hausman, Karol and Herzog, Alex and Hsu, Jasmine and others},
  journal={arXiv preprint arXiv:2212.06817},
  year={2022}
}

@article{brohan2023rt,
  title={Rt-2: Vision-language-action models transfer web knowledge to robotic control},
  author={Brohan, Anthony and Brown, Noah and Carbajal, Justice and Chebotar, Yevgen and Chen, Xi and Choromanski, Krzysztof and Ding, Tianli and Driess, Danny and Dubey, Avinava and Finn, Chelsea and others},
  journal={arXiv preprint arXiv:2307.15818},
  year={2023}
}

@article{liu2025embodied,
  title={Embodied intelligence: A synergy of morphology, action, perception and learning},
  author={Liu, Huaping and Guo, Di and Cangelosi, Angelo},
  journal={ACM Computing Surveys},
  volume={57},
  number={7},
  pages={1--36},
  year={2025},
  publisher={ACM New York, NY}
}

@article{kim2024openvla,
  title={OpenVLA: An Open-Source Vision-Language-Action Model},
  author={Kim, Moo Jin and Pertsch, Karl and Karamcheti, Siddharth and Xiao, Ted and Balakrishna, Ashwin and Nair, Suraj and Rafailov, Rafael and Foster, Ethan and Lam, Grace and Sanketi, Pannag and others},
  journal={arXiv preprint arXiv:2406.09246},
  year={2024}
}

@article{li2023vision,
  title={Vision-language foundation models as effective robot imitators},
  author={Li, Xinghang and Liu, Minghuan and Zhang, Hanbo and Yu, Cunjun and Xu, Jie and Wu, Hongtao and Cheang, Chilam and Jing, Ya and Zhang, Weinan and Liu, Huaping and others},
  journal={arXiv preprint arXiv:2311.01378},
  year={2023}
}

@article{li2024gr,
  title={GR-MG: Leveraging Partially Annotated Data via Multi-Modal Goal Conditioned Policy},
  author={Li, Peiyan and Wu, Hongtao and Huang, Yan and Cheang, Chilam and Wang, Liang and Kong, Tao},
  journal={arXiv preprint arXiv:2408.14368},
  year={2024}
}

@article{mees2022calvin,
  title={Calvin: A benchmark for language-conditioned policy learning for long-horizon robot manipulation tasks},
  author={Mees, Oier and Hermann, Lukas and Rosete-Beas, Erick and Burgard, Wolfram},
  journal={IEEE Robotics and Automation Letters},
  volume={7},
  number={3},
  pages={7327--7334},
  year={2022},
  publisher={IEEE}
}

@article{li2024evaluating,
  title={Evaluating Real-World Robot Manipulation Policies in Simulation},
  author={Li, Xuanlin and Hsu, Kyle and Gu, Jiayuan and Pertsch, Karl and Mees, Oier and Walke, Homer Rich and Fu, Chuyuan and Lunawat, Ishikaa and Sieh, Isabel and Kirmani, Sean and others},
  journal={arXiv preprint arXiv:2405.05941},
  year={2024}
}

@article{cheang2024gr,
  title={GR-2: A Generative Video-Language-Action Model with Web-Scale Knowledge for Robot Manipulation},
  author={Cheang, Chi-Lam and Chen, Guangzeng and Jing, Ya and Kong, Tao and Li, Hang and Li, Yifeng and Liu, Yuxiao and Wu, Hongtao and Xu, Jiafeng and Yang, Yichu and others},
  journal={arXiv preprint arXiv:2410.06158},
  year={2024}
}

@misc{jang2022bc,
  title={Bc-z: Zero-shot task generalization with robotic imitation learning},
  author={Jang, Eric and Irpan, Alex and Khansari, Mohi and Kappler, Daniel and Ebert, Frederik and Lynch, Corey and Levine, Sergey and Finn, Chelsea},
  booktitle={Conference on Robot Learning},
  pages={991--1002},
  year={2022},
  organization={PMLR}
}

@article{torne2024reconciling,
  title={Reconciling reality through simulation: A real-to-sim-to-real approach for robust manipulation},
  author={Torne, Marcel and Simeonov, Anthony and Li, Zechu and Chan, April and Chen, Tao and Gupta, Abhishek and Agrawal, Pulkit},
  journal={arXiv preprint arXiv:2403.03949},
  year={2024}
}

@misc{radosavovic2023real,
  title={Real-world robot learning with masked visual pre-training},
  author={Radosavovic, Ilija and Xiao, Tete and James, Stephen and Abbeel, Pieter and Malik, Jitendra and Darrell, Trevor},
  booktitle={Conference on Robot Learning},
  pages={416--426},
  year={2023},
  organization={PMLR}
}

@article{nair2022r3m,
  title={R3m: A universal visual representation for robot manipulation},
  author={Nair, Suraj and Rajeswaran, Aravind and Kumar, Vikash and Finn, Chelsea and Gupta, Abhinav},
  journal={arXiv preprint arXiv:2203.12601},
  year={2022}
}

@article{wu2023unleashing,
  title={Unleashing large-scale video generative pre-training for visual robot manipulation},
  author={Wu, Hongtao and Jing, Ya and Cheang, Chilam and Chen, Guangzeng and Xu, Jiafeng and Li, Xinghang and Liu, Minghuan and Li, Hang and Kong, Tao},
  journal={arXiv preprint arXiv:2312.13139},
  year={2023}
}

@article{zhao2023learning,
  title={Learning fine-grained bimanual manipulation with low-cost hardware},
  author={Zhao, Tony Z and Kumar, Vikash and Levine, Sergey and Finn, Chelsea},
  journal={arXiv preprint arXiv:2304.13705},
  year={2023}
}

@article{zawalski2024robotic,
  title={Robotic control via embodied chain-of-thought reasoning},
  author={Zawalski, Micha{\l} and Chen, William and Pertsch, Karl and Mees, Oier and Finn, Chelsea and Levine, Sergey},
  journal={arXiv preprint arXiv:2407.08693},
  year={2024}
}

@misc{zhao2020sim,
  title={Sim-to-real transfer in deep reinforcement learning for robotics: a survey},
  author={Zhao, Wenshuai and Queralta, Jorge Pe{\~n}a and Westerlund, Tomi},
  booktitle={2020 IEEE symposium series on computational intelligence (SSCI)},
  pages={737--744},
  year={2020},
  organization={IEEE}
}

@article{graves2012long, 
    title={Long short-term memory}, 
    author={Graves, Alex and Graves, Alex}, 
    journal={Supervised sequence labelling with recurrent neural networks}, 
    pages={37--45}, 
    year={2012}, 
    publisher={Springer} 
}

@article{chi2023diffusion,
  title={Diffusion policy: Visuomotor policy learning via action diffusion},
  author={Chi, Cheng and Xu, Zhenjia and Feng, Siyuan and Cousineau, Eric and Du, Yilun and Burchfiel, Benjamin and Tedrake, Russ and Song, Shuran},
  journal={The International Journal of Robotics Research},
  pages={02783649241273668},
  year={2023},
  publisher={SAGE Publications Sage UK: London, England}
}

@article{floridi2020gpt,
  title={GPT-3: Its nature, scope, limits, and consequences},
  author={Floridi, Luciano and Chiriatti, Massimo},
  journal={Minds and Machines},
  volume={30},
  pages={681--694},
  year={2020},
  publisher={Springer}
}

@article{vaswani2017attention,
  title={Attention is all you need},
  author={Vaswani, A},
  journal={Advances in Neural Information Processing Systems},
  year={2017}
}

@article{chung2014empirical,
  title={Empirical evaluation of gated recurrent neural networks on sequence modeling},
  author={Chung, Junyoung and Gulcehre, Caglar and Cho, KyungHyun and Bengio, Yoshua},
  journal={arXiv preprint arXiv:1412.3555},
  year={2014}
}

@article{medsker2001recurrent,
  title={Recurrent neural networks},
  author={Medsker, Larry R and Jain, Lakhmi and others},
  journal={Design and Applications},
  volume={5},
  number={64-67},
  pages={2},
  year={2001}
}

@article{ke20243d,
  title={3d diffuser actor: Policy diffusion with 3d scene representations},
  author={Ke, Tsung-Wei and Gkanatsios, Nikolaos and Fragkiadaki, Katerina},
  journal={arXiv preprint arXiv:2402.10885},
  year={2024}
}

@article{yue2024deer,
  title={DeeR-VLA: Dynamic Inference of Multimodal Large Language Models for Efficient Robot Execution},
  author={Yue, Yang and Wang, Yulin and Kang, Bingyi and Han, Yizeng and Wang, Shenzhi and Song, Shiji and Feng, Jiashi and Huang, Gao},
  journal={arXiv preprint arXiv:2411.02359},
  year={2024}
}

@article{liu2024robouniview,
  title={RoboUniView: Visual-Language Model with Unified View Representation for Robotic Manipulation},
  author={Liu, Fanfan and Yan, Feng and Zheng, Liming and Feng, Chengjian and Huang, Yiyang and Ma, Lin},
  journal={arXiv preprint arXiv:2406.18977},
  year={2024}
}

@article{liu2024robomamba,
  title={RoboMamba: Multimodal State Space Model for Efficient Robot Reasoning and Manipulation},
  author={Liu, Jiaming and Liu, Mengzhen and Wang, Zhenyu and Lee, Lily and Zhou, Kaichen and An, Pengju and Yang, Senqiao and Zhang, Renrui and Guo, Yandong and Zhang, Shanghang},
  journal={arXiv preprint arXiv:2406.04339},
  year={2024}
}

@article{reed2022generalist,
  title={A generalist agent},
  author={Reed, Scott and Zolna, Konrad and Parisotto, Emilio and Colmenarejo, Sergio Gomez and Novikov, Alexander and Barth-Maron, Gabriel and Gimenez, Mai and Sulsky, Yury and Kay, Jackie and Springenberg, Jost Tobias and others},
  journal={arXiv preprint arXiv:2205.06175},
  year={2022}
}

@article{bousmalis2023robocat,
  title={Robocat: A self-improving foundation agent for robotic manipulation},
  author={Bousmalis, Konstantinos and Vezzani, Giulia and Rao, Dushyant and Devin, Coline and Lee, Alex X and Bauza, Maria and Davchev, Todor and Zhou, Yuxiang and Gupta, Agrim and Raju, Akhil and others},
  journal={arXiv preprint arXiv:2306.11706},
  year={2023}
}

@misc{walke2023bridgedata,
    title={BridgeData V2: A Dataset for Robot Learning at Scale},
    author={Walke, Homer and Black, Kevin and Lee, Abraham and Kim, Moo Jin and Du, Max and Zheng, Chongyi and Zhao, Tony and Hansen-Estruch, Philippe and Vuong, Quan and He, Andre and Myers, Vivek and Fang, Kuan and Finn, Chelsea and Levine, Sergey},
    booktitle={Conference on Robot Learning (CoRL)},
    year={2023}
}

@article{black2024pi_0,
  title={$\pi_0$: A Vision-Language-Action Flow Model for General Robot Control},
  author={Black, Kevin and Brown, Noah and Driess, Danny and Esmail, Adnan and Equi, Michael and Finn, Chelsea and Fusai, Niccolo and Groom, Lachy and Hausman, Karol and Ichter, Brian and others},
  journal={arXiv preprint arXiv:2410.24164},
  year={2024}
}

@article{mees2022matters,
  title={What matters in language conditioned robotic imitation learning over unstructured data},
  author={Mees, Oier and Hermann, Lukas and Burgard, Wolfram},
  journal={IEEE Robotics and Automation Letters},
  volume={7},
  number={4},
  pages={11205--11212},
  year={2022},
  publisher={IEEE}
}

@article{alayrac2022flamingo,
  title={Flamingo: a visual language model for few-shot learning},
  author={Alayrac, Jean-Baptiste and Donahue, Jeff and Luc, Pauline and Miech, Antoine and Barr, Iain and Hasson, Yana and Lenc, Karel and Mensch, Arthur and Millican, Katherine and Reynolds, Malcolm and others},
  journal={Advances in neural information processing systems},
  volume={35},
  pages={23716--23736},
  year={2022}
}

@article{liu2024visual,
  title={Visual instruction tuning},
  author={Liu, Haotian and Li, Chunyuan and Wu, Qingyang and Lee, Yong Jae},
  journal={Advances in neural information processing systems},
  volume={36},
  year={2024}
}

@article{peng2023kosmos,
  title={Kosmos-2: Grounding multimodal large language models to the world},
  author={Peng, Zhiliang and Wang, Wenhui and Dong, Li and Hao, Yaru and Huang, Shaohan and Ma, Shuming and Wei, Furu},
  journal={arXiv preprint arXiv:2306.14824},
  year={2023}
}

@article{bai2023qwen,
  title={Qwen-vl: A frontier large vision-language model with versatile abilities},
  author={Bai, Jinze and Bai, Shuai and Yang, Shusheng and Wang, Shijie and Tan, Sinan and Wang, Peng and Lin, Junyang and Zhou, Chang and Zhou, Jingren},
  journal={arXiv preprint arXiv:2308.12966},
  year={2023}
}

@article{beyer2024paligemma,
  title={Paligemma: A versatile 3b vlm for transfer},
  author={Beyer, Lucas and Steiner, Andreas and Pinto, Andr{\'e} Susano and Kolesnikov, Alexander and Wang, Xiao and Salz, Daniel and Neumann, Maxim and Alabdulmohsin, Ibrahim and Tschannen, Michael and Bugliarello, Emanuele and others},
  journal={arXiv preprint arXiv:2407.07726},
  year={2024}
}

@misc{moondream,
  author = {Vikhyat},
  year = {2024},
  url = {https://github.com/vikhyat/moondream},
  title = {Moondream, tiny vision language model}
}

@misc{uform-gen,
  author = {Unum-cloud},
  year = {2024},
  url = {https://huggingface.co/unum-cloud/uform-gen2-qwen-500m},
  title = {UForm: Pocket-Sized Multimodal AI For Content Understanding and Generation}
}

@article{dosovitskiy2020image,
  title={An image is worth 16x16 words: Transformers for image recognition at scale},
  author={Dosovitskiy, Alexey},
  journal={arXiv preprint arXiv:2010.11929},
  year={2020}
}

@article{jiang2022vima,
  title={Vima: General robot manipulation with multimodal prompts},
  author={Jiang, Yunfan and Gupta, Agrim and Zhang, Zichen and Wang, Guanzhi and Dou, Yongqiang and Chen, Yanjun and Fei-Fei, Li and Anandkumar, Anima and Zhu, Yuke and Fan, Linxi},
  journal={arXiv preprint arXiv:2210.03094},
  year={2022}
}

@article{zhen20243d,
  title={3d-vla: A 3d vision-language-action generative world model},
  author={Zhen, Haoyu and Qiu, Xiaowen and Chen, Peihao and Yang, Jincheng and Yan, Xin and Du, Yilun and Hong, Yining and Gan, Chuang},
  journal={arXiv preprint arXiv:2403.09631},
  year={2024}
}

@misc{wang2022ofa,
  title={Ofa: Unifying architectures, tasks, and modalities through a simple sequence-to-sequence learning framework},
  author={Wang, Peng and Yang, An and Men, Rui and Lin, Junyang and Bai, Shuai and Li, Zhikang and Ma, Jianxin and Zhou, Chang and Zhou, Jingren and Yang, Hongxia},
  booktitle={International conference on machine learning},
  pages={23318--23340},
  year={2022},
  organization={PMLR}
}

@article{ye2024latent,
  title={Latent action pretraining from videos},
  author={Ye, Seonghyeon and Jang, Joel and Jeon, Byeongguk and Joo, Sejune and Yang, Jianwei and Peng, Baolin and Mandlekar, Ajay and Tan, Reuben and Chao, Yu-Wei and Lin, Bill Yuchen and others},
  journal={arXiv preprint arXiv:2410.11758},
  year={2024}
}

@article{nagrani2021attention,
  title={Attention bottlenecks for multimodal fusion},
  author={Nagrani, Arsha and Yang, Shan and Arnab, Anurag and Jansen, Aren and Schmid, Cordelia and Sun, Chen},
  journal={Advances in neural information processing systems},
  volume={34},
  pages={14200--14213},
  year={2021}
}

@article{xu2021vlm,
  title={Vlm: Task-agnostic video-language model pre-training for video understanding},
  author={Xu, Hu and Ghosh, Gargi and Huang, Po-Yao and Arora, Prahal and Aminzadeh, Masoumeh and Feichtenhofer, Christoph and Metze, Florian and Zettlemoyer, Luke},
  journal={arXiv preprint arXiv:2105.09996},
  year={2021}
}

@article{yang2023dawn,
  title={The dawn of lmms: Preliminary explorations with gpt-4v (ision)},
  author={Yang, Zhengyuan and Li, Linjie and Lin, Kevin and Wang, Jianfeng and Lin, Chung-Ching and Liu, Zicheng and Wang, Lijuan},
  journal={arXiv preprint arXiv:2309.17421},
  volume={9},
  number={1},
  pages={1},
  year={2023}
}

@article{moe,
  title={Outrageously large neural networks: The sparsely-gated mixture-of-experts layer},
  author={Shazeer, Noam and Mirhoseini, Azalia and Maziarz, Krzysztof and Davis, Andy and Le, Quoc and Hinton, Geoffrey and Dean, Jeff},
  journal={International Conference on Learning Representations},
  year={2017}
}

@article{lipman2022flow,
  title={Flow matching for generative modeling},
  author={Lipman, Yaron and Chen, Ricky TQ and Ben-Hamu, Heli and Nickel, Maximilian and Le, Matt},
  journal={arXiv preprint arXiv:2210.02747},
  year={2022}
}
\newpage

\begin{appendices}


\section{Discussion}\label{sec:discuss}

This empirical study mainly focuses on what matters in building Visual-Language-Action models (VLAs).
We raise four essential questions for building a VLA: {\bf Why} do we need VLAs instead of other generalist policies, and by outperforming the existing methods by a large margin, we illustrate the necessity of studying VLAs. For the next step, we describe the key components for building a VLM-based VLA: {\bf Which} kind of VLM backbone to utilize, {\bf How} to train the model to generate action, and {\bf When} should we add cross-embodiment data into training stages. To answer these questions, we built a unified framework for a fair comparison of VLAs and designed a series of bottom-up systematic experiments. To answer these questions, we conduct extensive experiments across three simulators and more than 240 rollouts within 20 tasks in real-world scenarios, and we can conclude from the experiments that: For the {\bf Why} question, VLAs could achieve high performance and generalization, and is a promising path to generalist robotics policy; For the {\bf Which} problem, we find that VLMs with sufficient vision-language pre-training over large scale vision-language datasets is suitable for constructing VLAs.
For the {\bf How} problem, we can investigate the performance, generalization, and data efficiency of different VLA structures, and find that integrating history observations is essential for VLAs, and policy head is a more effective and efficient history aggregating method compared with interleaved; For the {\bf When} problem, we compare three training recipes with cross-embodiment integrated at different stages, and conclude that extra in-domain data shows beneficial, and large-scale cross-embodiment pre-training further improves overall as well as few-shot performance. As a byproduct of answers to the raised questions, we built an easy-to-use framework for easily integrating arbitrary VLMs and turning them into VLAs, named {\bf RoboVLMs}. 

\noindent\textbf{Under investigated observations.}
During our experiments, we found that VLAs built upon Qwen-VL and LLaVA, the performance is surprisingly low, compared with their original performance on vision-language tasks. After adding a perceiver resampler after the vision encoder, we found that the VLAs based on Qwen-VL and LLaVA could obtain great performance gains and reach reasonable performance. We hypothesize that the performance gain is related to the image resolution and the number of vision tokens in the input token sequence.

\noindent\textbf{Limitations.}
Although we make every effort to investigate the key challenges in building Vision-Language Agents (VLAs), this work remains preliminary and has several limitations at the current stage.
(1) The action tokenization, policy head, and corresponding training objectives are not fully explored in this work. For example, techniques like VQ-VAE, VQGAN, and FAST action tokenizer remain under-explored in the context of VLAs.
(2) The set of VLM backbones considered in this study is limited and can be actively expanded.
(3) Deploying such large models for real-time robotic control remains a significant challenge.

\noindent\textbf{Future works.} For future work, we envision several potential directions for advancing generalist robot policies. 
Beyond semantic generalization, an ideal generalist robot policy should be capable of handling long-horizon, complex task instructions (e.g., \texttt{make breakfast}), reasoning through executable actions step by step, and generating meaningful physical interactions with its environment. In our future work, we aim to explore the key elements required to develop policies with these advanced capabilities.

We have open-sourced our codebase with detailed guidelines, model weights of the strongest VLAs built by RoboVLMs, along with the real-world dataset used in our experiments. 
We anticipate that our research will bolster the community and expedite progress in the realms of vision-language learning and foundational models for robotics.

\section{Implementation Details}
\label{ap:impl_detail}

\paragraph{Hyper-parameters and Training Details.}
With different formulations, the best setting of hyperparameters like batch size, weight decay, and learning rate could be varied. Although OpenVLA suggests utilizing the same hyperparameters as in the VLM pretrain phase, we find that a varied setting of the hyperparameters could improve the performance.

The hyperparameters for fine-tuning VLAs are mainly derived from the VLMs training setups, for example, we select the weight decay from $[0 , 1e-1]$, and the learning rate as one of $[1e-4, 5e-5, 2e-5, 1e-5]$. 
We conduct a grid search and select the one with the best performance. We set the global batch size as 128 and the warmup ratio is 0.25 epoch (5K steps for OpenX Embodiment pre-train). All models included in this paper are trained on a cluster of 4 x 8 A100-80G GPUs.

\begin{supplementarytable*}[htbp]
\caption{Hyper Parameters setup for different experiments. ``EP" denotes the epoch. ``Iters" represents iterations.}
\centering
\resizebox{\textwidth}{!}{
\begin{tabular}{llccccccccc}
\toprule
\multirow{2}{*}{\makecell{Experiment Name}} & \multirow{2}{*}{Backbone} & \multirow{2}{*}{\makecell{Window\\ Size}} & \multirow{2}{*}{\makecell{Chunk\\ Size}} & \multirow{2}{*}{\makecell{Input\\ View}} & \multirow{2}{*}{\makecell{Batch\\ Size}} & \multirow{2}{*}{Warmup} & \multirow{2}{*}{Scheduler} & \multirow{2}{*}{Optimizer} & \multirow{2}{*}{\makecell{Learning\\ Rate}} & \multirow{2}{*}{\makecell{Total\\ Epochs/Iters}}\\ \\ \midrule
CALVIN Perform (Extended Table.~2) & KosMos-2 & 16 & 10 & Side+Wrist & 128 & 0.25 Ep & Constant & AdamW & 1e-4 & 5 Ep \\ \midrule
\multirow{2}{*}{SimplerEnv Training Recipe (Supplementary Fig.~1, Extended Fig.4)} & PaliGemma (Co-Train) & 1 & 10 & Side & 1024 & 500 Iters & Constant & AdamW & 2e-5 & 20K Iters  \\
 & PaliGemma (Post-Train \& Finetune) & 1 & 10 & Side & 128 & 0.25 Ep & Constant & AdamW & 5e-5 & 5 Ep  \\ \midrule
CALVIN VL Pre-train (Fig.~4b) & Flamingo \& KosMos-2 & 16 & 10 & Side+Wrist & 128 & 0.25 Ep & Constant & AdamW & 1e-4 & 5 Ep  \\ \midrule
Real Perform (Fig.~4d) & KosMos-2 & 8 & 10 & Side+Wrist & 128 & 0.25 Ep & Constant & AdamW & 1e-4 & 5 Ep  \\ \midrule
\multirow{2}{*}{VLA Structure (Table.~I)} & LLaVA & 8 & 10 & Side+Wrist & 128 & 0.25 Ep & Constant & AdamW & 2e-5 & 5 Ep  \\
 & Flamingo \& KosMos-2 \& PaliGemma & 16 & 10 & Side+Wrist & 128 & 0.25 Ep & Constant & AdamW & 1e-4 & 5 Ep  \\ \midrule
CALVIN Generalization (Fig.~4c) & All & 16 & 10 & Side+Wrist & 128 & 0.25 Ep & Constant & AdamW & 1e-4 & 5 Ep  \\ \midrule
CALVIN Data Efficiency (Extended Table.~4) & All & 16 & 10 & Side+Wrist & 128 & 0.25 Ep & Constant & AdamW & 1e-4 & 5 Ep  \\ \midrule
CALVIN Backbone (Extended Table.~5) & All & 8 & 10 & Side & 128 & 0.25 Ep & Constant & AdamW & 2e-5 & 5 Ep  \\ \midrule
CALVIN Training Objective (Table.~IIa) & All & 1 & 10 & Side+Wrist & 128 & 0.25 Ep & Constant & AdamW & 2e-5 & 5 Ep  \\ \midrule
CALVIN MoE (Table.~IIb) & All & 1 & 10 & Side+Wrist & 128 & 0.25 Ep & Constant & AdamW & 2e-5 & 5 Ep  \\ \midrule
CALVIN few-shot (Extended Fig.~5) & All & 16 & 10 & Side & 128 & 0 Iter & Constant & AdamW & 2e-5 & 5K Iters  \\ 
\bottomrule
\end{tabular}
\label{tab:train_detail}
}
\end{supplementarytable*}

\paragraph{Checkpoint selection.} 
We find out that, normally, the performance of robot policies does not fully depend on offline evaluation metrics, such as the validation loss, due to the compounding error in long-horizon rollouts.
Therefore, it is challenging to select the best checkpoint during training.
For fair comparisons, we train all VLAs for a fixed number of epochs or timesteps.
Concretely, on CALVIN, we train each model for 5 epochs with a batch size of 128 truncated trajectories and report the performance of the final model.
For SimplerEnv, we train the model for 100K iterations with a batch size of 512 truncated trajectories and report the best-performing model with a 10K-iteration training interval. 
In real-world experiments, we train the model for 5 epochs with a batch size of 512 truncated trajectories, and we only report the performance of the last model.

\section{Benchmark Details}
\label{ap:bench_details}
\noindent{\textbf{CALVIN}} is a simulated benchmark for multi-task table-top manipulation. It provides 24k human-teleoperated demonstrations annotated with language instruction. Each trajectory is less than 64-time steps, which includes 34 pre-defined basic skills: \texttt{rotate blue block right}, \texttt{move slider right}, \texttt{lift red block slider}, \texttt{place slider}, \texttt{turn off light bulb}, \texttt{turn off led light}, \texttt{push in drawer}, \texttt{lift blue block drawer}, \texttt{close drawer}, \texttt{lift pink block slider}, \texttt{lift pink block table}, \texttt{move slider left}, \texttt{open drawer}, \texttt{turn on light bulb}, \texttt{rotate blue block left}, \texttt{push blue block left},\texttt{rotate red block right}, \texttt{turn on led light},\\ \texttt{push pink block right}, \texttt{push red block left}, \texttt{lift blue block table}, \texttt{place in drawer}, \texttt{rotate red block left}, \texttt{push pink block left}, \texttt{lift stacked blocks}, \texttt{lift blue block slider},
\texttt{push red block right}.
The dataset contains four splits as scene A, B, C, and D.
We train and test VLAs on different training/test splits to fully analyze the capabilities, data and training efficiencies.
During evaluation, the robot is required to complete a set of 5 consecutive tasks. 
The metrics are the success rates of finishing these sequential tasks and the average length of completed tasks.
All evaluations are implemented on D split, with 1000 rollouts and 5 consecutive sub-tasks for each rollout.
    
\noindent{\textbf{SimplerEnv}} are designed as a suite of real-to-sim environments, which enables evaluating robotic policies in simulation as efficient, scalable, and informative alternative to real-world evaluation. SimplerEnv created a comparable arena for benchmarking robotics policies on private real-world setups as Google and BridgeData V2.

We adopt the following tasks in the Google Robot setting:
\begin{itemize}
    \item \textbf{\texttt{pick coke can}}. The task assigned to the robot is to pick up an empty Coke can from the table and lift it. Under the standard configuration, the environment is kept free of any distracting elements. The Coke can is arranged in three distinct positions: lying flat horizontally, lying flat vertically, and standing upright. For each of these positions, the can is placed at 25 specific grid points within a defined rectangular area on the table. This setup results in 25 experiments per position, totaling 75 trials across all orientations.
    \item \textbf{\texttt{move \{\textit{obj1}\} near \{\textit{obj2}\}}}. In the experiment, a set of three objects was arranged on the table in a triangular formation. For each trial, one object was assigned the role of the source, another was designated as the target, and the third served as a distractor. This setup resulted in six distinct trials for each triplet and triangular configuration. From a total of eight objects—blue plastic bottle, Pepsi can, orange, 7up can, apple, sponge, Coke can, and Redbull can—five triplets were randomly selected. Additionally, two triangular patterns, upright and inverted, were employed. This design produced a total of 60 trials. 
    \item \textbf{\texttt{(open/close) (top / middle/bottom) drawer}}. In this setup, the robot is placed facing a cabinet equipped with three drawers and tasked with opening or closing a specific drawer. This experiment evaluates the robot's capability to handle articulated objects. The robot is positioned at nine distinct locations on a predefined grid within a rectangular area on the floor. With three drawers and two possible actions (opening or closing), the setup results in a total of 54 trials.
    \item \textbf{\texttt{open top drawer}; \texttt{place apple into top drawer}}. In this experiment, the robot is tasked with opening the top drawer and transferring an apple from the surface of the cabinet into the drawer. This setup evaluates the robot's ability to execute tasks that require multiple sequential actions. The robot is positioned in three distinct locations on the floor, while the apple is placed at nine specific grid points on the cabinet surface, resulting in a total of 27 trials. At the start, the robot operates under the instruction to open the top drawer. Once the robot either signals task completion with a "terminate" token or reaches the midpoint of the allotted time, the instruction transitions to directing the robot to place the apple into the drawer.
\end{itemize}

For the WidowX + Bridge setting, we test the following tasks:
\begin{itemize}
    \item \textbf{\texttt{put the spoon on the towel}}. In this setup, the spoon is positioned at one corner of a square on the tabletop, with the towel placed at a different corner. The square has sides measuring 15 cm in length. The orientation of the spoon alternates between horizontal and vertical, requiring the robot to adjust the orientation of its gripper accordingly. This configuration results in a total of 24 trials.
    \item \textbf{\texttt{put carrot on plate}}. This setup is similar to \texttt{put the spoon on the towel}, but the spoon is replaced with a carrot and the towel is substituted with a plate.
    \item \textbf{\texttt{stack the green block on the yellow block}}. In this experiment, a green block is positioned at one corner of a square on the tabletop, while a yellow block is placed at a different corner. Both blocks measure 3 cm in size. Two square configurations with 10 cm and 20 cm side lengths are used. This setup results in a total of 24 trials.
    \item \textbf{\texttt{put eggplant into yellow basket}}. An eggplant is positioned randomly within the right basin of a sink, while a yellow basket is placed in the left basin. The eggplant's placement varies in both location and orientation but is carefully arranged to remain easily graspable, avoiding proximity to the sink's edges. A total of 24 trials are conducted under this setup.
\end{itemize}
    
\newpage
\section{Detailed Performance on CALVIN}
\label{ap:calvin}
\begin{supplementarytable*}[htbp]
\caption{Generalization and data efficiency of VLAs with or without vision-language pre-train on different settings of the CALVIN benchmark. ``Inter." denotes interleaved modeling. ``P.H." denotes policy head. ``No VL" suggests models without VL pre-training.
``5x" represents training with 5x re-generated training data. 
The results for 5x data are the model performance at 1 epoch, and the best-behaved model checkpoints within 5 epochs for the others.
}
\centering
\begin{tabularx}{\textwidth}{l|>{\centering\arraybackslash}X|>{\centering\arraybackslash}X>{\centering\arraybackslash}X>{\centering\arraybackslash}X>{\centering\arraybackslash}X>{\centering\arraybackslash}X|>{\centering\arraybackslash}X>{\centering\arraybackslash}X}
\toprule
\multirow{2}{*}{Architecture} & \multirow{2}{*}{Train} & \multicolumn{5}{c|}{Consecutive tasks success rates} & \multirow{2}{*}{\makecell{\textit{Avg.}\\\textit{Len.}}} \\ \cmidrule{3-7} 
 &  & 1 & 2 & 3 & 4 & 5 &  \\ \midrule
 KosMos Inter. No VL & \multirow{6}{*}{ABCD} & 0.692 & 0.382 & 0189 & 0.085 & 0.036 & 1.38  \\
 KosMos Inter. &  & 0.987 & 0.915 & 0.824 & 0.737 & 0.660 & 4.12 \\
 \cmidrule{1-1}\cmidrule{3-8}
  KosMos P.H. No VL &  & 0.815 & 0.626 & 0.473 & 0.349 & 0.245 & 2.51  \\
 KosMos P.H. &  & 0.967 & 0.930 & 0.899 & 0.865 & 0.826 & 4.49 \\ 
 \cmidrule{1-1}\cmidrule{3-8}
  Flamingo P.H. 9B No VL &  & 0.733 & 0.444 & 0.270 & 0.161 & 0.077 & 1.69 \\
  Flamingo P.H. 9B &  & 0.955 & 0.879 & 0.784 & 0.714 & 0.634 & 3.97 \\
 \midrule\midrule
 KosMos Inter. No VL & \multirow{4}{*}{5x ABCD} & 0.833 & 0.588 & 0.421 & 0.297 & 0.209 & 2.35\\
KosMos Inter. &  & 0.989 & 0.940 & 0.892 & 0.842 & 0.795 & 4.46 \\
\cmidrule{1-1}\cmidrule{3-8}
KosMos P.H. No VL &  & 0.893 & 0.755 & 0.644 & 0.564 & 0.462 & 3.32 \\
KosMos P.H. &  & 0.968 & 0.937 & 0.903 & 0.872 & 0.830 & 4.51 \\ \midrule\midrule
 KosMos Inter. No VL & \multirow{4}{*}{ABC} & 0.432 & 0.162 & 0.044 & 0.007 & 0.002 & 0.65 \\
 KosMos Inter. &  & 0.824 & 0.684 & 0.524 & 0.376 & 0.296 & 2.70 \\
 \cmidrule{1-1}\cmidrule{3-8}
 KosMos P.H. No VL &  & 0.389 & 0.121 & 0.038 & 0.008 & 0.001 & 0.56 \\
 KosMos P.H. &  & 0.980 & 0.936 & 0.854 & 0.778 & 0.704 & 4.25 \\ 
 \bottomrule
\end{tabularx}
\label{tab:generalization_novl_calvin}
\end{supplementarytable*}

\newpage
\section{Diverse Backbone}
\label{ap:diverse_backbone}
\begin{supplementarytable*}[htbp]  
\caption{Sub-task level success rates by tasks in CALVIN under different VLM backbones. All models trained with 5 epochs on CALVIN ABCD training splits.} 
\label{tab:diverse_backbone}
\centering  
\begin{tabularx}{\textwidth}{l|*{8}{X}} 
\toprule  
Task Name & Flamingo-3B & Flamingo-4B & Flamingo-9B  & KosMos & MoonDream & Paligemma & Qwen & Uform \\
\midrule
\texttt{rotate blue block right} & 0.222 & 0.300 & 0.322  & 0.493 & 0.234 & 0.816 & 0.057 & 0.689 \\
\texttt{move slider right} & 0.736 & 0.989 & 0.805  & 0.987 & 0.988 & 1.000 & 0.442 & 1.000 \\
\texttt{lift red block slider} & 0.681 & 0.620 & 0.301  & 0.856 & 0.500 & 0.975 & 0.051 & 0.387 \\
\texttt{place in slider} & 0.683 & 0.813 & 0.765  & 0.874 & 0.361 & 0.957 & 0.308 & 0.490 \\
\texttt{turn off lightbulb} & 0.935 & 0.920 & 0.857  & 0.927 & 0.988 & 0.992 & 0.051 & 0.907 \\
\texttt{turn off led} & 0.990 & 0.925 & 0.966 & 0.970 & 0.990 & 0.987 & 0.793 & 0.754 \\
\texttt{push into drawer} & 0.422 & 0.318 & 0.296 & 0.705 & 0.688 & 0.814 & 0.000 & 0.743 \\
\texttt{lift blue block drawer} & 1.000 & 1.000 & 0.250 & 0.917 & 0.833 & 1.000 & -- & 0.917 \\
\texttt{close drawer} & 0.118 & 1.00 & 0.990  & 0.986 & 0.991 & 1.000 & 0.160 & 0.950 \\
\texttt{lift pink block slider} & 0.754 & 0.633 & 0.471 & 0.861 & 0.444 & 0.945 & 0.059 & 0.703 \\
\texttt{lift pink block table} & 0.708 & 0.541 & 0.630  & 0.543 & 0.544 & 0.936 & 0.229 & 0.892 \\
\texttt{move slider left} & 0.952 & 0.952 & 0.961  & 0.970 & 1.000 & 1.000 & 0.787 & 0.994 \\
\texttt{open drawer} & 0.986 & 0.796 & 0.983  & 0.980 & 1.000 & 1.000 & 0.231 & 0.992 \\
\texttt{turn on lightbulb} & 0.737 & 0.225 & 0.561  & 0.949 & 0.990 & 1.000 & 0.170 & 0.775 \\
\texttt{rotate blue block left} & 0.358 & 0.588 & 0.596  & 0.636 & 0.500 & 0.955 & 0.000 & 0.964 \\
\texttt{push blue block left} & 0.554 & 0.509 & 0.696  & 0.677 & 0.424 & 0.727 & 0.000 & 0.661 \\
\texttt{rotate red block right} & 0.267 & 0.317 & 0.355 & 0.591 & 0.444 & 0.853 & 0.162 & 0.610 \\
\texttt{turn on led} & 0.962 & 0.864 & 0.902  & 0.985 & 0.973 & 0.970 & 0.562 & 0.526 \\
\texttt{push pink block right} & 0.365 & 0.574 & 0.538  & 0.627 & 0.271 & 0.559 & 0.027 & 0.412 \\
\texttt{push red block left} & 0.383 & 0.542 & 0.727  & 0.562 & 0.338 & 0.653 & 0.043 & 0.435 \\
\texttt{lift blue block table} & 0.470 & 0.474 & 0.564 & 0.611 & 0.691 & 0.955 & 0.111 & 0.889 \\
\texttt{place in drawer} & 0.969 & 0.932 & 0.928  & 0.971 & 0.958 & 0.994 & 0.750 & 0.931 \\
\texttt{rotate red block left} & 0.469 & 0.562 & 0.588  & 0.677 & 0.327 & 0.937 & 0.000 & 0.978 \\
\texttt{push pink block left} & 0.661 & 0.672 & 0.697  & 0.747 & 0.338 & 0.857 & 0.000 & 0.554 \\
\texttt{stack block} & 0.130 & 0.029 & 0.070  & 0.569 & 0.448 & 0.646 & 0.000 & 0.495 \\
\texttt{lift blue block slider} & 0.458 & 0.405 & 0.282  & 0.769 & 0.867 & 0.953 & 0.211 & 0.367 \\
\texttt{push red block right} & 0.233 & 0.561 & 0.557  & 0.457 & 0.194 & 0.414 & 0.103 & 0.350 \\
\texttt{lift red block table} & 0.629 & 0.608 & 0.554  & 0.606 & 0.640 & 0.967 & 0.091 & 0.835 \\
\texttt{lift pink block drawer} & 0.800 & 0.667 & 0.333  & 0.778 & 0.667 & 0.929 & -- & 0.400 \\
\texttt{rotate pink block right} & 0.182 & 0.358 & 0.377  & 0.478 & 0.462 & 0.620 & 0.179 & 0.431 \\
\texttt{unstack block} & 1.000 & 0.500 & 1.000 & 0.946 & 0.950 & 0.935 & -- & 0.880 \\
\texttt{rotate pink block left} & 0.578 & 0.442 & 0.479  & 0.698 & 0.146 & 0.964 & 0.034 & 1.000 \\
\texttt{push blue block right} & 0.091 & 0.421 & 0.320 & 0.400 & 0.066 & 0.324 & 0.000 & 0.500 \\
\texttt{lift red block drawer} & 0.400 & 0.625 & 0.800  & 0.769 & 0.667 & 1.000 & -- & 0.556 \\
\bottomrule  
\end{tabularx}
\end{supplementarytable*}

\newpage
\section{Diverse ph}
\label{ap:diverse_ph}
\begin{supplementarytable*}[htbp]  
\centering  
\caption{Sub-task level success rates by tasks in CALVIN under different training splits and VLM backbones. All models are trained with maximal 5 epochs.} 
\label{tab:diverse_ph}
\small
\begin{tabularx}{\textwidth}{l|*{7}{X}} 
\toprule  
Task Name & flamingo-3b-abc & flamingo-3b-abcd & flamingo-4b-abcd & flamingo-9b-abc & flamingo-9b-abcd & kosmos-abc & kosmos-abcd \\
\midrule
\texttt{rotate blue block right}   & 0.831 & 0.893 & 0.770 & 0.722 & 0.882 & 0.960 & 0.974 \\
\texttt{move slider right}   & 0.163 & 0.993 & 0.992 & 0.471 & 0.996 & 1.000 & 1.000 \\
\texttt{lift red block slider}  & 0.642 & 0.970 & 0.858 & 0.610 & 0.927 & 0.906 & 0.993 \\
\texttt{place in slider}  & 0.739 & 0.828 & 0.911 & 0.654 & 0.910 & 0.744 & 0.960 \\
\texttt{turn off lightbulb} &  0.852 & 1.000 & 0.956 & 0.135 & 0.964 & 1.000 & 1.000 \\
\texttt{turn off led} &  0.835 & 1.000 & 0.994 & 0.765 & 0.981 & 0.970 & 1.000 \\
\texttt{push into drawer} &   0.656 & 0.821 & 0.770 & 0.688 & 0.703 & 0.805 & 0.849 \\
\texttt{lift blue block drawer}  & 0.818 & 0.950 & 1.000 & 0.800 & 0.737 & 0.944 & 1.000 \\
\texttt{close drawer} &  1.000 & 1.000 & 1.000 & 0.978 & 0.995 & 1.000 & 1.000 \\
\texttt{lift pink block slider} &   0.674 & 0.971 & 0.862 & 0.733 & 0.918 & 0.925 & 0.993 \\
\texttt{lift pink block table} &  0.871 & 0.851 & 0.892 & 0.872 & 0.899 & 0.913 & 0.961 \\
\texttt{move slider left} &   0.517 & 0.996 & 1.000 & 0.798 & 0.996 & 0.992 & 1.000 \\
\texttt{open drawer} &   1.000 & 0.997 & 0.982 & 0.991 & 0.997 & 1.000 & 1.000 \\
\texttt{turn on lightbulb} &   0.922 & 0.994 & 0.988 & 0.241 & 0.988 & 0.989 & 1.000 \\
\texttt{rotate blue block left} &  0.915 & 0.939 & 0.925 & 0.755 & 0.848 & 0.985 & 1.000 \\
\texttt{push blue block left} &   0.909 & 0.955 & 0.836 & 0.736 & 0.909 & 0.914 & 0.826 \\
\texttt{rotate red block right} &  0.893 & 0.972 & 0.905 & 0.727 & 0.959 & 1.000 & 0.973 \\
\texttt{turn on led} &  0.956 & 0.988 & 0.976 & 0.777 & 0.994 & 0.977 & 0.995 \\
\texttt{push pink block right} &  0.706 & 0.754 & 0.651 & 0.652 & 0.750 & 0.708 & 0.779 \\
\texttt{push red block left} &   0.983 & 0.920 & 0.949 & 0.911 & 0.908 & 0.910 & 0.848 \\
\texttt{lift blue block table} &   0.936 & 0.956 & 0.927 & 0.876 & 0.931 & 0.995 & 0.995 \\
\texttt{place in drawer} &  1.000 & 0.989 & 0.975 & 0.969 & 0.976 & 1.000 & 1.000 \\
\texttt{rotate red block left} &  0.942 & 0.908 & 0.953 & 0.961 & 0.952 & 1.000 & 1.000 \\
\texttt{push pink block left} &   0.889 & 0.920 & 0.973 & 0.959 & 0.933 & 0.959 & 0.883 \\
\texttt{stack block} & 0.489 & 0.641 & 0.595 & 0.588 & 0.604 & 0.801 & 0.908 \\
\texttt{lift blue block slider} &  0.607 & 0.963 & 0.826 & 0.595 & 0.869 & 0.835 & 0.978 \\
\texttt{push red block right}   & 0.825 & 0.732 & 0.451 & 0.764 & 0.653 & 0.871 & 0.681 \\
\texttt{lift red block table} & 0.891 & 0.939 & 0.975 & 0.958 & 0.989 & 0.989 & 0.984 \\
\texttt{lift pink block drawer} & 0.444 & 0.800 & 0.714 & 0.333 & 0.923 & 0.846 & 0.857 \\
\texttt{rotate pink block right} & 0.768 & 0.896 & 0.794 & 0.787 & 0.789 & 0.985 & 0.986 \\
\texttt{unstack block} &  0.880 & 0.982 & 0.980 & 0.909 & 0.979 & 0.985 & 1.000 \\
\texttt{rotate pink block left} &  0.974 & 0.839 & 0.818 & 0.947 & 0.927 & 1.000 & 0.965 \\
\texttt{push blue block right} &  0.620 & 0.597 & 0.478 & 0.511 & 0.594 & 0.870 & 0.657 \\
\texttt{lift red block drawer} &  0.714 & 1.000 & 1.000 & 0.833 & 0.933 & 0.950 & 1.000 \\
\bottomrule  
\end{tabularx}
\end{supplementarytable*}

\newpage
\newpage



\section{Rollout Examples in SimplerEnv}
\label{ap:simpler_quality}

\begin{supplementaryfigure*}[htbp]
  \centering
    \includegraphics[width=0.8\textwidth]{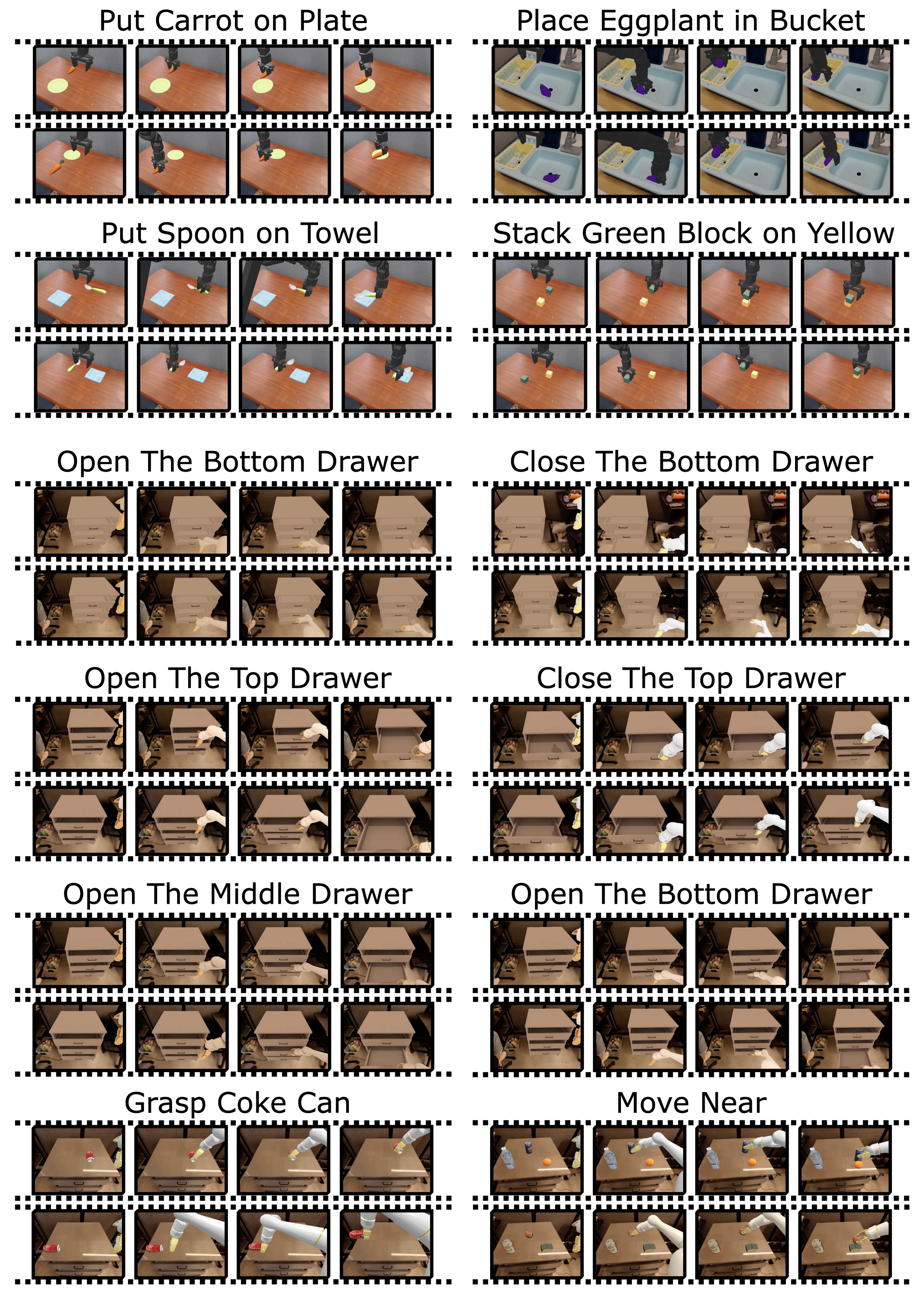}
  \caption{Qualitative results for SimplerEnv tasks.}
  \label{fig:simpler_ana}
\end{supplementaryfigure*}

\newpage

\section{Rollout Examples in Real-World Experiments}
\label{ap:real_quality}
\begin{supplementaryfigure*}[htbp]
    \centering
    \vspace{-10pt}
    \includegraphics[width=0.78\textwidth]{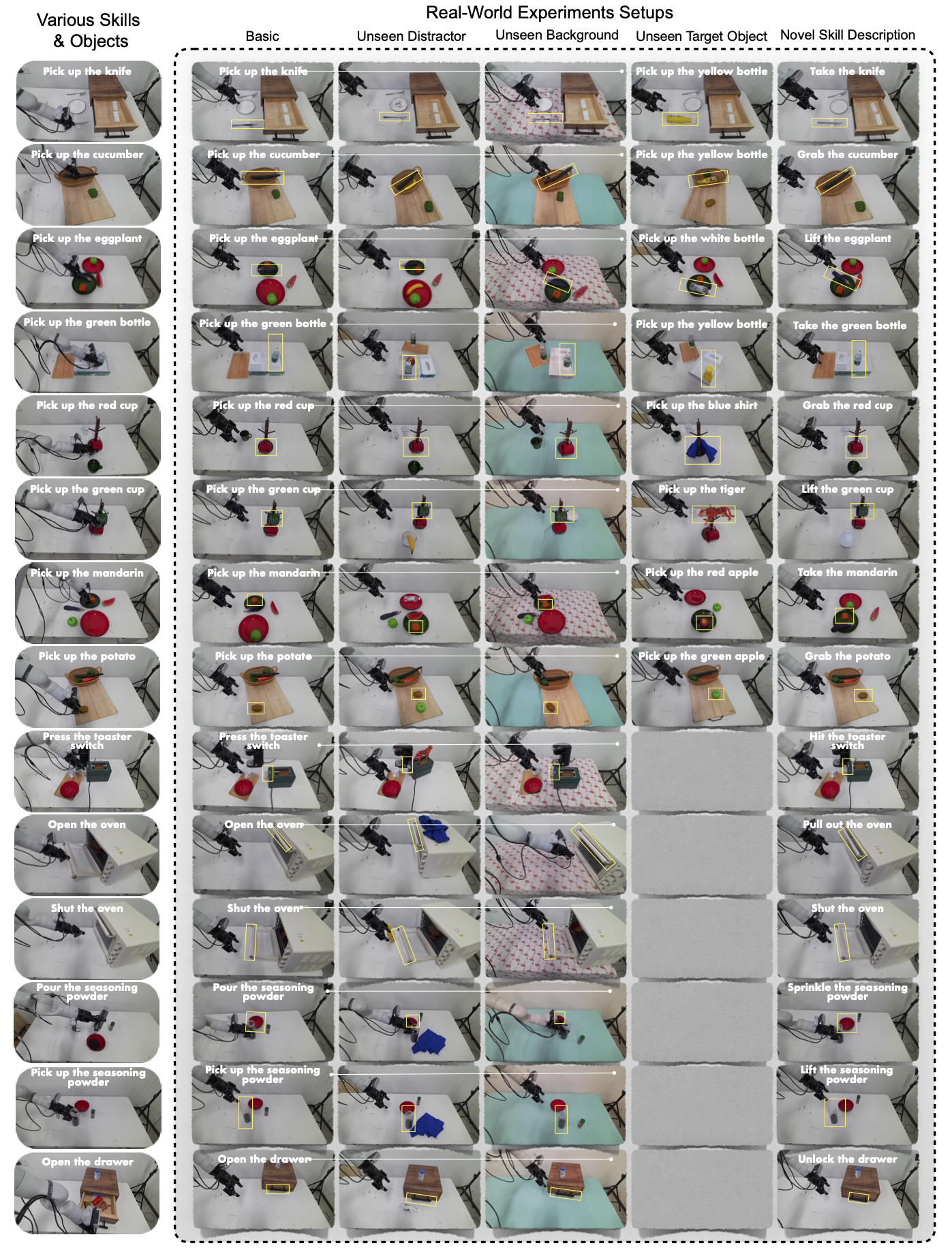}
    \caption{This figure illustrates the experimental setup of some real-world tasks. The models are evaluated across 20 tasks, each with 5 rollouts, involving unseen distractors, unseen backgrounds, unseen target objects, and novel skill descriptions. Note that some tasks exclude the unseen target object setting due to the lack of suitable alternative unseen objects.
    }
    \label{fig:real_illustration_full}
\end{supplementaryfigure*}

\begin{supplementaryfigure*}[htbp]
  \centering
    \includegraphics[width=0.8\textwidth]{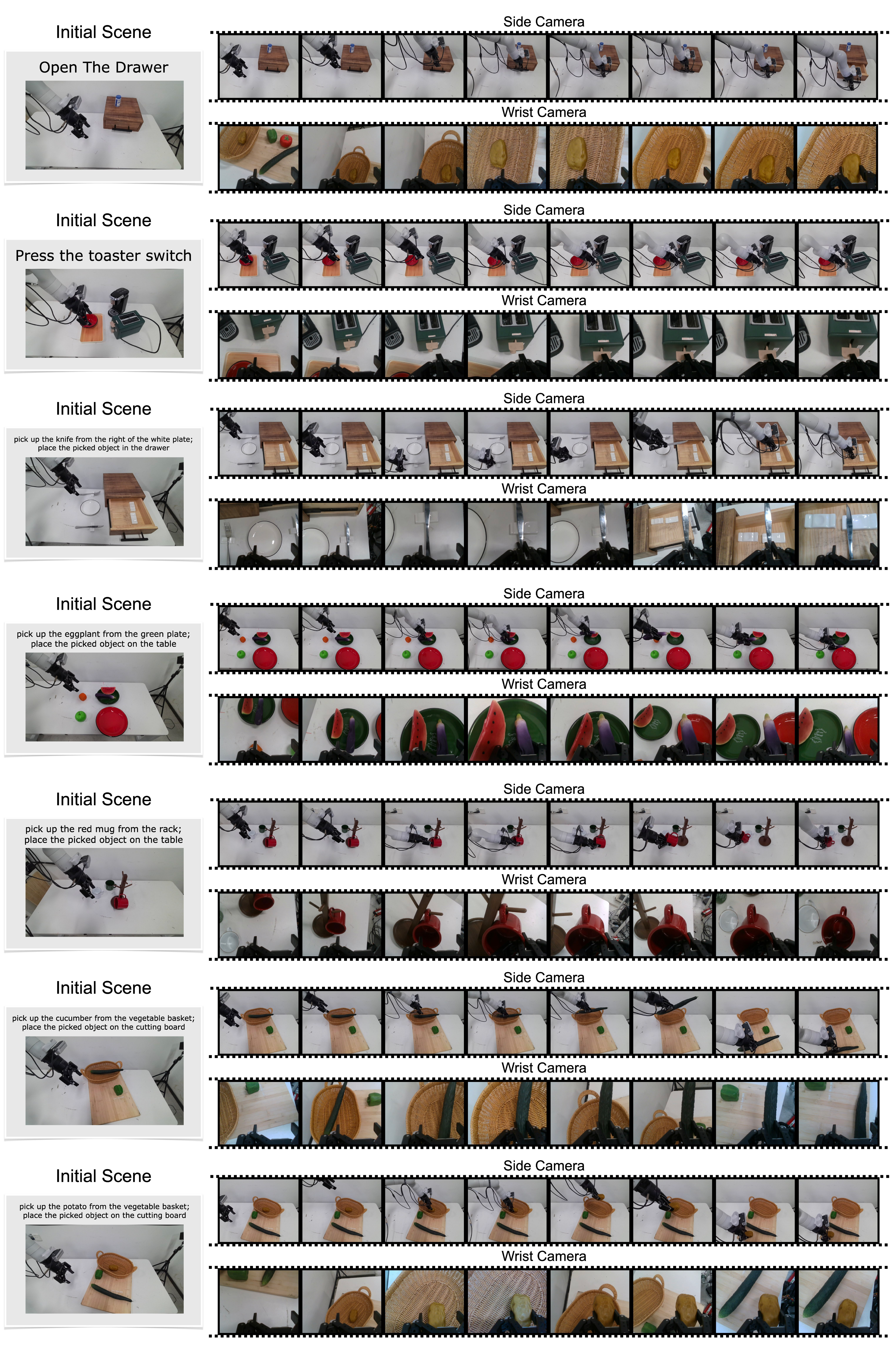}
  \caption{Qualitative results for basic setting in real-world experiments.}
  \label{fig:real_basic_ana}
\end{supplementaryfigure*}


\begin{supplementaryfigure*}[htbp]
  \centering
    \includegraphics[width=0.8\textwidth]{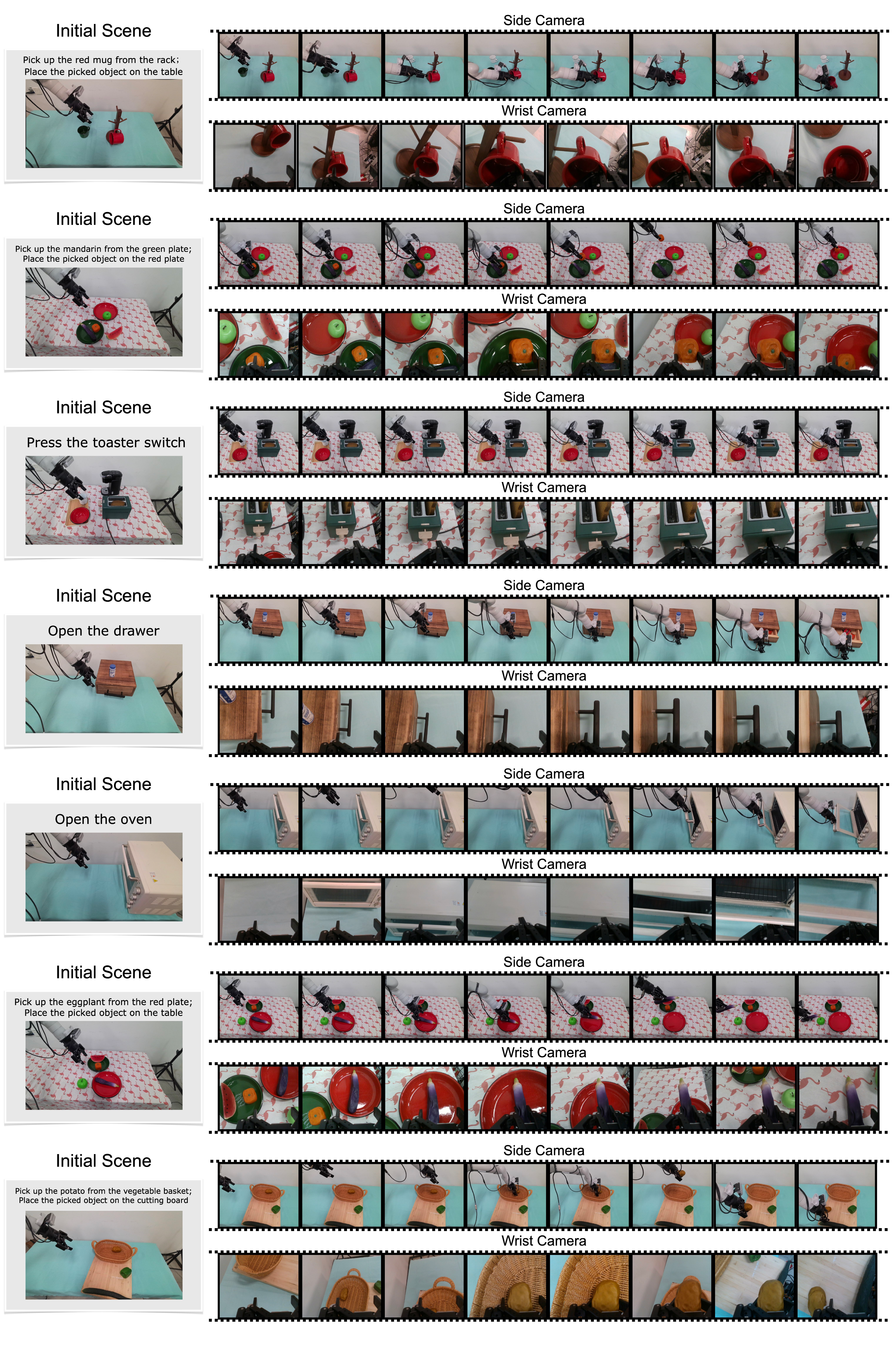}
  \caption{Qualitative results for unseen background.}
  \label{fig:real_unseen_bg}
\end{supplementaryfigure*}

\begin{supplementaryfigure*}[htbp]
  \centering
    \includegraphics[width=0.8\textwidth]{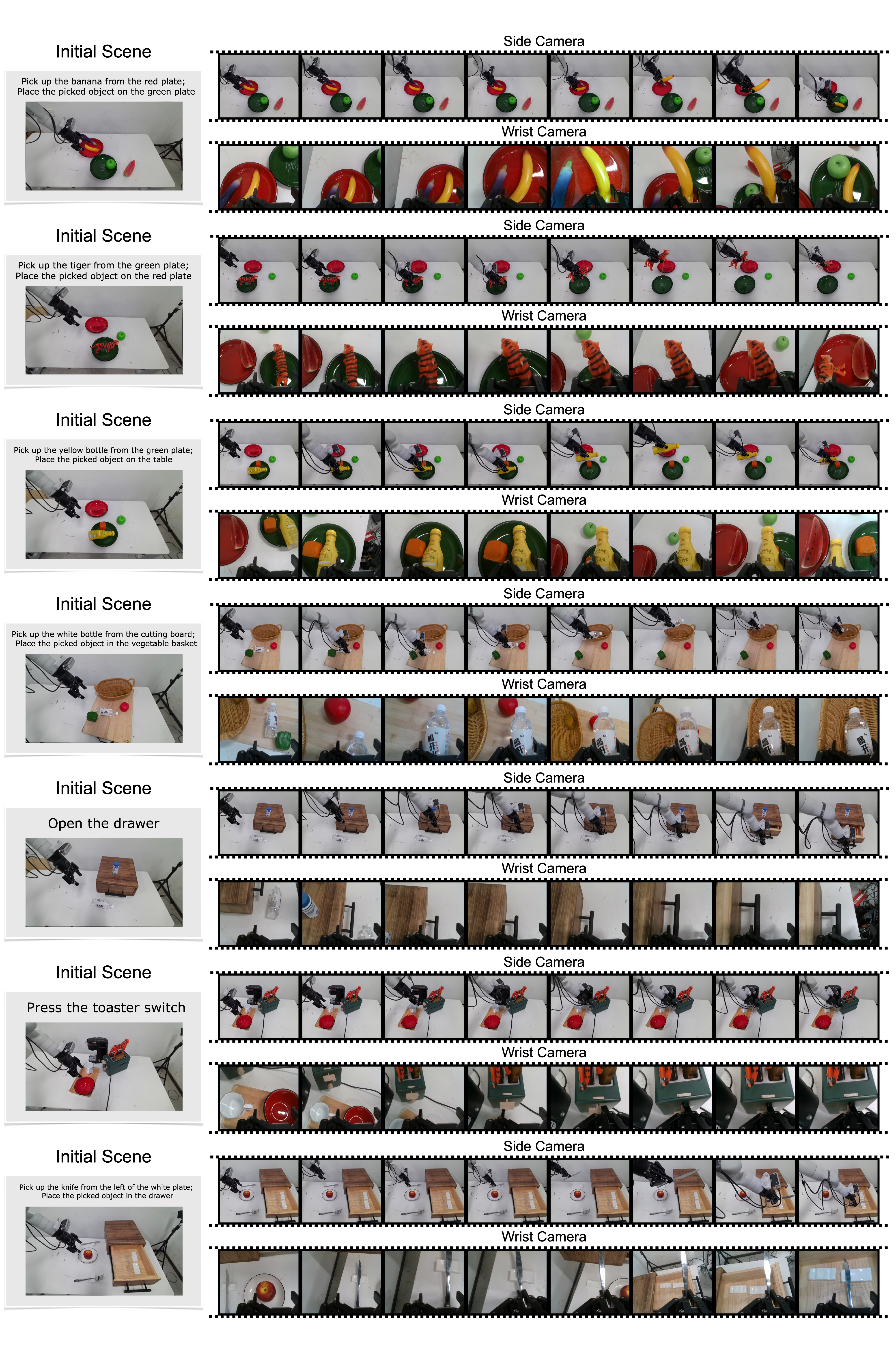}
  \caption{Qualitative results for unseen distractors and objects.}
  \label{fig:real_unseen_distractor_object}
\end{supplementaryfigure*}

\begin{supplementaryfigure*}[htbp]
  \centering
    \includegraphics[width=0.8\textwidth]{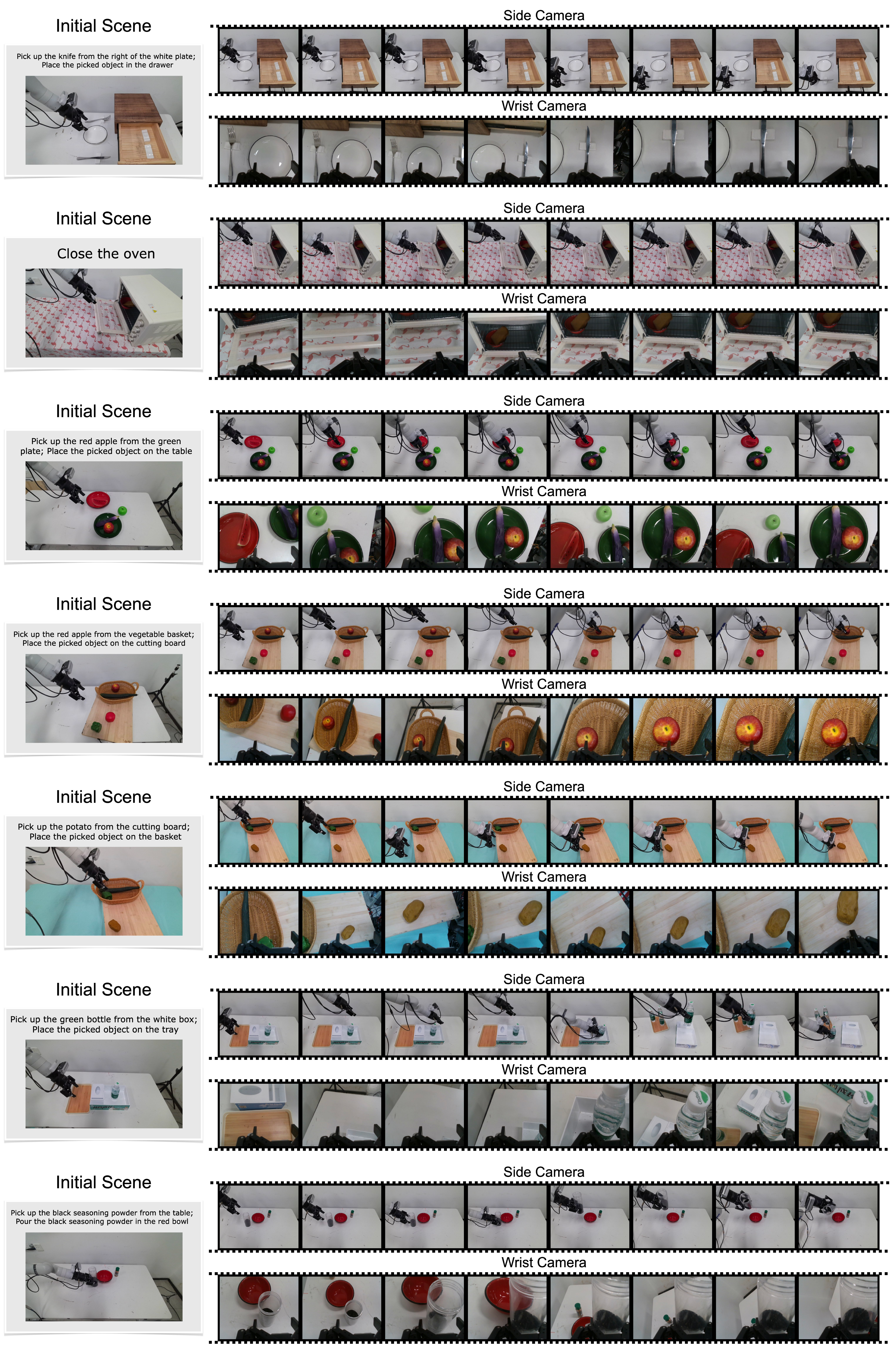}
  \caption{Our model exhibits several typical failure cases. For instance, it might prematurely close the gripper, fail to accurately grasp the target object, exhibit repeated oscillations, or successfully pick up an object but cannot place it in the correct location.}
  \label{fig:real_failure}
\end{supplementaryfigure*}

\end{appendices}
\end{document}